\documentclass[lettersize, journal,twoside]{IEEEtran}
\newcommand\highlightReference[1]{%
  \expandafter\newcommand\csname highlightReference-#1\endcsname{}%
}
\let\oldbibitem\bibitem
\def\bibitem#1 #2\par{%
  \expandafter\ifx\csname highlightReference-#1\endcsname\relax
    \oldbibitem{#1}#2\par
  \else
  \oldbibitem{#1}\highlight{#2}\par
  \fi
}
\usepackage[makeroom]{cancel}
\usepackage{subcaption}
\usepackage{amsmath}
 
\DeclareMathOperator*{\argmin}{argmin}
\usepackage{graphicx}
\usepackage{subfiles}
\usepackage[ruled,vlined]{algorithm2e}
\usepackage{multirow}
\usepackage{cite}
\usepackage{xcolor}
\newcommand\hlbreakable[1]{\textcolor{black}{#1}}
\usepackage{amsthm}
\usepackage{ulem}
\usepackage{color,soul}
\soulregister\cite7
\newcommand{\rs}[1]{}
\newcommand\highlight[1]{\textcolor{black}{#1}}
\usepackage[switch]{lineno}
\theoremstyle{definition}

\begin{document}
\title{SemiPFL: Personalized Semi-Supervised Federated Learning Framework for Edge Intelligence}
\author{Arvin~Tashakori, Wenwen~Zhang, Z.~Jane~Wang, and~Peyman~Servati\\
Department of Electrical and Computer Engineering\\
University of British Columbia\\
Vancouver BC, Canada
\thanks{}        
\thanks{}
\thanks{A. Tashakori is the corresponding author: arvin@ece.ubc.ca}}
\markboth{}%
{}
\maketitle
\begin{abstract}
Recent advances in wearable devices and Internet-of-Things (IoT) have led to massive growth in sensor data generated in edge devices. Labeling such massive data for classification tasks has proven to be challenging. In addition, data generated by different users bear various personal attributes and edge heterogeneity, rendering it impractical to develop a global model that adapts well to all users. Concerns over data privacy and communication costs also prohibit centralized data accumulation and training. \hlbreakable{We propose SemiPFL that supports} edge users having no label or limited labeled datasets and a sizable amount of unlabeled data that is insufficient to train a well-performing model. In this work, edge users collaborate to train a Hyper-network in the server, generating personalized autoencoders for each user. After receiving updates from edge users, the server produces a set of base models for each user, which the users locally aggregate them using their own labeled dataset. We comprehensively evaluate our proposed framework on various public datasets \hlbreakable{from a wide range of application scenarios, from wearable health to IoT,} and demonstrate that SemiPFL outperforms state-of-art federated learning frameworks under the same assumptions \hlbreakable{regarding user performance, network footprint, and computational consumption.} We also show that the solution performs well for users without \hlbreakable{label} or having limited labeled datasets and increasing performance for increased labeled data and number of users, signifying the effectiveness of SemiPFL for handling \hlbreakable{data} heterogeneity and limited annotation. \hlbreakable{We also demonstrate the stability of SemiPFL for handling user hardware resource heterogeneity in three real-time scenarios.}
\end{abstract}
\begin{IEEEkeywords}
Edge Computing, \hlbreakable{Edge Heterogeneity, }Personalized Federated Learning, Federated Learning, Edge Intelligence, Semi-Supervised Learning, Transfer Learning, Meta Learning, Sensor Analytics.
\end{IEEEkeywords}
\IEEEpeerreviewmaketitle
\section{Introduction}
\IEEEPARstart{O}{ver} the past years, the evolution of sensor and wearable technologies and the Internet of Things (IoT) devices has led to a wealth of personalized time-series multi-sensory data, tackling various real-life problems, including human activity recognition \cite{yousefi2017survey,chen2012sensor,ravi2005activity}, sleep stage identification \cite{supratak2017deepsleepnet} and fall detection \cite{mubashir2013survey}. The success in these application domains relies on supervised learning frameworks \cite{kim2009human, Wang2019DeepSurvey,lara2012survey}, where high-quality labeled data is necessary to train classification models in a strictly controlled context. However, acquiring a large amount of personalized labeled data in a centralized server is cost-prohibitive, time-consuming, and impractical due to complexity and privacy concerns \hlbreakable{\mbox{\cite{Saeed2021FederatedIntelligence,nguyen2021federated,ferrag2021federated}}}. \hlbreakable{On the other hand, users own devices with different processing powers, which makes training on user devices impractical for applications requiring extensive calculations \hlbreakable{\mbox{\cite{zhao2020semi}}}.} In summary, these problems have made it challenging to use an extensive amount of multi-sensor time-series data for predictive analytics tasks for each edge user.

Addressing the above concerns motivates us to answer the following research questions: Can we develop a framework to help users participate in cooperative training without violating their privacy? Can \hlbreakable{it} perform better than supervised learning methods with a few labeled data points? What is the influence of the number of available labeled instances on quality of training? Can the framework provide personalized models adapted to individual users instead of one global model? What is the impact of the number of users who are collaborating? \hlbreakable{In case of required processing on the user side, what is the effect of user hardware resource heterogeneity on the performance? And what is the performance of the proposed system compared to other methods?}

\begin{figure}
  \centering
   \includegraphics[width=1\linewidth]{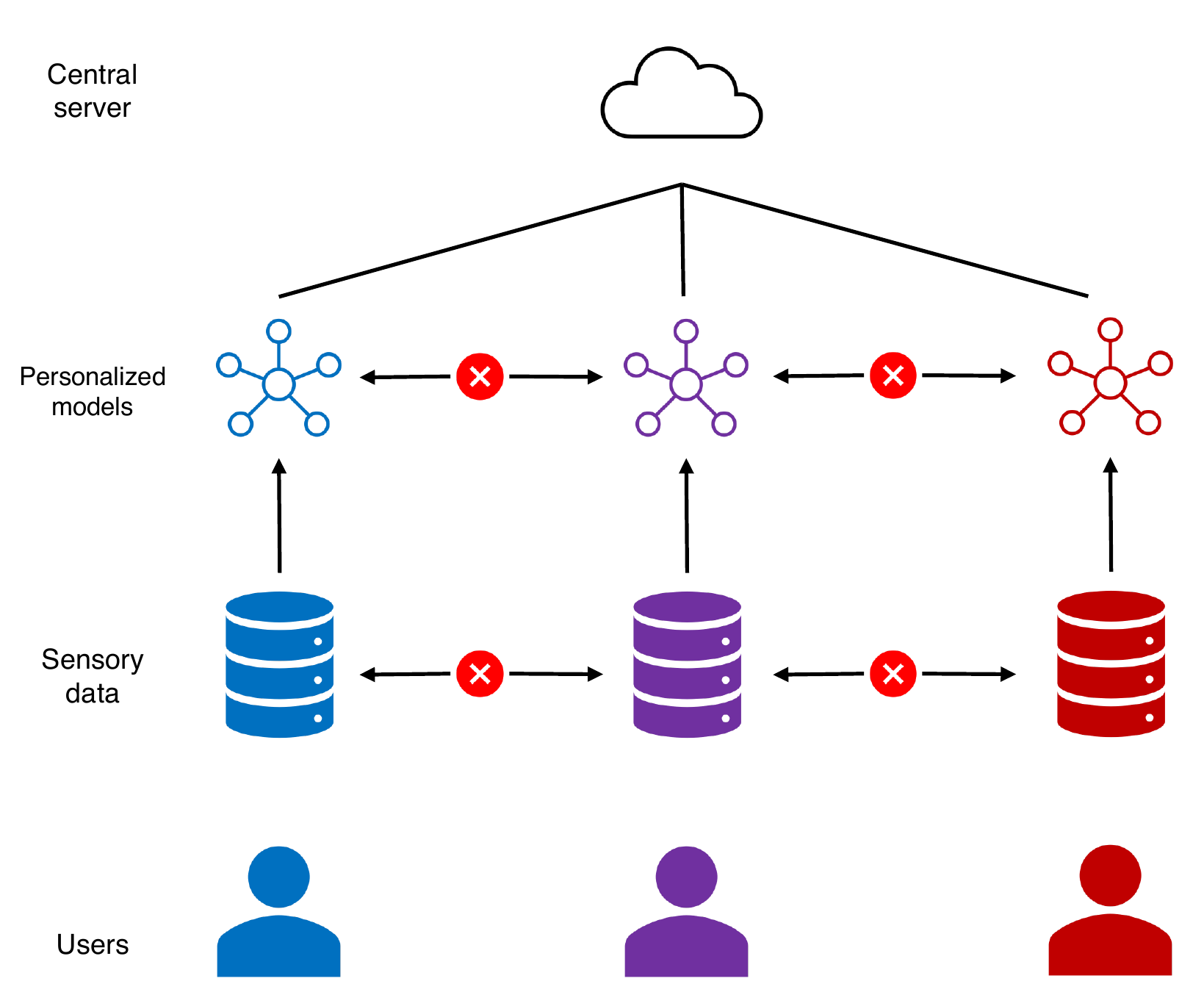}
    \caption{Overview of the general structure of personalized federated learning. The server communicates a personalized model to each user. Each user locally updates the model using their sensory data, and return the fine-tuned model back to the server.}
    \label{fig:overal1}
  \end{figure}

Federated learning can address the privacy concern regarding sharing of users' data with a centralized server \hlbreakable{\cite{nguyen2021federated,ferrag2021federated,imteaj2021survey}}. It has gained much attention in recent years, with applications in detection of industrial anomalies \cite{Liu2020DeepApproach,zhang2020blockchain}, air quality sensing \cite{liu2020federated}, traffic forecast \cite{Liuyi2020IPT}\hlbreakable{\cite{pei2022personalized}}, resource management in wireless networks \cite{yu2020deep,saha2020fogfl,nguyen2020efficient,xue2021resource}, human-computer interaction \cite{chhikara2020federated}, smart ocean \cite{kwon2020multiagent}, COVID-19 detection \cite{zhang2021dynamic,dayan2021federated}, medical imaging \cite{kaissis2020secure}, clinical decision systems \cite{xue2021resource,lim2020dynamic}, and embedded intelligence \cite{Saeed2021FederatedIntelligence,Tang2021SelfHAR:Data,Zhao2020Semi-supervisedRecognition,Chen2021FedHealthHealthcare,Chen2019FedHealth:Healthcare,yu2021fedhar}. In federated learning frameworks, users collaboratively train a global model while keeping their data isolated on the edge devices \cite{Lim2019FederatedSurvey,MLSYS2019_bd686fd6, xu2021federated}. The first method introduced by Google in 2016, FedAVG \cite{mcmahan2017communicationefficient}, is widely used as the baseline \cite{Bettini2021PersonalizedRecognition, yu2021fedhar, li2020federated}. In FedAVG, the server first sends a global model to each user. Later, edge users update the model using their locally labeled data and communicate the results back with the server. Finally, the server updates the global model by averaging the received model parameters from the users. 
Despite its simple framework, FedAVG demonstrated exceptional performances in several scenarios and applications \cite{Lim2019FederatedSurvey,Aledhari2020FederatedApplications,mcmahan2017communicationefficient}. At the same time, the performance of federated learning methods including FedAVG, degrades in scenarios where edge users have heterogeneous attributes\hlbreakable{, such as a) heterogeneity in computational power, which means users have different hardware settings, resulting in various processing delays \mbox{\cite{brik2020federated,Lim2019FederatedSurvey, cho2022flame,tan2022towards}}, b) heterogeneity in the distribution of data and labels \mbox{\cite{li2021fedrs,zhang2022federated,li2022data,huang2022learn,mendieta2022local,tan2022towards}}, c) heterogeneity in the learning task \mbox{\cite{brik2020federated,Lim2019FederatedSurvey,tan2022towards}},  d) heterogeneity in learning models \mbox{\cite{Tang2021SelfHAR:Data,tan2022towards}}, and e) heterogeneity in the number of devices per users \cite{cho2022flame}.} To address this challenge, recent works aimed at developing a personalized model for each user \cite{Chen2019FedHealth:Healthcare,Chen2021FedHealthHealthcare},\hlbreakable{\cite{cho2022flame,tan2022towards,wu2020personalized,hu2020personalized}}. An overview of personalized federated learning models is illustrated in Fig. \ref{fig:overal1}. In personalized federated learning, users collaborate to develop personalized models instead of generating a global model \cite{fallah2020personalized,Shamsian2021PersonalizedHypernetworks,Dinh2020PersonalizedEnvelopes,Bettini2021PersonalizedRecognition,yu2021fedhar,achituve2021personalized,Fallah2020PersonalizedApproach,li2021ditto}.
However, these works primarily focus on image classification benchmarks such as CIFAR10 \cite{cifar10}, and CIFAR100 \cite{cifar100}, and applications such as multi-sensor and \hlbreakable{IoT} classifications are left \hlbreakable{less }unattended. Also, in most of these works the main assumption is that sufficient labeled data exists at each user side, which may be impractical.

\begin{figure*}
     \centering
     \begin{subfigure}[b]{0.49\textwidth}
         \centering
         \includegraphics[width=\textwidth]{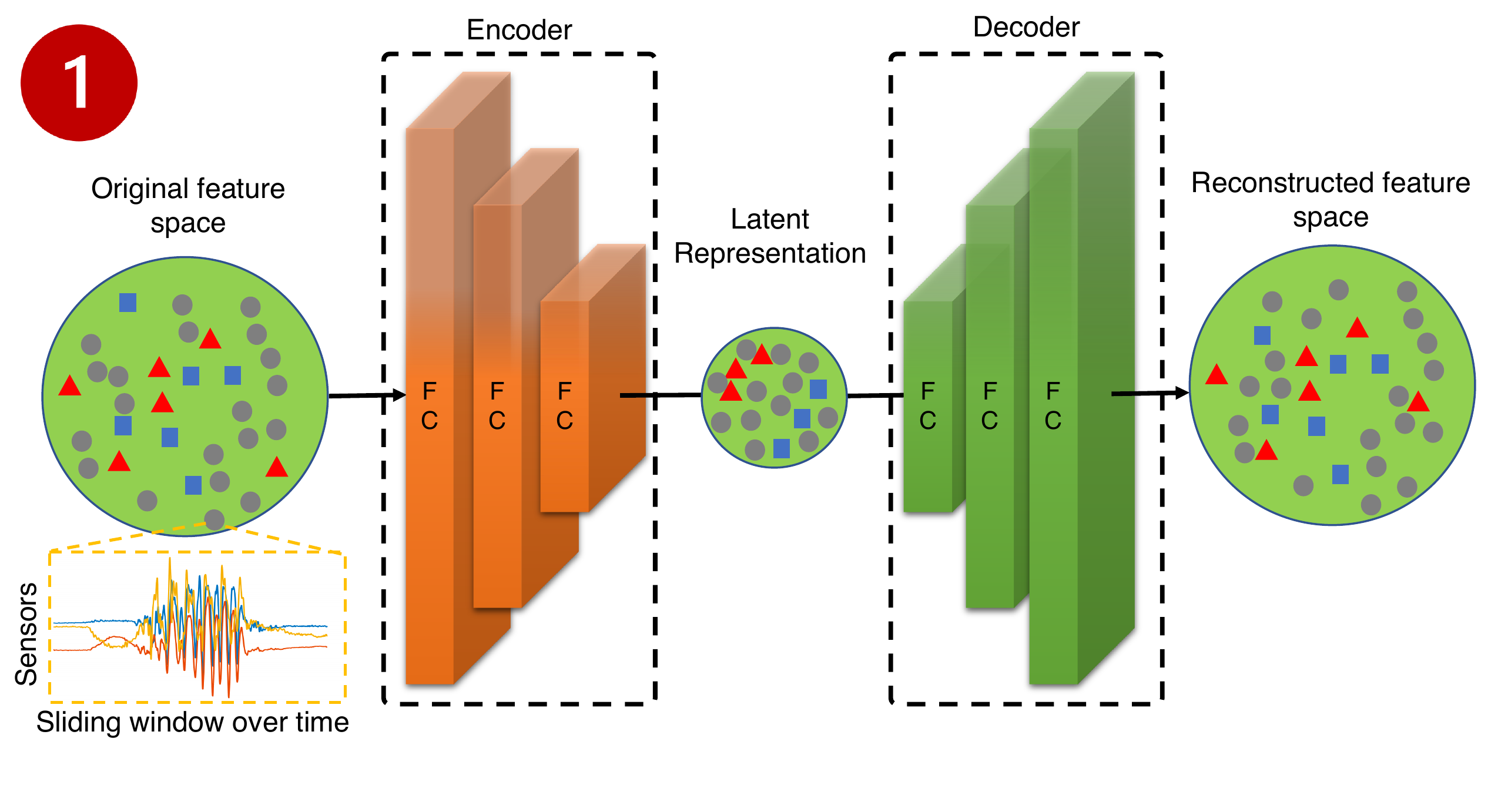}
     \end{subfigure}
     \hfill
     \begin{subfigure}[b]{0.49\textwidth}
         \centering
         \includegraphics[width=\textwidth]{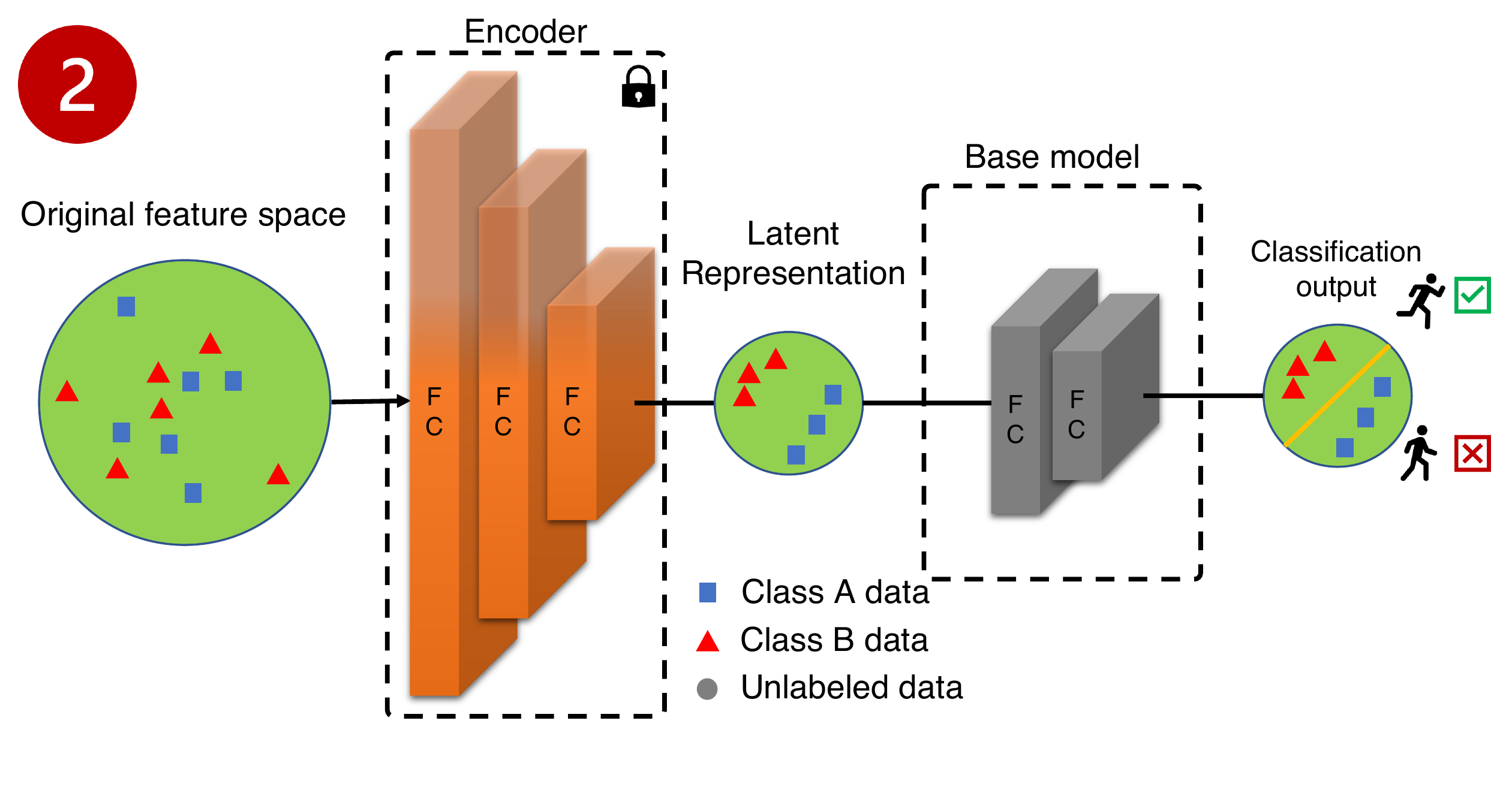}
     \end{subfigure}
     \caption{Overview of the semi-supervised learning method used in this work. First, we train an autoencoder on the whole labeled and unlabeled data. Then, we use the encoder to transform the data to its latent representation, and we use the labeled data to train a base classifier. The final model is the trained encoder followed by the base classifier.}
      \label{fig:semi}
\end{figure*}

Data annotation is another big challenge to address. Current solutions mainly employ supervised learning for sensor-based predictive analytic tasks. For example, in activity recognition, supervised learning is used to deduce activity categories from time-series sensory input data generated by edge sensors \cite{yang2015deep, ronao2016human,hammerla2016deep}. Data annotation is required by an expert in the field. In practice, we mostly have either no label or some labeled data leaving the bulk of data unlabeled on the edge devices. 

Although federated learning can protect user privacy, most of the existing methods rely on high degree of annotated data. However, due to the unpredictably variable characteristics of edge users and devices, achieving labeled data at the edge makes the current solutions impractical. To address these shortcomings, semi-supervised learning, which is defined between supervised and unsupervised learning, is proposed to deal with the insufficient labeled data problem \cite{van2020survey}. In semi-supervised learning, we have a small amount of labeled data and a large amount of unlabeled data. There have been multiple attempts to unify federated learning with semi-supervised learning \cite{zhao2020semi,yu2021fedhar,Bettini2021PersonalizedRecognition}\hlbreakable{,\mbox{\cite{Saeed2021FederatedIntelligence}}}. Zhao et al. consider users have a large amount of \hlbreakable{un}labeled data, while the server has a set of labeled data \cite{zhao2020semi}. Saeed et al. developed a self-supervised method to learn valuable representations from users' unlabeled data \cite{Saeed2021FederatedIntelligence}. None of these work investigates methods to solve the data heterogeneity problem. 

In this work, we aim to integrate semi-supervised learning with personalized federated learning for multi-sensory time-series-based classification. We assume that we have a set of labeled data from different distributions at the server and a small set of labeled data, and a large set of unlabeled data at the users' side. The proposed framework contains three main steps: first, the server learns to generate a personalized autoencoder using a Hyper-network for each user. Second, the server selects samples close to each user distribution, transforms its dataset using the corresponding user's encoder, and trains a set of base models that map the latent representation to output classes and send them to that specific user. Finally, the user aggregates the received models using its labeled datasets.

Our main contributions are as follows:
\begin{enumerate}
\item We propose SemiPFL, a semi-supervised personalized federated learning framework, for edge intelligence. SemiPFL applies to a wide range of scenarios, from supervised learning settings to having no labeled data available at the user side. To the best of our knowledge, our solutions is the first attempt to integrate personalized federated learning and semi-supervised learning for multi-sensory classifications. 
\item To our knowledge, this is the first work developing personalized models using a Hyper-network for multi-sensory classification.
\item \hlbreakable{To our knowledge, this is the first work that studies the effect of hardware resource heterogeneity among users.}
\item The proposed method outperforms recognition accuracy compared to recent publications using fully connected neural network architectures. 
\item We extensively evaluate the proposed framework on publicly available datasets gathered from edge devices such as smartphones and wearable devices. We show better performance of the proposed method as more users collaborate in the framework and more labeled instances are available on the user side.
\end{enumerate}

\section{Background and related work}
\hlbreakable{IoT solutions face several challenges, including a) scarcity and heterogeneity of computational and storage resources on edge devices \cite{Saeed2021FederatedIntelligence, zhao2020semi, cho2022flame}, b) heterogeneity of data on edge devices from distribution \cite{yu2021fedhar, cho2022flame}, availability of labeled data, and c) the requirement for having a smaller and less complex model architecture \cite{Saeed2021FederatedIntelligence, yu2021fedhar, cho2022flame}.}
The proposed SemiPFL aims at unifying semi-supervised learning with personalized federated learning for multi-sensory time-series edge inputs. In subsequent sections, we present the details of the main building blocks of our method and related works.

\begin{table}[t]
\caption{\textsc{Summary of notations used in this study.}}
\begin{center}
\resizebox{0.5\textwidth}{!}{\begin{tabular}{l | l} 
\hline
$R$ & Number of training rounds\\
\hline
$K$ & Number of users\\
\hline
$S$ & Number of sensors\\
\hline
$W$ & Sliding window size\\
\hline
$M$ & Number of different samples available on server side\\
\hline
$T, T'$ & Number of fine tuning epochs on user side\\
\hline
$h(.,.)$ & Hyper-network model\\
\hline
$\theta$ & Hyper-network parameters\\
\hline
$\{\alpha_j\}_{j=1}^K$ & User embedding vector\\
\hline
$D_S$ & Server data\\
\hline
$D_j$ & User j labeled data\\
\hline
$U_j$ & User j unlabeled data\\
\hline
$E_j$ & User j evaluation data\\
\hline
$\eta, \mu, \zeta$ & Learning rates\\
\hline
$\tau$ & Sample selection threshold\\
\hline
$\mathcal{L}^a(.,.)$ & Autoencoder loss function\\
\hline
$\mathcal{L}(.,.)$ & Supervised Learning loss function\\
\hline
\hlbreakable{$\mathcal{L}_j(.)$} & \hlbreakable{Local loss function for user j}\\
\hline
$f_r^{b,j}(.)$ & Base model for user j in round r\\
\hline
$f_{r,m}^{b,j}(.)$ & m-th base model for user j in round r\\
\hline
$f_r^{a_j}(.)$ & Autoencoder model for user j in round r\\
\hline
$\chi_{r,m}^{b,j}$ & Averaging weight for m-th base model for user j in round r\\
\hline
$f_j(.)$ & Final classification model for user j\\
\hline
\hlbreakable{$\Omega(.,...,.)$} & \hlbreakable{Regularization term}\\
\hline
\end{tabular}}
\end{center}
\label{table:notations}
\end{table}
\subsection{Semi-supervised learning}
Semi-supervised learning is one division of machine learning that falls between supervised and unsupervised learning \cite{van2020survey}, where generally we have a few labeled instances and a large number of unlabeled instances, and the number of labeled data points is insufficient to train a desirable supervised model \cite{van2020survey}. 

One major challenge in supervised learning frameworks is the availability of labeled instances. Data annotation is usually time-consuming, expensive, and not accessible at each edge point. \hlbreakable{SemiFL \mbox{\cite{Zhao2020Semi-supervisedRecognition}} addresses scenarios where users have unlabeled instances and collaborate to generate a global model. The authors assume that a set of labeled instances are available on the server-side. FedSCN \mbox{\cite{Saeed2021FederatedIntelligenceb}} proposes a self-supervised approach based on wavelet transform (WT) to learn valuable representations from unlabeled sensor inputs. However, both methods try to generate a global model for users, which does not perform well in scenarios where data heterogeneity exists among users.}

One popular approach to tackling semi-supervised learning is feature extraction by training an autoencoder on unlabeled instances. Autoencoder, an artificial neural network, learns data representation from unlabeled data. Its objective is to transform the original data to its compressed representation and reconstruct it back to its original form without losing valuable information. An autoencoder $(f^a(.) = f^a_{dec}(f^a_{enc}(.)))$ contains two sections, the encoder $f^a_{enc}(.)$, which maps instances to their latent representation, and the decoder $f^a_{dec}(.)$, which reconstructs the original data from its simplified representation. The learning objective of the autoencoder is to minimize the following loss function \cite{goodfellow2016deep}:

\begin{equation}
\mathcal{L}^a(X, f^a(X)) = \mathcal{L}^a(X, f^a_{dec}(f^a_{enc}(X)))
\label{eqn:genlossae}
\end{equation}
where $\mathcal{L}^a(.,.)$ is the autoencoder loss function, and $X \in {\rm I\!R}^{S \times W}$ the input data, where $S$ is the number of sensors, and $W$ the window width. This loss function penalizes the output of the autoencoder for being dissimilar to input $X$ \cite{goodfellow2016deep}. Training a classification model under a semi-supervised learning using the autoencoder requires two steps. In the first step, we train an autoencoder on all labeled and unlabeled instances. Then using the encoder part of the trained autoencoder, we compress the original data to its latent representation, which provides a compact representation of original instances. Subsequently, we train a base classifier using the transformed labeled data points. The final classifier will be the encoder, followed by the base classifier\cite{goodfellow2016deep,Zhao2020Semi-supervisedRecognition}. Fig. \ref{fig:semi} gives an overview of the proposed approach.

\begin{figure*}
  \centering
   \includegraphics[width=1\linewidth]{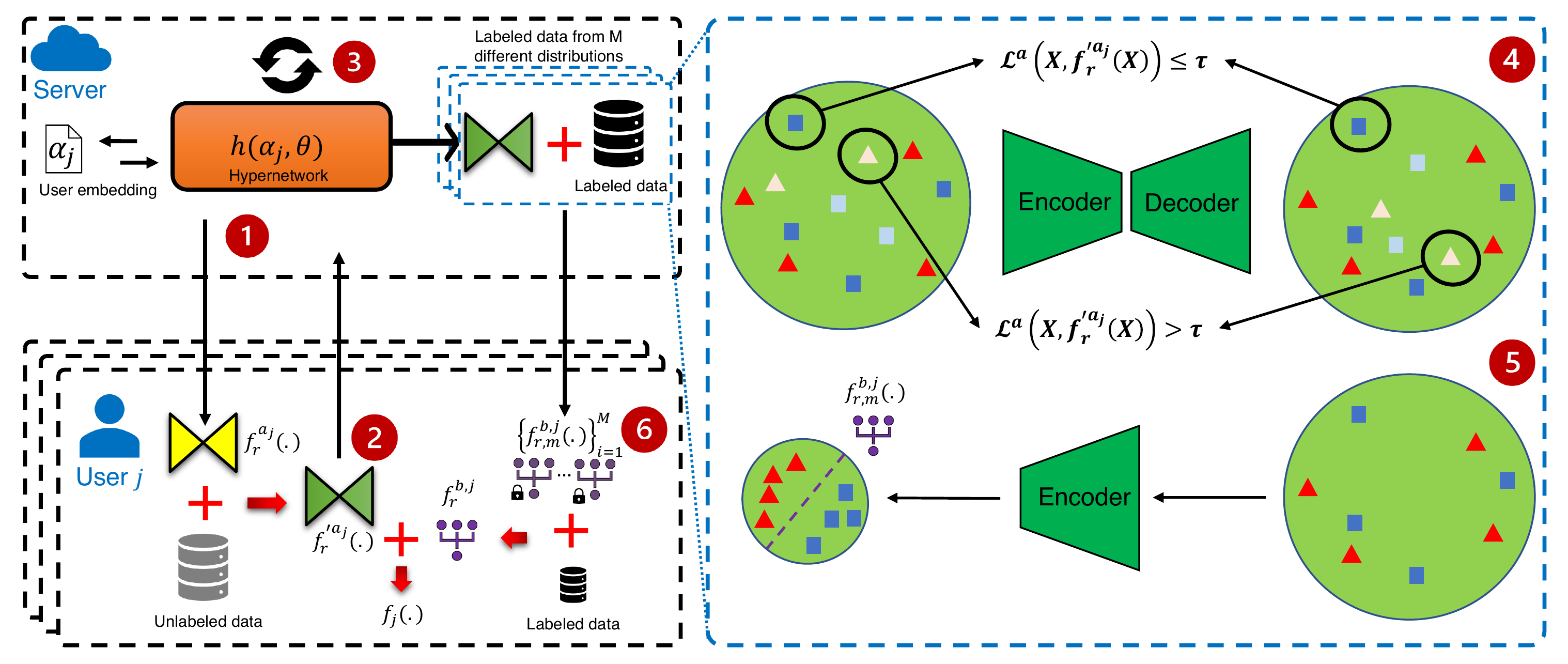}
    \caption{Overall architecture of the proposed SemiPFL. 1) The server selects a user, generates a personalized autoencoder using the user embedding $\alpha_j$ via a Hyper-network, and sends it to the user. 2) The user fine-tunes the model using its unlabeled data and sends it back to the server. 3) The server updates the user embedding document and Hyper-network using the fine-tuned model. 4) The server encodes its labeled data using the encoding part of the user autoencoder, and 5) trains a set of base models using supervised learning and sends it to the user. 6) The user generates its base model by aggregating them using labeled data.}
    \label{fig:overal}
  \end{figure*}

\subsection{Personalized Federated learning}
IoT devices such as smartphones and wearable devices generate a massive amount of data every day. Traditional machine learning approaches require us to accumulate user data in a centralized database to train supervised models. However, this task is not practical due to several challenges such as privacy and bandwidth limitations \cite{Aledhari2020FederatedApplications,Saeed2021FederatedIntelligence,Kairouz2019AdvancesLearning}. Moreover, the growing computational power of edge devices makes them suitable for local computation and data storage \cite{zhao2020semi}.

Federated learning seeks to provide the same collaborative training without sharing data. In FedAVG \cite{mcmahan2017communicationefficient}, the server aggregates all user models without particular non-iid data operations (Eqn. (\ref{fedavg})). $f_{r}^{j}(.)$ means user-j model parameters in the r-th round, K is the total number of users, and $f_{r+1}^{S}(.)$ the global model parameters for round $r+1$ that will be sent to all users at the beginning of round $r+1$. \hlbreakable{However, in IoT applications, heterogeneity always exists among users, such as data heterogeneity or hardware resource heterogeneity, making it challenging to train a global model that performs adequately well for all users.}

\begin{equation}
f_{r+1}^S(.) \gets \frac{1}{K} \sum_{j=1}^K f_{r}^{j}(.)
\label{fedavg}
\end{equation}

\hlbreakable{Recent literature focuses on personalized federated learning to address data heterogeneity, generating personalized models for different users.} FedBN \cite{li2021fedbn} assumes that each user keeps the local batch normalization while collaborating to generate a global model. FedPer \cite{arivazhagan2019federated} tries to generate a personalized model for each user by preserving some local layers. FedProx \cite{li2020federated} is a generalized version of FedAVG. It allows partial information aggregation and adds a proximal term to FedAVG. Fedhealth \cite{Chen2019FedHealth:Healthcare} is the first personalized federated learning method introduced for wearable healthcare devices through transfer learning. In Fedhealth, users freeze the global model and fine-tune the last layers using their labeled dataset. FedHealth 2 \cite{Chen2021FedHealthHealthcare} measures users' similarities with the pre-trained model and then aggregates all weighted models while users keep their batch normalization layer. \hlbreakable{Flame \cite{cho2022flame} focuses on a multi-device-environment (MDE), where each user can have multiple devices with different processing power. They propose personalized federated learning for scenarios where users can have sufficient labeled data on one of their edge devices. In PfedHN \cite{shamsian2021personalized}, authors generate personalized models for users using Hyper-networks. Hyper-networks are one of the widely used meta-learning techniques that output the weights of another target network that performs the learning task \cite{shamsian2021personalized,ha2016hypernetworks}. The idea is that the output weights vary depending on the input to the Hyper-network \cite{shamsian2021personalized,ha2016hypernetworks}.
Hyper-networks are widely used in various machine learning domains, including language modeling, computer vision, continual learning, hyper-parameter optimization, multi-objective optimization, and decoding block codes \cite{shamsian2021personalized,ha2016hypernetworks}.
Hyper-networks are naturally suitable for learning a diverse set of personalized models, as Hyper-networks dynamically generate target networks conditioned on the input \cite{shamsian2021personalized,ha2016hypernetworks}.}

\hlbreakable{In IoT applications, hardware heterogeneity is another major challenge that requires users to have personalized models. However, in the literature, no work has focused on the effect of differences in user processing power on the overall performance of the method. In this work, we provide a baseline to compare the effect of hardware resource heterogeneity on the overall performance of federated learning algorithms, which we will discuss in section IV.}

\section{SemiPFL: Proposed framework}

This section introduces SemiPFL, our personalized semi-supervised federated learning framework for time-series multi-sensory data. SemiPFL consists of three main steps: first, the server generates a personalized autoencoder using a Hyper-network for each user. Second, the server selects samples close to each user distribution, transforms its dataset using the corresponding user's encoder, and trains a set of base models that map latent representation to classes and send it to that specific user. In the third and final step, the user aggregates the received models using its labeled dataset. We will first describe the problem formulation in subsequent sections and then discuss our training pipeline.

\subsection{Problem description}
\quad In SemiPFL, similar to a semi-supervised learning scenario, we assume that users have a small set of labeled and a large set of unlabeled datasets. We also consider a set of labeled instances in the server. In other words, we assume that we have $K$ users and each user $j$ has a small set of labeled data $D_j$:
\begin{equation}
D_j = \left\{ X_j^i, Y_j^i \right\}_{i=1}^{l_j} \forall j \in {1, 2, ..., K}
\end{equation}
where $\left\{ X_j^i, Y_j^i \right\}$ are the i-th input and output pairs for user j respectively. We also have a large unlabeled dataset for user j, denoted as $U_j$:
\begin{equation}
U_j = \left\{ \overline{X_j^{i}}\right\}_{i=1}^{u_j} \forall j \in {1, 2, ..., K}
\label{eqn:unlabel}
\end{equation}
To evaluate our method, we define $E_j$ as user j-th evaluation dataset, where $\left\{ \widehat{X_j^{i}}, \widehat{Y_j^{i}} \right\}$ are the i-th input and output pairs for user j evaluation dataset, respectively:
\begin{equation}
E_j = \left\{ \widehat{X_j^{i}}, \widehat{Y_j^{i}} \right\}_{i=1}^{e_j} \forall j \in {1, 2, ..., K}
\label{eqn:eval}
\end{equation}
Also, there is a high-resolution dataset from $M$ different distributions available on the server-side:
\begin{equation}
D_S = \left\{D_{S_m}\right\}_{m=1}^M =\left\{\{ X_{S_m}^i, Y_{S_m}^i \}_{i=1}^{l_{S_m}}\right\}_{m=1}^{M}
\end{equation}
The goal of traditional federated learning methods such as FedAVG is to find a global model $f(.)$ through minimizing Eqn. (\ref{eqn:genloss}):
\begin{algorithm}[ht]
    \SetKwInOut{Input}{Input}
    \SetKwInOut{Output}{Output}
    \SetAlgoLined
    \Output{Personalized models:$\{f_j(.)\}_{j=1}^K$}
    \For{$r = 0, 1, \cdots,  R-1$}{
        1.1- Server randomly select user $j \in [K]$\;
        \text{1.2-} ${f_r}^{a_j}(.) = h (\alpha_j, \theta)$\;
        1.3- ${f'}_r^{a_j}(.)\gets f_r^{a_j}(.)$\;
        2- \For{$t \in [T]$}{
        2.1- User j sample mini-batch $\gamma \subset U_j \bigcup \{X|(X,Y)\in D_j\}$\;
        2.2- ${f'}_r^{a_j}(.) \gets {f'}_r^{a_j}(.) - \eta \nabla_{{f'}_r^{a_j}(.)} \mathcal{L}^a_j(\gamma)$\;
        }
        3- Eqn. (\ref{equ:update_ae})\;
        4- Eqn. (\ref{equ:SampleSelection})\;
        5.1- ${D'}^j_{S_m}\gets\big\{({f'}_{r,enc}^{a_j}(X),Y) | (X,Y) \in D_{S_m}\big\} \forall m \in [M]$\;
        5.2- $\{f_{r,m}^{b,j}(.)\}_{m=1}^M \gets \{f^b(.)\}_{m=1}^M$\;
        5.3- \For{$\{t,m\} \in [T\times M]$}{
        5.3.1- Server sample mini-batch $\gamma \subset {D'}^j_{S_m}$\;
        5.3.2- $f_{r,m}^{b,j}(.) \gets f_{r,m}^{b,j}(.) - \eta \nabla_{f_{r,m}^{b,j}(.)} \mathcal{L}(\gamma)$\;
        }
        6.1 Freeze $\{f_{r,m}^{b,j}(.)\}_{m=1}^M$\;
        6.2 Initialize $\left\{\chi_m\right\}{m=1}^M = \left\{\frac{1}{M}\right\}_{m=1}^M$\;
        6.2- ${f'}_r^{b,j}(.) \gets \sum_{m=1}^M \chi_m . f_{r,m}^{b,j}(.)$\;
        6.3- ${D'}_j \gets \Big\{({f'}_{r,enc}^{a_j}(X),Y)|(X,Y) \in D_j\Big\}$\;
        6.4- \For{$t \in [T']$}{
        6.4.1- User j sample mini-batch $\gamma \subset {D'}_j$\;
        6.4.2-${f'}_r^{b,j}(.) \gets {f'}_r^{b,j}(.) - \eta \nabla_{{f'}_r^{b,j}(.)} \mathcal{L}_j(\gamma)$\;
        Subject to: $\sum_{m=1}^M \chi_m =1$\;
        }
        6.5- $f_j(.) \gets {f'}_r^{b,j}({f'}_{r,enc}^{a_j}(.))$ \;
        }
 \caption{SemiPFL}
 \label{algorithmSemiPFL}
\end{algorithm}
\begin{equation}
f(.) = \argmin_{f(.)} \frac{1}{K}\sum_{j=1}^{K} \frac{1}{e_j}\sum_{i=1}^{e_j}\hlbreakable{\mathcal{L}}\big(\widehat{Y_j^{i}}, f(\widehat{X_j^{i}})\big) 
\label{eqn:genloss}
\end{equation}
where $\mathcal{L}(.,.)$ denotes the loss function. In traditional models, although we could find a model that minimizes the general loss (Eqn. (\ref{eqn:genloss})), due to the domain shift between users, the general model would not perform satisfactorily for all users. The goal of SemiPFL is to generate a set of personalized models for each user. Particularly, our objective is to propose a framework where the server provides a personalized model $\left\{f_j(.)\right\}_{i=1}^{K}$ for each user. \hlbreakable{Based on \cite{ding2022federated}, the overall objective of personalized federated learning from the server's perspective can be formulated as Eqn. (\ref{eqn:genlosspflnew}):}
\begin{table*}[ht]
\caption{\textsc{Summary of datasets used in this study.}}
\begin{center}
\resizebox{0.93\textwidth}{!}{\begin{tabular}{c c c c c c c c c c} 
\hline
 \multirow{2}{*}{Task} & \multirow{2}{*}{Dataset} & \multirow{2}{*}{Users}& \multirow{2}{*}{Trials}& \multirow{2}{*}{Dimensions} & \multirow{2}{*}{Output} & \multicolumn{2}{c}{\hlbreakable{Intra-subject CV}} & \multicolumn{2}{c}{\hlbreakable{Inter-subject CV}} \\
&&&&&&\hlbreakable{F1 score} & \hlbreakable{Kappa} & \hlbreakable{F1 score} & \hlbreakable{Kappa} \\
\hline
 \multirow{6}{*}{Activity recognition} & \multirow{2}{*}{Mobiact \cite{10.1007/978-3-319-62704-5_7}} & \multirow{2}{*}{61} &\multirow{2}{*}{6}& \multirow{2}{*}{9}&5 & \hlbreakable{86.92} & \hlbreakable{80.21} &  \hlbreakable{86.34} & \hlbreakable{82.52}\\
 &  & & &&10 & \hlbreakable{65.24} & \hlbreakable{59.89} & \hlbreakable{88.49} & \hlbreakable{82.54} \\
\cline{2-6}
& WISDM \cite{weiss2019smartphone} & 51 & 1 & 6 & 6 & \hlbreakable{83.21} & \hlbreakable{80.32} & \hlbreakable{93.87} & \hlbreakable{92.31}\\
\cline{2-6}
& HAR-UCI \cite{anguita2013public} & 30 & 1 & 6 & 6 & \hlbreakable{83.54} & \hlbreakable{80.73} & \hlbreakable{98.76} & \hlbreakable{98.51}\\
\cline{2-6}
& HHAR \cite{stisen2015smart}& 9 & 1 & 12 & 6 & \hlbreakable{63.35} & \hlbreakable{55.44} & \hlbreakable{96.53} & \hlbreakable{95.50}\\
\cline{2-6}
& PAMAP2 \cite{reiss2012introducing,reiss2012creating}& 8 & 1 & 39 & 10 & \hlbreakable{77.67} & \hlbreakable{76.49} & \hlbreakable{99.99} & \hlbreakable{99.99}\\
\hline
Stress Detection & WESAD \cite{schmidt2018introducing} & 15 & 1 & 7 & 3 & \hlbreakable{78.87} & \hlbreakable{70.60} & \hlbreakable{91.03} & \hlbreakable{87.44}\\
\hline
\hlbreakable{Sleep stage scoring} & \hlbreakable{Sleep-EDF \cite{goldberger2000physiobank,stisen2015smart}} & \hlbreakable{20} & \hlbreakable{1} &\hlbreakable{7} & \hlbreakable{5} & \hlbreakable{51.97} & \hlbreakable{48.78} & \hlbreakable{82.32} & \hlbreakable{76.77}\\
\hline
\end{tabular}}
\end{center}
\label{table:datasets}
\end{table*}
\begin{table*}[t]
\caption{\textsc{Performance comparisons with other federated learning methods on several datasets}}
\begin{center}
\resizebox{0.93\textwidth}{!}{\begin{tabular}{c c c c c c c c c c} 
\hline
Dataset&method&	personalized&label in server&label in user&\hlbreakable{devic heterogeneity}&model architecture&\multicolumn{2}{c}{Metric}&\textbf{SemiPFL (FCNN)}\\
\hline
HAR-UCI \cite{anguita2013public}& FedHAR \cite{yu2021fedhar}& Yes & No & Few & \hlbreakable{No} &FCNN & F1 score & 92.32 & \textbf{95.29}\\
\hline
\multirow{1}{*}{WISDM \cite{weiss2019smartphone}}&\multirow{2}{*}{FedAR \cite{Bettini2021PersonalizedRecognition}} & \multirow{2}{*}{Yes} & \multirow{2}{*}{No} & \multirow{2}{*}{Few} &\multirow{2}{*}{\hlbreakable{No}} &\multirow{2}{*}{FCNN} & F1 score &65&\textbf{89.05}\\
\cline{1-1}
\cline{8-10}
\multirow{1}{*}{Mobiact \cite{10.1007/978-3-319-62704-5_7} (5 activities)}&&&&&&&F1 score &86&\textbf{98.13}\\
\hline
\multirow{2}{*}{Mobiact \cite{10.1007/978-3-319-62704-5_7} (10 activities)}&\multirow{8}{*}{FedSCN \cite{Saeed2021FederatedIntelligence}} & \multirow{8}{*}{No} & \multirow{8}{*}{No} & \multirow{8}{*}{Few} &\multirow{8}{*}{\hlbreakable{ No}}&\multirow{8}{*}{CNN+Linear} & F1 score&90&\textbf{93.98}\\
&&&&&&&Kappa&87&\textbf{86.94}\\
\cline{1-1}
\cline{8-10}
\multirow{2}{*}{HHAR \cite{stisen2015smart}}&&&&&&&F1 score&80&\textbf{88.35}\\
&&&&&&&Kappa&76&\textbf{86.51}\\
\cline{1-1}
\cline{8-10}
\multirow{2}{*}{WESAD \cite{schmidt2018introducing}}&&&&&&&F1 score&82&\textbf{88.03}\\
&&&&&&&Kappa&70&\textbf{84.46}\\
\cline{1-1}
\cline{8-10}
\multirow{2}{*}{\hlbreakable{Sleep-EDF \cite{goldberger2000physiobank,stisen2015smart}}}&&&&&&&\hlbreakable{F1 score}&\hlbreakable{77}&\hlbreakable{\textbf{79.07}}\\
&&&&&&&\hlbreakable{Kappa}&\hlbreakable{69}&\hlbreakable{\textbf{74.06}}\\
\hline

\multirow{7}{*}{PAMAP2 \cite{reiss2012introducing,reiss2012creating}}&\multirow{1}{*}{Fedhealth2 \cite{Chen2021FedHealthHealthcare}} & \multirow{1}{*}{Yes} & \multirow{1}{*}{No} & \multirow{1}{*}{Yes} &\hlbreakable{No} & \multirow{1}{*}{FCNN} & F1 score&81.14&\multirow{7}{*}{\textbf{91.03}}\\
\cline{2-9}

&\multirow{1}{*}{FedAVG \cite{mcmahan2017communicationefficient}} & \multirow{1}{*}{No} & \multirow{1}{*}{No} & \multirow{1}{*}{Yes} &\hlbreakable{No} & \multirow{1}{*}{FCNN}  &F1 score&67.58&\\
\cline{2-9}

&\multirow{1}{*}{FedBN \cite{li2021fedbn}} & \multirow{1}{*}{Yes} & \multirow{1}{*}{No} & \multirow{1}{*}{Yes} &\hlbreakable{No} & \multirow{1}{*}{FCNN}  &F1 score&67.56&\\
\cline{2-9}

&\multirow{1}{*}{FedProx \cite{li2020federated}} & \multirow{1}{*}{Yes} & \multirow{1}{*}{No} & \multirow{1}{*}{Yes} &\hlbreakable{No} & \multirow{1}{*}{FCNN} &F1 score&67.52&\\
\cline{2-9}

&\multirow{1}{*}{FedPer\cite{arivazhagan2019federated}} & \multirow{1}{*}{Yes} & \multirow{1}{*}{No} & \multirow{1}{*}{Yes} &\hlbreakable{No} & \multirow{1}{*}{FCNN}  &F1 score&64.59&\\
\cline{2-9}

&\multirow{1}{*}{SemiFL\cite{zhao2020semi}} & \multirow{1}{*}{No} & \multirow{1}{*}{Yes} & \multirow{1}{*}{No} & \hlbreakable{No} &\multirow{1}{*}{LSTM + FCNN}  &F1 score&82&\\
\cline{2-9}
&\multirow{1}{*}{\hlbreakable{Flame\cite{cho2022flame}}} & \multirow{1}{*}{\hlbreakable{Yes}} & \multirow{1}{*}{\hlbreakable{No}} & \multirow{1}{*}{\hlbreakable{Yes}} &\hlbreakable{Yes} & \multirow{1}{*}{\hlbreakable{Deep ConvLSTM \cite{khatun2022deep}}}  &\hlbreakable{F1 score}&\hlbreakable{53}&\\

\hline
\end{tabular}}
\end{center}
\label{table:litreview}
\end{table*}
{\color{black}\begin{equation}
\left\{f_j(.)\right\}_{j=1}^K=\argmin_{\left\{f_j(.)\right\}_{j=1}^K}\frac{1}{K} \sum_{j=1}^{K} \mathcal{L}_j(f_j(E_j))+\Omega(f_1,...,f_K)
\label{eqn:genlosspflnew}
\end{equation}}
\hlbreakable{where $\mathcal{L}_j(.)$ is user j's local loss function (Eqn. (\ref{eqn:localloss})), and $\Omega(.,...,.)$  denotes a regularization term that differentiates personalized federated learning from separately training $K$ personalized models. Therefore, our objective is to find personalized models for each user, assuming that we have limited or no labeled data, large set of unlabeled data available on the users' side, and a set of publicly available labeled data on the server side.} 
{\color{black}\begin{equation}
    \mathcal{L}_j(f_j(E_j))=\frac{1}{e_j}\sum_{i=1}^{e_j} \mathcal{L}\big(\widehat{Y_j^{i}}, f_j(\widehat{X_j^{i}})\big)
    \label{eqn:localloss}
\end{equation}}
\rs{the overall objective can be formulated as Eqn. (\mbox{\ref{eqn:genlosspfl}}):}

\rs{In Eqn. (\mbox{\ref{eqn:genlosspfl}}), since there is no dependence among $\xcancel{\left\{f_j(.)\right\}_{j=1}^K}$, we can simplify Eqn. (\mbox{\ref{eqn:genlosspfl}}) to Eqn. (\mbox{\ref{eqn:genlosspflsim}}) as:}
\rs{In other words, the personalized federated learning objective is to evaluate each personalized model generated for each user on that user's dataset. In the experimental evaluation section, we measure the performance based on the average of each user's F1 and Kappa scores on their own dataset.}
\subsection{Training pipeline}
This section introduces SemiPFL, our novel personalized semi-supervised federated learning framework. It consists of several steps that require communication between the server and users. The overall system model can be found in Fig. \ref{fig:overal}.

The server has a list of user embeddings $\{\alpha_j\}_{i=1}^K$ for every user. At the first step, the server randomly selects one user in every round and generates a personalized autoencoder model with its Hyper-network model using user embedding $\alpha_j$\hlbreakable{, Hyper-network parameters $\theta$,} and sends it to the selected user (Eqn. (\ref{equ:hypernetwork})). The Hyper-network implementation is the same as in \cite{shamsian2021personalized}.
\begin{equation}
f_r^{a_j}(.) = h (\alpha_j, \theta)
\label{equ:hypernetwork}
\end{equation}
In the second step, user updates the received autoencoder model using the whole unlabeled and labeled dataset over $T$ epochs and sends the updated autoencoder back to the server. In every epoch, user selects a mini-batch $\gamma \subset U_j \bigcup \{X|(X,Y)\in D_j\}$, and update the received autoencoder via Eqn. (\ref{equ:update_user_first}).
\begin{equation}
{f'}_r^{a_j}(.) \gets {f'}_r^{a_j}(.) - \eta \nabla_{{f'}_r^{a_j}(.)} \mathcal{L}^a_j(\gamma)
\label{equ:update_user_first}
\end{equation}
In the third step, the server updates the Hyper-network parameters \hlbreakable{$\theta$} and the corresponding user embedding feature $\alpha_j$. This means that the server calculates the difference between the sent and the received model parameters and updates the Hyper-network parameter and user embedding as in Eqn. (\ref{equ:update_ae})\cite{shamsian2021personalized}.
\begin{equation}
\begin{matrix}
\Delta f_r^{a_j}(.) \gets {f'}_r^{a_j}(.) - f_r^{a_j}(.)\\\\
\theta \gets \theta - \mu\nabla_{\theta} (f_r^{a_j}(.))^T \Delta f_r^{a_j}(.)\\\\
\alpha_j \gets \alpha_j - \zeta\nabla_{\alpha_j}\theta^T \nabla_{\theta} (f_r^{a_j}(.))^T \Delta f_r^{a_j}(.)
\end{matrix}
\label{equ:update_ae}
\end{equation}
In the fourth step, the server generates a set of $M$ personalized classifiers for user $j$. The server owns a labeled data set from $M (\ll K)$ different distributions. For each of those $M$ datasets, the user first selects samples more similar to the user $j$ dataset. In order to do that, the server checks the distance between its data values and the reconstructed values using user $j$ fine-tuned autoencoder model through calculating an autoencoder loss function (Eqn. (\ref{equ:SampleSelection})):
\begin{figure*}
     \centering
     \begin{subfigure}[b]{0.32\textwidth}
         \centering
         \includegraphics[width=\textwidth]{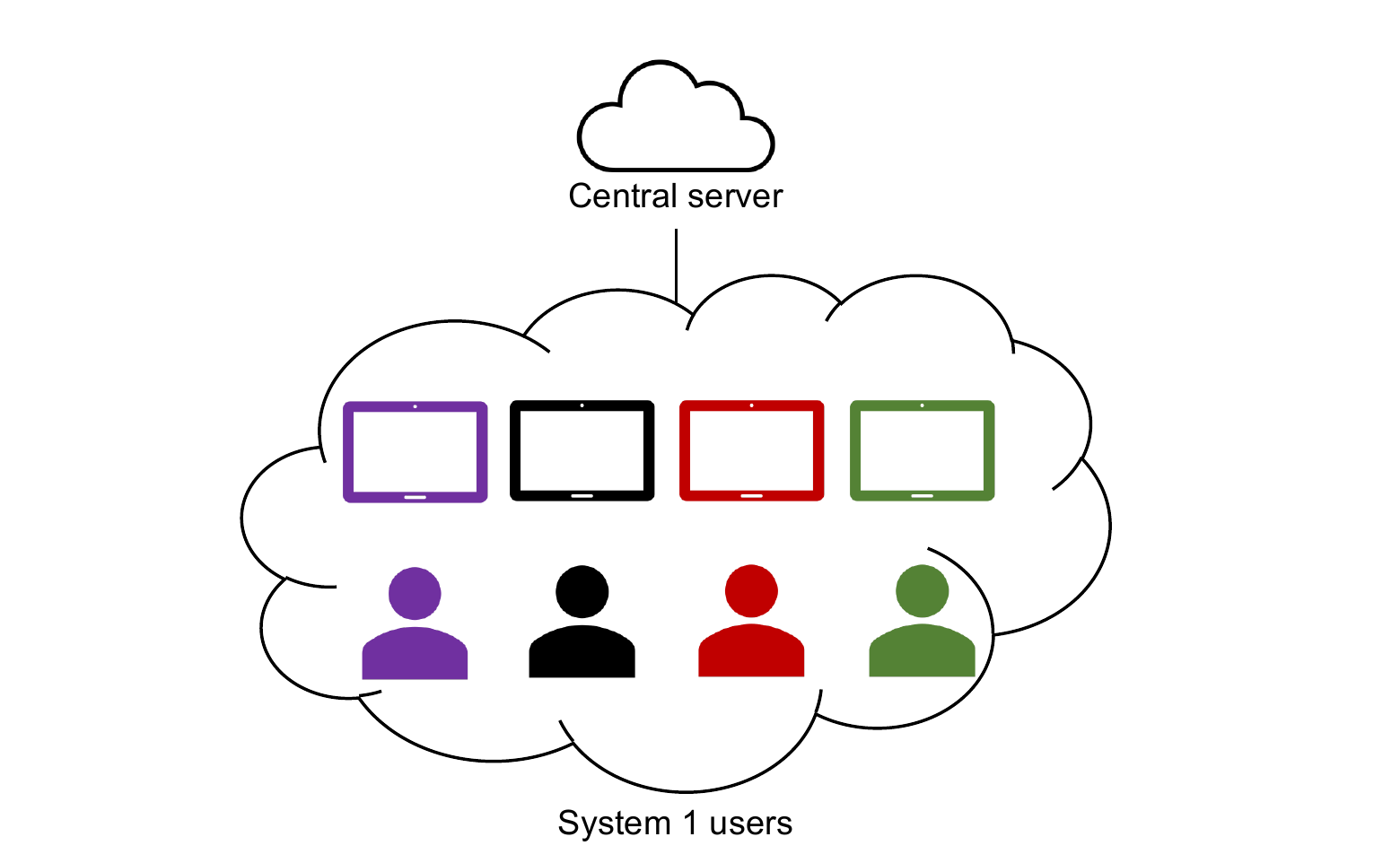}
         \caption{\hlbreakable{Scenario 1}}
     \end{subfigure}
     \hfill
     \begin{subfigure}[b]{0.32\textwidth}
         \centering
         \includegraphics[width=\textwidth]{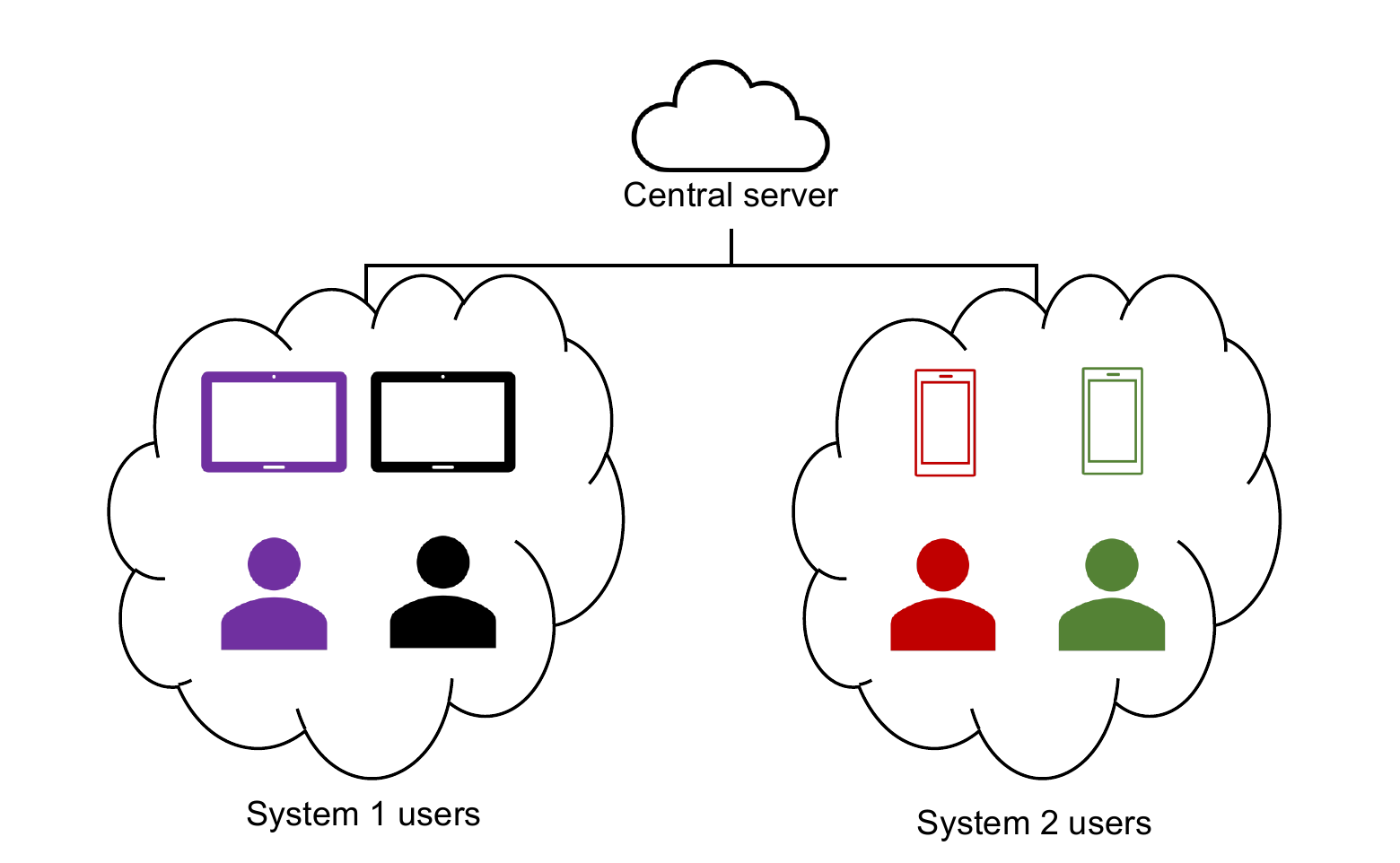}
         \caption{\hlbreakable{Scenario 2}}
     \end{subfigure}
     \hfill
     \begin{subfigure}[b]{0.32\textwidth}
         \centering
         \includegraphics[width=\textwidth]{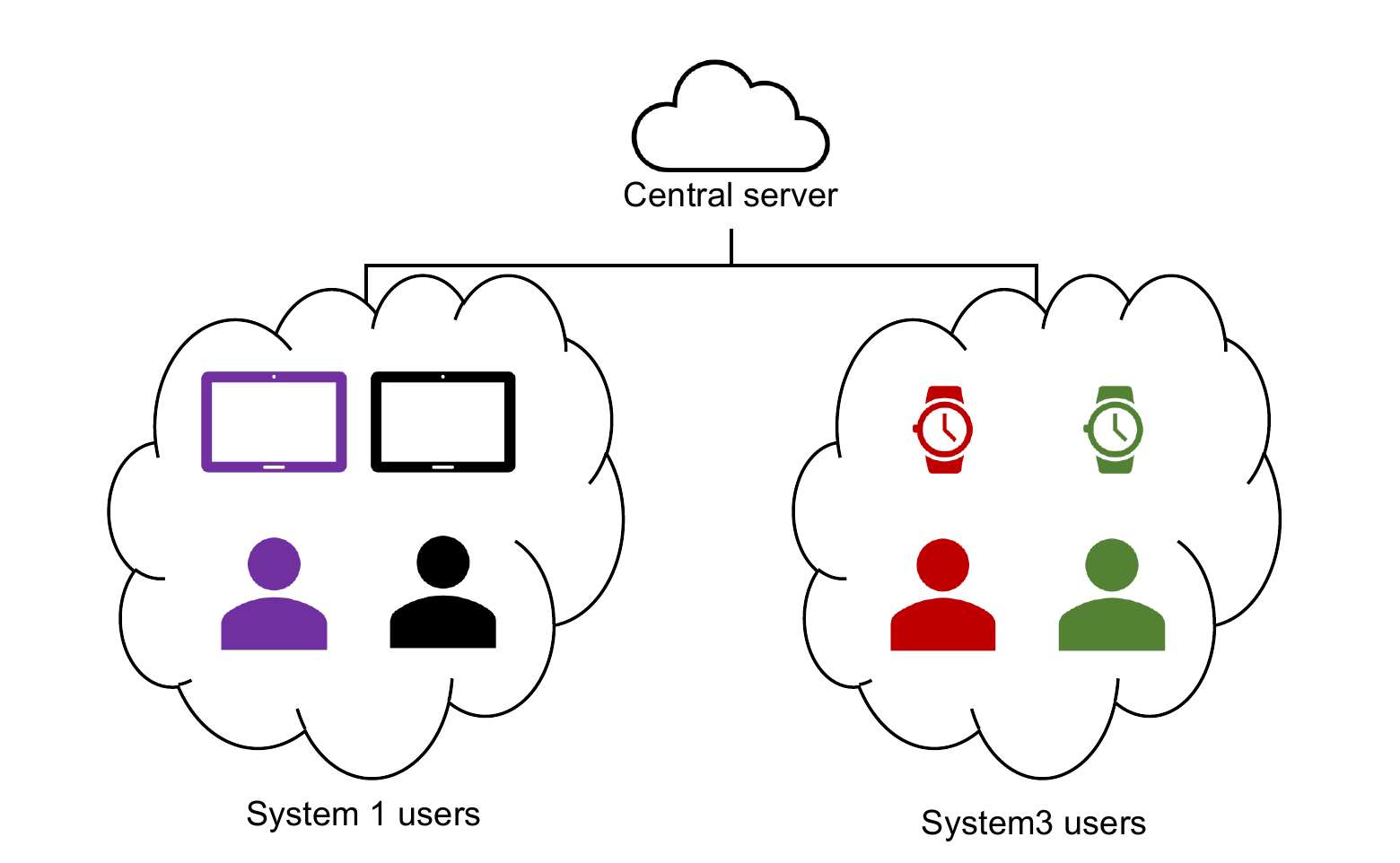}
         \caption{\hlbreakable{Scenario 3}}
     \end{subfigure}     \caption{\hlbreakable{Overview of experimental evaluation scenarios. (a) Scenario 1, in which all users have system 1 hardware resources. (b) Scenario 2, in which half of users have system 1 hardware configurations, while the rest of users have system 2 hardware configurations. (c) Scenario 3, in which half of users have system 1 hardware configurations, while the rest of users have system 3 hardware configurations.}}
      \label{fig:exp}
\end{figure*}
\begin{table*}[t]
\caption{\hlbreakable{Hardware resources comparison between system 1, system 2, and system 3.}}
\begin{center}
{\resizebox{0.7\textwidth}{!}{\color{black}\begin{tabular}{ c | c | c | c }
   \hline
  Feature & System 1 & System 2 & system 3\\
  \hline
  Board & Raspberry Pi 4 Model B - 8GB&Raspberry Pi 4 Model B - 4GB&Raspberry Pi 4 Model B - 2GB\\
   \hline
  CPU & \multicolumn{3}{c}{Quad-core Cortex-A72 (ARM v8) 64-bit SoC 1.5 GHz} \\
  \hline
  RAM & 8GB LPDDR4-3200 SDRAM & 4GB LPDDR4-3200 SDRAM & 2GB LPDDR4-3200 SDRAM\\
  \hline
  Storage & \multicolumn{3}{c}{SanDisk 128GB microSDXC Card}\\
  \hline
  OS & \multicolumn{3}{c}{Raspberrypi OS 64-bit}\\
  \hline
\end{tabular}}}
\end{center}
\label{table:system1and2}
\end{table*}
\begin{equation}
D^j_{S_m}\gets\Big\{(X,Y)|\mathcal{L}^a(X,{f'}_r^{a_j}(X))<\tau,(X,Y)\subset D_{S_m}\Big\} 
\label{equ:SampleSelection}
\end{equation}
In the fifth step, server encodes its samples using the encoder part of user $j$ autoencoder (Eqn. (\ref{equ:servertransform})):
\begin{equation}
{D'}^j_{S_m} \gets \Big\{({f'}_{r,enc}^{a_j}(X),Y)|(X,Y) \in D_{S_m}\Big\}
\label{equ:servertransform}
\end{equation}
Server trains a set of base models using selected samples from each of $M$ available datasets $\{f^{b,j}_{r,m}(.)\}_{m=1}^M$ and send\hlbreakable{s} them to the corresponding user. \hlbreakable{The }user initializes a set of weight\hlbreakable{s} for each of these base models and forms an initial personalized base model \hlbreakable{(Eqn. (\ref{equ:base}))}:
 \begin{equation}
 \begin{matrix}
    \left\{\chi_m\right\}_{m=1}^M = \left\{\frac{1}{M}\right\}_{m=1}^M\\
    \\
    {f'}_r^{b,j}(.) \gets \sum_{m=1}^M \chi_m \cdot f_{r,m}^{b,j}(.)
\end{matrix}
\label{equ:base}
\end{equation}

In the sixth step, The user encodes its labeled instances using its fine-tuned autoencoder that was calculated earlier, freezes base model parameters in Eqn. (\ref{equ:base}), and optimizes model weights via Eqn. (\ref{equ:optimization}) using its labeled dataset:

 \begin{equation}
 \begin{matrix}
    \{\chi_m\}_{m=1}^M=\argmin_{\{\chi_m\}_{m=1}^M} \frac{1}{l_j}\sum_{i=1}^{l_j} \mathcal{L}_j \left( Y_j^{i}, {f'}_r^{b,j}(X_j^{i})\right)\\\\
    \text{Subject to:} \sum_{m=1}^M \chi_m = 1
\end{matrix}
\label{equ:optimization}
\end{equation}

A summary of the SemiPFL algorithm can be found at algorithm \ref{algorithmSemiPFL}.

\begin{table*}[t]
\caption{\textsc{\hlbreakable{Mobiact dataset: Performance evaluation of SemiPFL for the different number of users and number of available labeled data per class at the user side. }}}
\begin{subtable}[ht]{1\textwidth}
\caption{\textsc{\hlbreakable{5 activities - scenario 1}}}
\vspace{-12pt}
\begin{center}
\resizebox{0.90\textwidth}{!}{{\color{black}
\begin{tabular}{c c c c c c c c c c c c} 
\hline

 \multirow{1}{*}{Number of labeled}&

 \multirow{1}{*}{Metric} &
 \multicolumn{10}{c}{\hlbreakable{Number of system 1 users - number of system 2 users - number of system 3 users}}\\
\cline{3-12}
 instances per class&\hlbreakable{Average(Std)}&5-0-0&10-0-0&15-0-0&20-0-0&25-0-0&30-0-0&35-0-0&40-0-0&45-0-0&50-0-0\\
\hline
\multirow{2}{*}{0}&F1 score&92.47(3.06)&92.51(1.87)&92.67(1.37)&93.00(1.05)&93.31(0.81)&93.78(0.62)&94.33(0.46)&94.98(0.29)&95.73(0.21)&96.62(0.09)\\ 
& Kappa&86.49(3.31)&86.59(2.02)&86.72(1.48)&87.00(1.13)&87.34(0.87)&87.76(0.67)&88.33(0.50)&89.00(0.31)&89.74(0.23)&90.69(0.12)\\
\hline
\multirow{2}{*}{10}&F1 score&94.59(2.57)&94.64(1.57)&94.75(1.15)&95.03(0.88)&95.41(0.68)&95.89(0.52)&96.39(0.39)&97.12(0.24)&97.86(0.18)&98.71(0.08)\\ 
& Kappa&88.55(2.96)&88.64(1.80)&88.78(1.32)&89.01(1.01)&89.43(0.78)&89.85(0.60)&90.40(0.45)&91.06(0.28)&91.87(0.21)&92.74(0.11)\\
\hline
\multirow{2}{*}{20}&F1 score&95.22(2.18)&95.28(1.33)&95.38(0.97)&95.65(0.74)&96.04(0.57)&96.52(0.44)&97.02(0.33)&97.75(0.20)&98.49(0.15)&99.34(0.07)\\ 

& Kappa&89.17(2.47)&89.27(1.51)&89.40(1.10)&89.63(0.84)&90.06(0.65)&90.48(0.50)&91.03(0.37)&91.69(0.23)&92.50(0.17)&93.37(0.09)\\
\hline

\end{tabular}}}
\end{center}
\label{table:mobiact5}
\end{subtable}

\begin{subtable}[ht]{1\textwidth}
\caption{\textsc{\hlbreakable{5 activities - scenario 2}}}
\vspace{-12pt}
\begin{center}
\resizebox{0.90\textwidth}{!}{{\color{black}\begin{tabular}{c c c c c c c c c c c c} 
\hline

 \multirow{1}{*}{Number of labeled}&

 \multirow{1}{*}{Metric} &
 \multicolumn{10}{c}{Number of system 1 users - number of system 2 users - number of system 3 users}\\
\cline{3-12}
 instances per class&\hlbreakable{Average(Std)}&3-2-0&5-5-0&8-7-0&10-10-0&13-12-0&15-15-0&18-17-0&20-20-0&23-22-0&25-25-0\\
\hline
\multirow{2}{*}{0}&F1 score&90.33(3.07)&90.57(1.93)&90.73(1.53)&91.12(1.25)&91.52(0.87)&91.63(0.69)&91.77(0.51)&92.86(0.34)&92.96(0.32)&94.60(0.12)\\ 
& Kappa&84.19(3.52)&84.21(2.34)&84.32(1.53)&85.53(1.32)&85.70(0.93)&85.76(0.71)&86.23(0.61)&87.51(0.34)&87.74(0.25)&88.69(0.15)\\
\hline
\multirow{2}{*}{10}&F1 score&93.47(2.63)&93.59(1.64)&93.73(1.23)&94.12(0.91)&94.50(0.72)&94.88(0.53)&95.41(0.41)&96.32(0.30)&96.81(0.21)&97.78(0.11)\\ 
& Kappa&87.05(2.96)&87.12(1.80)&87.34(1.32)&87.61(1.01)&87.92(0.78)&88.23(0.60)&88.29(0.45)&89.06(0.28)&89.27(0.21)&90.74(0.11)\\
\hline
\multirow{2}{*}{20}&F1 score&94.93(2.20)&94.98(1.34)&95.03(0.98)&95.21(0.76)&95.80(0.61)&96.21(0.47)&96.02(0.37)&97.34(0.25)&98.37(0.18)&99.05(0.09)\\ 

& Kappa&88.97(2.47)&89.03(1.51)&89.12(1.10)&89.32(0.84)&89.86(0.65)&90.23(0.50)&90.88(0.37)&91.47(0.23)&92.43(0.17)&93.21(0.09)\\
\hline

\end{tabular}}}
\end{center}
\label{table:mobiact5sc2}
\end{subtable}
\begin{subtable}[ht]{1\textwidth}
\caption{\textsc{\hlbreakable{5 activities - scenario 3}}}
\vspace{-12pt}
\begin{center}
\resizebox{0.90\textwidth}{!}{{\color{black}\begin{tabular}{c c c c c c c c c c c c} 
\hline

 \multirow{1}{*}{Number of labeled}&

 \multirow{1}{*}{Metric} &
 \multicolumn{10}{c}{Number of system 1 users - number of system 2 users - number of system 3 users}\\
\cline{3-12}
 instances per class&\hlbreakable{Average(Std)}&3-0-2&5-0-5&8-0-7&10-0-10&13-0-12&15-0-15&18-0-17&20-0-20&23-0-22&25-0-25\\
\hline
\multirow{2}{*}{0}&F1 score&90.01(3.27)&90.22(1.95)&90.42(1.57)&91.01(1.27)&91.32(0.88)&91.54(0.70)&91.53(0.52)&92.61(0.34)&92.66(0.33)&94.32(0.12)\\ 
& Kappa&83.98(3.52)&84.02(2.34)&84.12(1.53)&85.24(1.32)&85.61(0.93)&85.67(0.71)&86.10(0.61)&87.42(0.34)&87.59(0.25)&88.58(0.15)\\
\hline
\multirow{2}{*}{10}&F1 score&93.24(2.93)&93.33(1.94)&93.69(1.55)&94.01(0.98)&94.34(0.75)&94.53(0.54)&95.24(0.42)&96.13(0.31)&96.66(0.21)&97.52(0.11)\\ 
& Kappa&86.92(3.23)&87.07(3.03)&87.28(1.55)&87.54(1.21)&87.82(0.82)&88.03(0.62)&88.19(0.46)&88.97(0.29)&89.21(0.22)&90.68(0.11)\\
\hline
\multirow{2}{*}{20}&F1 score&94.82(2.30)&94.87(1.38)&94.95(0.99)&95.11(0.77)&95.69(0.61)&96.11(0.47)&95.92(0.37)&97.25(0.25)&98.24(0.19)&98.99(0.11)\\ 

& Kappa&88.89(2.59)&88.97(1.59)&89.01(1.17)&89.27(0.89)&89.82(0.69)&90.15(0.53)&90.78(0.39)&91.41(0.24)&92.38(0.19)&93.12(0.09)\\
\hline

\end{tabular}}}
\end{center}
\label{table:mobiact5sc3}
\end{subtable}
\begin{subtable}[ht]{1\textwidth}

\caption{\textsc{\hlbreakable{10 activities - scenario 1}}}
\vspace{-12pt}
\begin{center}
\resizebox{0.90\textwidth}{!}{{\color{black}\begin{tabular}{c c c c c c c c c c c c} 
\hline

 \multirow{1}{*}{Number of labeled }&

 \multirow{1}{*}{Metric} &
 \multicolumn{10}{c}{\hlbreakable{Number of system 1 users - number of system 2 users - number of system 3 users}}\\
\cline{3-12}
 instances per class&\hlbreakable{Average(Std)}&5-0-0&10-0-0&15-0-0&20-0-0&25-0-0&30-0-0&35-0-0&40-0-0&45-0-0&50-0-0\\
\hline
\multirow{2}{*}{0}&F1 score&88.55(4.14)&88.60(2.52)&88.75(1.85)&89.01(1.41)&89.37(1.09)&89.83(0.84)&90.39(0.62)&91.06(0.39)&91.83(0.29)&92.70(0.15)\\ 
& Kappa&82.49(2.77)&82.51(1.71)&82.67(1.26)&82.95(1.11)&83.34(0.71)&83.79(0.55)&84.30(0.45)&85.06(0.29)&85.73(0.22)&86.67(0.09)\\

\hline
\multirow{2}{*}{10}&F1 score&90.38(3.69)&90.43(2.25)&90.58(1.65)&90.84(1.26)&91.20(0.97)&91.66(0.75)&92.22(0.56)&92.89(0.35)&93.65(0.26)&94.52(0.13)\\ 

& Kappa&84.30(2.70)&84.37(1.47)&84.50(1.13)&84.76(0.83)&85.17(0.67)&85.60(0.60)&86.13(0.40)&86.80(0.23)&87.62(0.17)&88.49(0.08)\\ 
\hline
\multirow{2}{*}{20}&F1 score&90.99(3.09)&91.04(1.88)&91.19(1.38)&91.45(1.05)&91.81(0.81)&92.27(0.62)&92.83(0.47)&93.49(0.29)&94.26(0.21)&95.13(0.11)\\ 
& Kappa&84.91(2.21)&84.99(1.48)&85.11(0.97)&85.37(0.81)&85.78(0.55)&86.21(0.50)&86.73(0.32)&87.41(0.24)&88.23(0.16)&89.09(0.07)\\
\hline

\end{tabular}}}
\end{center}
\label{table:mobiact11}
\end{subtable}

\begin{subtable}[ht]{1\textwidth}

\caption{\textsc{\hlbreakable{10 activities - scenario 2}}}
\vspace{-12pt}
\begin{center}
\resizebox{0.90\textwidth}{!}{{\color{black}\begin{tabular}{c c c c c c c c c c c c} 
\hline

 \multirow{1}{*}{Number of labeled }&

 \multirow{1}{*}{Metric} &
 \multicolumn{10}{c}{Number of system 1 users - number of system 2 users - number of system 3 users}\\
\cline{3-12}
instances per class&\hlbreakable{Average(Std)}&3-2-0&5-5-0&8-7-0&10-10-0&13-12-0&15-15-0&18-17-0&20-20-0&23-22-0&25-25-0\\
\hline
\multirow{2}{*}{0}&F1 score&86.63(4.47)&86.73(2.43)&86.79(1.70)&87.12(1.53)&87.43(1.13)&87.91(0.92)&88.24(0.69)&89.13(0.42)&89.95(0.33)&90.87(0.17)\\ 
& Kappa&80.57(2.84)&80.64(1.78)&80.72(1.31)&83.03(1.17)&81.47(0.79)&81.93(0.61)&82.41(0.52)&83.17(0.37)&83.84(0.29)&84.81(0.11)\\

\hline
\multirow{2}{*}{10}&F1 score&89.43(3.71)&89.54(2.27)&89.61(1.66)&89.72(1.28)&90.11(0.99)&90.57(0.77)&91.31(0.58)&91.92(0.36)&92.70(0.28)&93.49(0.14)\\ 

& Kappa&83.41(2.72)&83.45(1.49)&83.52(1.15)&83.83(0.85)&84.22(0.69)&84.74(0.62)&85.02(0.42)&85.92(0.25)&86.62(0.19)&87.54(0.09)\\ 
\hline
\multirow{2}{*}{20}&F1 score&90.34(3.09)&90.53(1.88)&90.78(1.38)&90.02(1.05)&91.27(0.81)&91.78(0.62)&92.31(0.47)&93.01(0.29)&93.73(0.21)&94.63(0.11)\\ 
& Kappa&84.44(2.21)&84.52(1.48)&84.67(0.97)&84.91(0.81)&85.32(0.55)&85.82(0.50)&86.24(0.32)&87.12(0.24)&88.77(0.16)&88.63(0.07)\\
\hline

\end{tabular}}}
\end{center}
\label{table:mobiact11sc2}
\end{subtable}
\begin{subtable}[ht]{1\textwidth}

\caption{\textsc{\hlbreakable{10 activities - scenario 3}}}
\vspace{-12pt}
\begin{center}
\resizebox{0.90\textwidth}{!}{{\color{black}\begin{tabular}{c c c c c c c c c c c c} 
\hline

 \multirow{1}{*}{Number of labeled }&

 \multirow{1}{*}{Metric} &
 \multicolumn{10}{c}{Number of system 1 users - number of system 2 users - number of system 3 users}\\
\cline{3-12}
instances per class&\hlbreakable{Average(Std)}&3-0-2&5-0-5&8-0-7&10-0-10&13-0-12&15-0-15&18-0-17&20-0-20&23-0-22&25-0-25\\
\hline
\multirow{2}{*}{0}&F1 score&86.32(4.64)&86.44(2.57)&86.51(1.81)&86.86(1.62)&87.12(1.20)&87.62(0.99)&88.01(0.75)&88.89(0.49)&89.64(0.36)&90.71(0.20)\\ 
& Kappa&80.24(2.91)&80.30(1.85)&80.43(1.54)&82.74(1.20)&81.12(0.99)&81.64(0.82)&82.12(0.71)&82.87(0.54)&83.54(0.42)&84.53(0.27)\\

\hline
\multirow{2}{*}{10}&F1 score&89.27(3.80)&89.36(2.35)&89.43(1.71)&89.54(1.35)&89.91(1.09)&90.34(0.92)&91.13(0.62)&91.69(0.40)&92.52(0.36)&93.28(0.23)\\ 

& Kappa&83.20(2.93)&83.26(1.62)&83.33(1.34)&83.67(0.93)&84.01(0.79)&84.52(0.72)&84.82(0.55)&85.76(0.30)&86.41(0.23)&87.31(0.10)\\ 
\hline
\multirow{2}{*}{20}&F1 score&90.26(3.11)&90.42(1.91)&90.66(1.41)&89.92(1.10)&91.13(0.89)&91.70(0.66)&92.22(0.55)&93.92(0.33)&93.64(0.24)&94.54(0.12)\\ 
& Kappa&84.32(2.23)&84.44(1.49)&84.58(0.97)&84.80(0.82)&85.22(0.55)&85.73(0.51)&86.11(0.33)&87.02(0.24)&88.68(0.17)&88.57(0.08)\\
\hline

\end{tabular}}}
\end{center}
\label{table:mobiact11sc3}
\end{subtable}

\label{table:mobiact-tot}
\end{table*}

\begin{table*}[t]
\caption{\textsc{\hlbreakable{WISDM dataset: Performance evaluation of SemiPFL for the different number of users and number of available labeled data per class at the user side.}}}
\vspace{-8pt}
\begin{subtable}[ht]{1\textwidth}
\caption{\textsc{\hlbreakable{scenario 1}}}
\vspace{-12pt}
\begin{center}
\resizebox{0.76\textwidth}{!}{{\color{black}\begin{tabular}{c c c c c c c c c c c} 
\hline
 \multirow{1}{*}{Number of labeled }&

 \multirow{1}{*}{Metric} &
 \multicolumn{9}{c}{\hlbreakable{Number of system 1 users - number of system 2 users - number of system 3 users}}\\
\cline{3-11}
 instances per class&\hlbreakable{Average(Std)}&5-0-0&10-0-0&15-0-0&20-0-0&25-0-0&30-0-0&35-0-0&40-0-0&45-0-0\\
\hline
\multirow{2}{*}{0}&F1 score&83.88(3.17)&83.93(1.65)&84.08(1.30)&84.34(1.02)&84.70(0.87)&85.16(0.56)&85.72(0.44)&86.39(0.28)&87.16(0.19)\\ 
& Kappa&80.58(3.68)&80.66(2.24)&80.84(1.64)&81.07(1.26)&81.47(0.97)&81.92(0.74)&82.41(0.56)&83.17(0.34)&83.93(0.26)\\  
\hline
\multirow{2}{*}{10}&F1 score&86.39(2.73)&86.44(1.56)&86.59(1.11)&86.85(0.85)&87.21(0.71)&87.67(0.54)&88.23(0.38)&88.90(0.25)&89.66(0.17)\\ 
& Kappa&83.15(3.28)&83.17(2.00)&83.32(1.47)&83.61(1.12)&83.95(0.86)&84.37(0.66)&84.94(0.50)&85.63(0.31)&86.40(0.23)\\   
\hline
\multirow{2}{*}{20}&F1 score&87.04(2.61)&87.09(1.37)&87.24(1.12)&87.50(0.73)&87.86(0.67)&88.32(0.48)&88.88(0.33)&89.55(0.22)&90.31(0.16)\\ 
& Kappa&83.80(2.74)&83.82(1.67)&83.97(1.23)&84.26(0.94)&84.60(0.72)&85.02(0.56)&85.59(0.41)&86.28(0.26)&87.05(0.19)\\
\hline
\end{tabular}}}
\end{center}
\label{table:wisdm}
\end{subtable}
\begin{subtable}[ht]{1\textwidth}
\caption{\textsc{\hlbreakable{scenario 2}}}
\vspace{-12pt}
\begin{center}
\resizebox{0.76\textwidth}{!}{{\color{black}\begin{tabular}{c c c c c c c c c c c} 
\hline
 \multirow{1}{*}{Number of labeled }&

 \multirow{1}{*}{Metric} &
 \multicolumn{9}{c}{Number of system 1 users - number of system 2 users - number of system 3 users}\\
\cline{3-11}
 instances per class&\hlbreakable{Average(Std)}&3-2-0&5-5-0&8-7-0&10-10-0&13-12-0&15-15-0&18-17-0&20-20-0&23-22-0\\
\hline
\multirow{2}{*}{0}&F1 score&81.83(3.24)&81.93(1.73)&82.12(1.42)&82.41(1.14)&82.78(0.97)&83.20(0.63)&83.68(0.53)&84.41(0.30)&85.21(0.22)\\ 
& Kappa&78.63(3.76)&78.72(2.31)&78.85(1.69)&79.17(1.32)&79.44(1.03)&77.96(0.79)&80.44(0.61)&82.23(0.39)&81.90(0.30)\\  
\hline
\multirow{2}{*}{10}&F1 score&85.41(2.79)&85.47(1.62)&85.63(1.18)&85.84(0.89)&86.34(0.76)&86.72(0.57)&87.32(0.41)&87.95(0.27)&88.52(0.18)\\ 
& Kappa&82.11(3.34)&82.23(2.08)&82.36(1.53)&82.68(1.17)&83.01(0.89)&83.44(0.69)&83.92(0.54)&84.59(0.34)&85.47(0.25)\\   
\hline
\multirow{2}{*}{20}&F1 score&86.52(2.68)&86.62(1.41)&86.79(1.14)&86.94(0.75)&87.33(0.69)&87.82(0.49)&88.42(0.34)&89.01(0.23)&89.87(0.16)\\ 
& Kappa&83.32(2.76)&83.54(1.69)&83.67(1.25)&83.81(0.96)&84.12(0.74)&84.53(0.58)&85.07(0.42)&85.79(0.27)&86.53(0.19)\\
\hline
\end{tabular}}}
\end{center}
\label{table:wisdmsc2}
\end{subtable}
\begin{subtable}[ht]{1\textwidth}
\caption{\textsc{\hlbreakable{scenario 3}}}
\vspace{-12pt}
\begin{center}
\resizebox{0.76\textwidth}{!}{{\color{black}\begin{tabular}{c c c c c c c c c c c} 
\hline
 \multirow{1}{*}{Number of labeled }&

 \multirow{1}{*}{Metric} &
 \multicolumn{9}{c}{Number of system 1 users - number of system 2 users - number of system 3 users}\\
\cline{3-11}
 instances per class&\hlbreakable{Average(Std)}&3-0-2&5-0-5&8-0-7&10-0-10&13-0-12&15-0-15&18-0-17&20-0-20&23-0-22\\
\hline
\multirow{2}{*}{0}&F1 score&81.41(3.36)&81.61(1.84)&81.82(1.55)&82.13(1.23)&82.34(1.06)&82.92(0.71)&83.34(0.62)&84.12(0.41)&85.01(0.32)\\ 
& Kappa&78.33(3.89)&78.44(2.44)&78.57(1.73)&78.91(1.37)&79.12(1.13)&79.68(0.85)&80.12(0.73)&82.00(0.44)&81.62(0.41)\\  
\hline
\multirow{2}{*}{10}&F1 score&85.22(2.84)&85.28(1.66)&85.42(1.23)&85.66(0.90)&86.12(0.76)&86.54(0.58)&87.13(0.42)&87.77(0.30)&88.35(0.20)\\ 
& Kappa&81.95(3.38)&82.04(2.11)&82.11(1.56)&82.46(1.19)&82.86(0.92)&83.27(0.71)&83.77(0.54)&84.42(0.34)&85.23(0.27)\\   
\hline
\multirow{2}{*}{20}&F1 score&86.41(2.69)&86.53(1.41)&86.67(1.17)&86.83(0.77)&87.20(0.69)&87.71(0.52)&88.32(0.38)&88.83(0.23)&89.74(0.17)\\ 
& Kappa&83.24(2.79)&83.44(1.69)&83.53(1.25)&83.73(0.98)&83.97(0.76)&84.42(0.60)&84.93(0.42)&85.67(0.27)&86.47(0.21)\\
\hline
\end{tabular}}}
\end{center}
\label{table:wisdmsc3}
\end{subtable}
\label{table:wisdm-tot}
\end{table*}

\begin{table*}[t]
\caption{\textsc{\hlbreakable{HAR-UCI dataset: Performance evaluation of SemiPFL for the different number of users and number of available labeled data per class at the user side.}}}
\vspace{-8pt}
\begin{subtable}[ht]{1\textwidth}
\caption{\textsc{\hlbreakable{scenario 1}}}
\vspace{-12pt}
\begin{center}
\resizebox{0.76\textwidth}{!}{{\color{black}\begin{tabular}{c c c c c c c c c c} 
\hline
 \multirow{1}{*}{Number of labeled }&

 \multirow{1}{*}{Metric} &
 \multicolumn{8}{c}{\hlbreakable{Number of system 1 users - number of system 2 users - number of system 3 users}}\\
\cline{3-10}
 instances per class&\hlbreakable{Average(Std)}&3-0-0&6-0-0&9-0-0&12-0-0&15-0-0&18-0-0&21-0-0&24-0-0\\
 \hline
\multirow{2}{*}{0}&F1 score&84.65(3.33)&84.72(1.93)&84.83(1.37)&85.10(1.09)&85.45(0.84)&85.98(0.68)&86.49(0.46)&87.15(0.29)\\ 
& Kappa&81.63(2.80)&81.69(1.71)&81.85(1.29)&82.07(1.12)&82.51(0.72)&82.93(0.59)&83.49(0.45)&84.16(0.30)\\  
\hline
\multirow{2}{*}{10}&F1 score&93.47(2.60)&93.53(1.61)&93.68(1.14)&93.99(0.87)&94.35(0.74)&94.81(0.50)&95.40(0.40)&95.98(0.26)\\ 
& Kappa&90.49(2.71)&90.54(1.69)&90.74(1.23)&91.02(0.83)&91.35(0.64)&91.83(0.53)&92.40(0.43)&92.99(0.26)\\   
\hline
\multirow{2}{*}{20}&F1 score&94.17(2.15)&94.23(1.52)&94.38(1.11)&94.68(0.87)&95.04(0.60)&95.50(0.48)&96.09(0.36)&96.68(0.21)\\ 
& Kappa&91.19(2.38)&91.23(1.35)&91.43(1.10)&91.72(0.86)&92.04(0.60)&92.52(0.54)&93.09(0.34)&93.68(0.25)\\
\hline

\end{tabular}}}
\end{center}
\label{table:uci}
\end{subtable}
\begin{subtable}[ht]{1\textwidth}
\caption{\textsc{\hlbreakable{scenario 2}}}
\vspace{-12pt}
\begin{center}
\resizebox{0.76\textwidth}{!}{{\color{black}\begin{tabular}{c c c c c c c c c c} 
\hline
 \multirow{1}{*}{Number of labeled }&

 \multirow{1}{*}{Metric} &
 \multicolumn{8}{c}{Number of system 1 users - number of system 2 users - number of system 3 users}\\
\cline{3-10}
 instances per class&\hlbreakable{Average(Std)}&2-1-0&3-3-0&5-4-0&6-6-0&8-7-0&9-9-0&11-10-0&12-12-0\\
 \hline
\multirow{2}{*}{0}&F1 score&82.73(3.33)&82.82(1.93)&82.90(1.37)&83.03(1.09)&83.51(0.84)&83.92(0.68)&84.52(0.46)&85.12(0.29)\\ 
& Kappa&79.67(2.80)&79.73(1.71)&79.83(1.29)&80.12(1.12)&80.43(0.72)&80.97(0.59)&81.44(0.45)&82.10(0.30)\\  
\hline
\multirow{2}{*}{10}&F1 score&92.52(2.60)&92.60(1.61)&92.74(1.14)&93.02(0.87)&93.41(0.74)&93.85(0.50)&94.49(0.40)&95.07(0.26)\\ 
& Kappa&89.52(2.71)&89.61(1.69)&89.82(1.23)&89.94(0.83)&90.36(0.64)&90.87(0.53)&91.48(0.43)&92.01(0.26)\\   
\hline
\multirow{2}{*}{20}&F1 score&93.73(2.21)&93.84(1.59)&93.91(1.18)&94.12(0.91)&94.53(0.65)&95.00(0.52)&95.49(0.41)&96.21(0.25)\\ 
& Kappa&90.72(2.50)&90.85(1.49)&90.97(1.15)&91.31(0.90)&91.58(0.60)&92.01(0.59)&92.62(0.38)&93.17(0.29)\\
\hline

\end{tabular}}}
\end{center}
\label{table:ucisc2}
\end{subtable}
\begin{subtable}[ht]{1\textwidth}
\caption{\textsc{\hlbreakable{scenario 3}}}
\vspace{-12pt}
\begin{center}
\resizebox{0.76\textwidth}{!}{{\color{black}\begin{tabular}{c c c c c c c c c c} 
\hline
 \multirow{1}{*}{Number of labeled }&

 \multirow{1}{*}{Metric} &
 \multicolumn{8}{c}{Number of system 1 users - number of system 2 users - number of system 3 users}\\
\cline{3-10}
 instances per class&\hlbreakable{Average(Std)}&2-0-1&3-0-3&5-0-4&6-0-6&8-0-7&9-0-9&11-0-10&12-0-12\\
 \hline
\multirow{2}{*}{0}&F1 score&82.42(3.54)&82.54(2.07)&82.61(1.51)&82.72(1.21)&83.19(0.97)&83.64(0.79)&84.25(0.57)&84.81(0.41)\\ 
& Kappa&79.34(2.89)&79.46(1.82)&79.54(1.41)&79.82(1.33)&80.12(0.83)&80.74(0.71)&81.07(0.51)&81.86(0.41)\\  
\hline
\multirow{2}{*}{10}&F1 score&92.40(2.67)&92.50(1.68)&92.53(1.19)&92.88(0.93)&93.27(0.81)&93.62(0.58)&94.21(0.47)&94.83(0.31)\\ 
& Kappa&89.31(2.79)&89.42(1.73)&89.63(1.29)&89.66(0.89)&90.17(0.69)&90.65(0.59)&91.23(0.49)&91.82(0.29)\\   
\hline
\multirow{2}{*}{20}&F1 score&93.61(2.23)&93.72(1.61)&93.80(1.20)&94.01(0.93)&94.44(0.67)&94.89(0.54)&95.39(0.43)&96.10(0.27)\\ 
& Kappa&90.59(2.53)&90.73(1.51)&90.82(1.17)&91.20(0.92)&91.41(0.62)&91.93(0.61)&92.53(0.41)&93.08(0.31)\\
\hline

\end{tabular}}}
\end{center}
\label{table:ucisc3}
\end{subtable}
\label{table:uci-tot}
\end{table*}

\begin{table*}[t]
\caption{\textsc{\hlbreakable{HHAR dataset: Performance evaluation of SemiPFL for the different number of users and number of available labeled data per class at the user side.}}}
\vspace{-8pt}
\begin{subtable}[ht]{1\textwidth}
\caption{\textsc{\hlbreakable{scenario 1}}}
\vspace{-12pt}
\begin{center}
\resizebox{0.58\textwidth}{!}{{\color{black}\begin{tabular}{c c c c c c c c} 
\hline
 \multirow{1}{*}{Number of labeled }&

 \multirow{1}{*}{Metric} &
 \multicolumn{6}{c}{\hlbreakable{Number of system 1 users - number of system 2 users - number of system 3 users}}\\
\cline{3-8}
instances per class&\hlbreakable{Average(Std)}&1-0-0&2-0-0&3-0-0&4-0-0&5-0-0&6-0-0\\
\hline
\multirow{2}{*}{0}&F1 score&86.18(2.73)&86.23(1.85)&86.38(1.22)&86.64(1.11)&87.00(0.82)&87.46(0.63)\\ 

& Kappa&84.28(3.31)&84.40(1.86)&84.50(1.22)&84.71(0.97)&85.14(0.77)&85.60(0.61)\\  

\hline
\multirow{2}{*}{10}&F1 score&87.54(2.78)&87.59(1.48)&87.74(1.22)&88.00(0.86)&88.36(0.69)&88.82(0.57)\\ 

& Kappa&85.64(2.79)&85.70(1.66)&85.83(1.21)&86.07(0.90)&86.45(0.66)&86.98(0.52)\\   
\hline
\multirow{2}{*}{20}&F1 score&88.09(2.25)&88.15(1.33)&88.30(1.05)&88.55(0.81)&88.91(0.60)&89.37(0.46)\\ 
& Kappa&86.20(2.18)&86.24(1.31)&86.39(1.16)&86.63(0.80)&87.00(0.66)&87.54(0.43)\\   
\hline

\multirow{2}{*}{40}&F1 score&88.59(1.99)&88.64(1.16)&88.79(0.79)&89.05(0.66)&89.41(0.55)&89.87(0.42)\\

& Kappa&86.69(1.92)&86.73(1.26)&86.88(0.84)&87.12(0.65)&87.49(0.49)&88.03(0.43)\\ 
\hline

\end{tabular}}}
\end{center}
\label{table:hhar}
\end{subtable}
\begin{subtable}[ht]{1\textwidth}
\caption{\textsc{\hlbreakable{scenario 2}}}
\vspace{-12pt}
\begin{center}
\resizebox{0.58\textwidth}{!}{{\color{black}\begin{tabular}{c c c c c c c c} 
\hline
 \multirow{1}{*}{Number of labeled }&

 \multirow{1}{*}{Metric} &
 \multicolumn{6}{c}{Number of system 1 users - number of system 2 users - number of system 3 users}\\
\cline{3-8}
instances per class&\hlbreakable{Average(Std)}&1-0-0&1-1-0&2-1-0&2-2-0&3-2-0&3-3-0\\
\hline
\multirow{2}{*}{0}&F1 score&84.21(2.73)&84.33(1.85)&84.49(1.22)&84.72(1.11)&85.12(0.82)&85.52(0.63)\\ 

& Kappa&82.34(3.31)&82.51(1.86)&82.62(1.22)&82.78(0.97)&83.27(0.77)&83.72(0.61)\\  

\hline
\multirow{2}{*}{10}&F1 score&86.57(2.78)&86.62(1.48)&86.81(1.22)&87.08(0.86)&87.42(0.69)&87.93(0.57)\\ 

& Kappa&84.71(2.79)&84.82(1.66)&84.91(1.21)&85.12(0.90)&85.54(0.66)&86.01(0.52)\\   
\hline
\multirow{2}{*}{20}&F1 score&87.53(2.25)&87.69(1.33)&87.86(1.05)&88.03(0.81)&88.43(0.60)&88.80(0.46)\\ 
& Kappa&85.73(2.18)&85.77(1.31)&85.93(1.16)&86.17(0.80)&86.61(0.66)&87.02(0.43)\\   
\hline

\multirow{2}{*}{40}&F1 score&88.53(1.99)&88.59(1.16)&88.71(0.79)&89.01(0.66)&89.38(0.55)&89.82(0.42)\\

& Kappa&86.65(1.92)&86.68(1.26)&86.84(0.84)&87.08(0.65)&87.41(0.49)&87.98(0.43)\\ 
\hline

\end{tabular}}}
\end{center}
\label{table:hharsc2}
\end{subtable}
\begin{subtable}[ht]{1\textwidth}
\caption{\textsc{\hlbreakable{scenario 3}}}
\vspace{-12pt}
\begin{center}
\resizebox{0.58\textwidth}{!}{{\color{black}\begin{tabular}{c c c c c c c c} 
\hline
 \multirow{1}{*}{Number of labeled }&

 \multirow{1}{*}{Metric} &
 \multicolumn{6}{c}{Number of system 1 users - number of system 2 users - number of system 3 users}\\
\cline{3-8}
instances per class&\hlbreakable{Average(Std)}&1-0-0&1-0-1&2-0-1&2-0-2&3-0-2&3-0-3\\
\hline
\multirow{2}{*}{0}&F1 score&83.91(2.81)&84.01(1.95)&84.12(1.41)&84.41(1.30)&84.80(0.93)&85.23(0.77)\\ 

& Kappa&82.01(3.39)&82.22(1.93)&82.33(1.29)&82.49(1.03)&83.00(0.83)&83.41(0.65)\\  

\hline
\multirow{2}{*}{10}&F1 score&86.35(2.81)&86.42(1.50)&86.63(1.23)&86.81(0.87)&87.21(0.70)&87.71(0.58)\\ 

& Kappa&84.52(2.81)&84.63(1.68)&84.73(1.23)&84.97(0.91)&85.41(0.66)&85.82(0.53)\\   
\hline
\multirow{2}{*}{20}&F1 score&87.44(2.25)&87.53(1.33)&87.72(1.06)&87.97(0.82)&88.35(0.60)&88.69(0.46)\\ 
& Kappa&85.62(2.18)&85.69(1.31)&85.81(1.17)&86.03(0.81)&86.51(0.66)&88.91(0.43)\\   
\hline

\multirow{2}{*}{40}&F1 score&88.48(1.99)&88.51(1.16)&88.63(0.80)&88.98(0.67)&89.31(0.55)&89.76(0.42)\\

& Kappa&86.60(1.92)&86.62(1.26)&86.79(0.86)&86.99(0.66)&87.35(0.49)&87.92(0.43)\\ 
\hline

\end{tabular}}}
\end{center}
\label{table:hharsc3}
\end{subtable}
\label{table:hhar-tot}
\end{table*}

\begin{table*}[t]
\caption{\textsc{\hlbreakable{PAMAP2 dataset: Performance evaluation of SemiPFL for the different number of users and number of available labeled data per class at the user side.}}}
\vspace{-8pt}
\begin{subtable}[ht]{1\textwidth}
\caption{\textsc{\hlbreakable{scenario 1}}}
\vspace{-12pt}
\begin{center}
\resizebox{0.58\textwidth}{!}{{\color{black}\begin{tabular}{c c c c c c c c} 
\hline
 \multirow{1}{*}{Number of labeled }&

 \multirow{1}{*}{Metric} &
 \multicolumn{6}{c}{\hlbreakable{Number of system 1 users - number of system 2 users - number of system 3 users}}\\
\cline{3-8}
instances per class&\hlbreakable{Average(Std)}&1-0-0&2-0-0&3-0-0&4-0-0&5-0-0&6-0-0\\
\hline
\multirow{2}{*}{0}&F1 score&88.90(3.23)&88.95(2.01)&89.10(1.31)&89.36(0.96)&89.72(0.76)&90.18(0.57)\\ 
& Kappa&87.57(2.70)&87.65(1.70)&87.81(1.30)&88.06(0.97)&88.45(0.77)&88.90(0.58)\\  
\hline
\multirow{2}{*}{10}&F1 score&90.21(2.98)&90.26(1.50)&90.41(1.13)&90.67(0.86)&91.03(0.68)&91.49(0.51)\\ 
& Kappa&88.94(2.81)&89.00(1.63)&89.13(1.19)&89.41(0.83)&89.75(0.75)&90.22(0.49)\\   
\hline
\multirow{2}{*}{20}&F1 score&90.75(2.11)&90.81(1.47)&90.96(0.99)&91.22(0.82)&91.57(0.64)&92.03(0.49)\\ 

& Kappa&89.50(2.21)&89.55(1.39)&89.68(1.04)&89.96(0.89)&90.30(0.65)&90.77(0.44)\\   
\hline
\multirow{2}{*}{40}&F1 score&91.24(1.87)&91.30(1.33)&91.45(0.96)&91.71(0.73)&92.06(0.47)&92.52(0.36)\\
& Kappa&89.99(2.01)&90.05(1.16)&90.17(0.95)&90.46(0.64)&90.79(0.57)&91.26(0.45)\\ 
\hline

\end{tabular}}}
\end{center}
\label{table:PAMAP2}
\end{subtable}

\begin{subtable}[ht]{1\textwidth}
\caption{\textsc{\hlbreakable{scenario 2}}}
\vspace{-12pt}
\begin{center}
\resizebox{0.58\textwidth}{!}{{\color{black}\begin{tabular}{c c c c c c c c} 
\hline
 \multirow{1}{*}{Number of labeled }&

 \multirow{1}{*}{Metric} &
 \multicolumn{6}{c}{Number of system 1 users - number of system 2 users - number of system 3 users}\\
\cline{3-8}
instances per class&\hlbreakable{Average(Std)}&1-0-0&1-1-0&2-1-0&2-2-0&3-2-0&3-3-0\\
\hline
\multirow{2}{*}{0}&F1 score&86.94(3.23)&86.98(2.01)&87.17(1.31)&87.44(0.96)&87.81(0.76)&88.21(0.57)\\ 
& Kappa&85.62(2.70)&85.72(1.70)&85.92(1.30)&86.12(0.97)&86.53(0.77)&86.98(0.58)\\  
\hline
\multirow{2}{*}{10}&F1 score&89.23(2.98)&89.37(1.50)&89.46(1.13)&89.73(0.86)&90.12(0.68)&91.49(0.51)\\ 
& Kappa&88.02(2.81)&88.15(1.63)&88.20(1.19)&88.49(0.83)&88.81(0.75)&89.32(0.49)\\   
\hline
\multirow{2}{*}{20}&F1 score&90.27(2.11)&90.34(1.47)&90.47(0.99)&90.91(0.82)&91.03(0.64)&91.54(0.49)\\ 

& Kappa&89.02(2.21)&89.12(1.39)&89.23(1.04)&89.49(0.89)&89.79(0.65)&90.24(0.44)\\   
\hline
\multirow{2}{*}{40}&F1 score&91.19(1.87)&91.27(1.33)&91.41(0.96)&91.67(0.73)&92.01(0.47)&92.48(0.36)\\
& Kappa&89.96(2.01)&90.01(1.16)&90.12(0.95)&90.39(0.64)&90.72(0.57)&91.19(0.45)\\ 
\hline

\end{tabular}}}
\end{center}
\label{table:PAMAP2sc2}
\end{subtable}
\begin{subtable}[ht]{1\textwidth}
\caption{\textsc{\hlbreakable{scenario 3}}}
\vspace{-12pt}
\begin{center}
\resizebox{0.58\textwidth}{!}{{\color{black}\begin{tabular}{c c c c c c c c} 
\hline
 \multirow{1}{*}{Number of labeled }&

 \multirow{1}{*}{Metric} &
 \multicolumn{6}{c}{Number of system 1 users - number of system 2 users - number of system 3 users}\\
\cline{3-8}
instances per class&\hlbreakable{Average(Std)}&1-0-0&1-0-1&2-0-1&2-0-2&3-0-2&3-0-3\\
\hline
\multirow{2}{*}{0}&F1 score&86.62(3.54)&86.69(2.33)&86.82(1.52)&87.11(1.02)&87.54(0.86)&88.00(0.63)\\ 
& Kappa&85.31(2.78)&85.42(1.77)&85.66(1.36)&85.88(1.05)&86.24(0.83)&86.70(0.63)\\  
\hline
\multirow{2}{*}{10}&F1 score&89.01(3.00)&89.13(1.55)&89.29(1.16)&89.57(0.89)&89.92(0.70)&91.29(0.55)\\ 
& Kappa&87.81(2.85)&88.01(1.67)&88.05(1.20)&88.23(0.89)&88.63(0.78)&89.13(0.54)\\   
\hline
\multirow{2}{*}{20}&F1 score&90.15(2.15)&90.22(1.49)&90.33(1.02)&90.80(0.84)&90.91(0.66)&91.43(0.50)\\ 

& Kappa&88.91(2.23)&89.01(1.41)&89.11(1.08)&89.38(0.91)&89.68(0.71)&90.13(0.45)\\   
\hline
\multirow{2}{*}{40}&F1 score&91.15(1.89)&91.23(1.35)&91.37(0.98)&91.61(0.75)&91.98(0.47)&92.45(0.36)\\
& Kappa&89.91(2.05)&89.92(1.19)&90.01(0.98)&90.33(0.66)&90.68(0.59)&91.12(0.45)\\ 
\hline

\end{tabular}}}
\end{center}
\label{table:PAMAP2sc3}
\end{subtable}
\label{table:PAMAP2-tot}
\end{table*}

\begin{table*}[t]
\caption{\textsc{\hlbreakable{WESAD dataset: Performance evaluation of SemiPFL for the different number of users and number of available labeled data per class at the user side.}}}
\vspace{-8pt}
\begin{subtable}[ht]{1\textwidth}
\caption{\textsc{\hlbreakable{scenario 1}}}
\vspace{-12pt}
\begin{center}
\resizebox{0.58\textwidth}{!}{{\color{black}\begin{tabular}{c c c c c c c c} 
\hline
 \multirow{1}{*}{Number of labeled }&

 \multirow{1}{*}{Metric} &
 \multicolumn{6}{c}{\hlbreakable{Number of system 1 users - number of system 2 users - number of system 3 users}}\\
\cline{3-8}
instances per class&\hlbreakable{Average(Std)}&2-0-0&4-0-0&6-0-0&8-0-0&10-0-0&12-0-0\\
\hline
\multirow{2}{*}{0}&F1 score&84.71(3.12)&84.81(1.76)&84.88(1.39)&85.16(1.12)&85.55(0.85)&86.01(0.67)\\ 
& Kappa&80.65(3.35)&80.72(2.01)&80.85(1.41)&81.10(1.01)&81.52(0.73)&81.97(0.67)\\  
\hline
\multirow{2}{*}{10}&F1 score&87.72(2.87)&87.72(1.79)&87.86(1.33)&88.18(1.04)&88.55(0.74)&88.73(0.56)\\ 
& Kappa&83.87(2.50)&83.91(1.66)&84.06(1.19)&84.29(0.84)&84.64(0.68)&85.11(0.54)\\   
\hline
\multirow{2}{*}{20}&F1 score&88.39(2.47)&88.38(1.51)&88.53(0.99)&88.84(0.90)&89.22(0.59)&89.68(0.46)\\ 

& Kappa&84.55(2.26)&84.58(1.53)&84.73(0.96)&84.96(0.84)&85.31(0.70)&85.78(0.50)\\   
\hline
\multirow{2}{*}{40}&F1 score&88.94(2.15)&88.93(1.27)&89.08(0.82)&89.39(0.68)&89.77(0.54)&90.23(0.45)\\
& Kappa&85.10(1.80)&85.13(1.15)&85.28(0.88)&85.51(0.63)&85.86(0.57)&86.33(0.38)\\ 
\hline

\end{tabular}}}
\end{center}
\label{table:WESAD}
\end{subtable}
\begin{subtable}[ht]{1\textwidth}
\caption{\textsc{\hlbreakable{scenario 2}}}
\vspace{-12pt}
\begin{center}
\resizebox{0.58\textwidth}{!}{{\color{black}\begin{tabular}{c c c c c c c c} 
\hline
 \multirow{1}{*}{Number of labeled }&

 \multirow{1}{*}{Metric} &
 \multicolumn{6}{c}{Number of system 1 users - number of system 2 users - number of system 3 users}\\
\cline{3-8}
instances per class&\hlbreakable{Average(Std)}&1-1-0&2-2-0&3-3-0&4-4-0&5-5-0&6-6-0\\
\hline
\multirow{2}{*}{0}&F1 score&82.77(3.12)&82.89(1.76)&82.93(1.39)&83.21(1.12)&83.59(0.85)&84.05(0.67)\\ 
& Kappa&78.63(3.35)&78.79(2.01)&79.92(1.41)&79.11(1.01)&79.59(0.73)&80.02(0.67)\\  
\hline
\multirow{2}{*}{10}&F1 score&86.69(2.87)&86.74(1.79)&86.91(1.33)&87.21(1.04)&87.64(0.74)&87.69(0.56)\\ 
& Kappa&82.85(2.50)&82.88(1.66)&83.11(1.19)&83.31(0.84)&83.66(0.68)&84.09(0.54)\\   
\hline
\multirow{2}{*}{20}&F1 score&87.91(2.47)&87.97(1.51)&87.04(0.99)&88.41(0.90)&88.76(0.59)&89.13(0.46)\\ 

& Kappa&84.02(2.26)&84.06(1.53)&84.12(0.96)&84.42(0.84)&84.86(0.70)&85.24(0.50)\\   
\hline
\multirow{2}{*}{40}&F1 score&88.88(2.15)&88.91(1.27)&89.00(0.82)&89.31(0.68)&89.64(0.54)&90.10(0.45)\\
& Kappa&84.98(1.80)&85.02(1.15)&85.19(0.88)&85.39(0.63)&85.79(0.57)&86.28(0.38)\\ 
\hline

\end{tabular}}}
\end{center}
\label{table:WESADsc2}
\end{subtable}
\begin{subtable}[ht]{1\textwidth}
\caption{\textsc{\hlbreakable{scenario 3}}}
\vspace{-12pt}
\begin{center}
\resizebox{0.58\textwidth}{!}{{\color{black}\begin{tabular}{c c c c c c c c} 
\hline
 \multirow{1}{*}{Number of labeled }&

 \multirow{1}{*}{Metric} &
 \multicolumn{6}{c}{Number of system 1 users - number of system 2 users - number of system 3 users}\\
\cline{3-8}
instances per class&\hlbreakable{Average(Std)}&1-0-1&2-0-2&3-0-3&4-0-4&5-0-5&6-0-6\\
\hline
\multirow{2}{*}{0}&F1 score&82.48(3.42)&82.57(1.95)&82.66(1.71)&82.95(1.52)&83.24(1.17)&83.75(0.98)\\ 
& Kappa&78.35(3.52)&78.44(2.51)&79.68(1.82)&78.88(1.51)&79.23(1.03)&79.71(0.98)\\  
\hline
\multirow{2}{*}{10}&F1 score&86.23(3.03)&86.29(1.92)&86.58(1.53)&86.82(1.30)&87.27(0.86)&87.32(0.63)\\ 
& Kappa&82.51(2.78)&82.72(1.83)&82.88(1.49)&82.95(0.94)&83.31(0.79)&83.72(0.63)\\   
\hline
\multirow{2}{*}{20}&F1 score&87.73(2.49)&87.79(1.53)&86.82(1.03)&88.21(0.92)&88.55(0.61)&88.91(0.48)\\ 

& Kappa&83.83(2.29)&83.89(1.58)&83.93(0.99)&84.20(0.88)&84.59(0.75)&85.01(0.54)\\   
\hline
\multirow{2}{*}{40}&F1 score&88.81(2.16)&88.83(1.28)&88.89(0.83)&89.21(0.68)&89.53(0.54)&89.98(0.46)\\
& Kappa&84.91(1.82)&84.92(1.17)&85.09(0.89)&85.25(0.64)&85.69(0.58)&86.17(0.38)\\ 
\hline

\end{tabular}}}
\end{center}
\label{table:WESADsc3}
\end{subtable}
\label{table:WESAD-tot}
\end{table*}

\begin{table*}[t]
\caption{\textsc{\hlbreakable{SleepEDF dataset: Performance evaluation of SemiPFL for the different number of users and number of available labeled data per class at the user side.}}}
\vspace{-8pt}
\begin{subtable}[ht]{1\textwidth}
\caption{\textsc{\hlbreakable{scenario 1}}}
\vspace{-12pt}
\begin{center}
\resizebox{0.58\textwidth}{!}{{\color{black}\begin{tabular}{c c c c c c c c} 
\hline
 \multirow{1}{*}{Number of labeled }&

 \multirow{1}{*}{Metric} &
 \multicolumn{6}{c}{Number of system 1 users - number of system 2 users - number of system 3 users}\\
\cline{3-8}
instances per class&\hlbreakable{Average(Std)}&3-0-0&6-0-0&9-0-0&12-0-0&15-0-0&18-0-0\\
\hline
\multirow{2}{*}{0}&F1 score&61.64(3.16)&61.72(1.87)&61.82(1.48)&62.09(1.01)&62.53(0.86)&62.92(0.65)\\ 
& Kappa&56.62(3.25)&56.71(1.97)&56.84(1.27)&57.14(1.01)&57.45(0.86)&57.94(0.65)\\
\hline
\multirow{2}{*}{10}&F1 score&76.56(2.64)&76.61(1.69)&76.76(1.26)&76.95(0.88)&77.39(0.66)&77.82(0.50)\\
& Kappa&71.54(2.75)&71.57(1.66)&71.77(1.35)&72.03(0.72)&72.35(0.54)&72.81(0.40)\\
\hline
\multirow{2}{*}{20}&F1 score&77.25(2.40)&77.30(1.36)&77.46(1.13)&77.65(0.89)&78.08(0.60)&78.51(0.49)\\ 
& Kappa&72.23(2.29)&72.26(1.29)&72.46(1.03)&72.73(0.89)&73.05(0.68)&73.50(0.47)\\
\hline
\multirow{2}{*}{40}&F1 score&77.81(1.37)&77.86(1.36)&78.02(0.96)&78.21(0.73)&78.64(0.54)&79.07(0.43)\\
& Kappa&72.79(2.03)&72.82(1.48)&73.02(1.03)&73.29(0.81)&73.61(0.53)&74.06(0.46)\\
\hline

\end{tabular}}}
\end{center}
\label{table:SleepEDF}
\end{subtable}
\begin{subtable}[ht]{1\textwidth}
\caption{\textsc{\hlbreakable{scenario 2}}}
\vspace{-12pt}
\begin{center}
\resizebox{0.58\textwidth}{!}{{\color{black}\begin{tabular}{c c c c c c c c} 
\hline
 \multirow{1}{*}{Number of labeled }&

 \multirow{1}{*}{Metric} &
 \multicolumn{6}{c}{Number of system 1 users - number of system 2 users - number of system 3 users}\\
\cline{3-8}
instances per class&\hlbreakable{Average(Std)}&2-1-0&3-3-0&5-4-0&6-6-0&8-7-0&9-9-0\\
\hline
\multirow{2}{*}{0}&F1 score&59.77(3.28)&59.89(1.89)&89.91(1.48)&60.13(1.11)&60.63(0.98)&60.98(0.73)\\ 
& Kappa&54.71(3.11)&54.84(2.07)&54.93(1.34)&55.19(1.13)&54.57(0.98)&56.03(0.79)\\  
\hline
\multirow{2}{*}{10}&F1 score&75.61(2.70)&75.68(1.73)&75.86(1.31)&76.02(0.93)&76.45(0.70)&76.92(0.58)\\ 
& Kappa&70.62(2.84)&70.77(1.71)&70.83(1.41)&71.12(0.79)&71.41(0.60)&71.79(0.51)\\   
\hline
\multirow{2}{*}{20}&F1 score&76.77(2.45)&76.85(1.42)&76.96(1.17)&77.11(0.92)&77.65(0.63)&78.03(0.54)\\ 

& Kappa&71.77(2.33)&71.89(1.34)&71.03(1.11)&72.57(0.94)&72.60(0.73)&72.98(0.52)\\   
\hline
\multirow{2}{*}{40}&F1 score&77.62(1.39)&77.67(1.40)&77.91(1.00)&78.07(0.79)&78.59(0.58)&78.98(0.48)\\
& Kappa&72.77(2.08)&72.71(1.53)&72.97(1.09)&73.17(0.87)&73.52(0.57)&73.99(0.49)\\ 
\hline

\end{tabular}}}
\end{center}
\label{table:SleepEDFsc2}
\end{subtable}
\begin{subtable}[ht]{1\textwidth}
\caption{\textsc{\hlbreakable{scenario 3}}}
\vspace{-12pt}
\begin{center}
\resizebox{0.58\textwidth}{!}{{\color{black}\begin{tabular}{c c c c c c c c} 
\hline
 \multirow{1}{*}{Number of labeled }&

 \multirow{1}{*}{Metric} &
 \multicolumn{6}{c}{Number of system 1 users - number of system 2 users - number of system 3 users}\\
\cline{3-8}
instances per class&\hlbreakable{Average(Std)}&2-0-1&3-0-3&5-0-4&6-0-6&8-0-7&9-0-9\\
\hline
\multirow{2}{*}{0}&F1 score&59.42(3.41)&59.61(1.92)&89.62(1.52)&59.82(1.15)&60.35(1.03)&60.66(0.79)\\ 
& Kappa&54.37(3.19)&54.53(2.15)&54.62(1.41)&54.89(1.19)&54.23(1.03)&55.72(0.84)\\  
\hline
\multirow{2}{*}{10}&F1 score&75.43(2.73)&75.48(1.75)&75.66(1.34)&75.80(0.95)&76.28(0.73)&76.74(0.59)\\ 
& Kappa&70.42(2.86)&70.53(1.73)&70.65(1.43)&70.97(0.81)&71.23(0.62)&71.61(0.53)\\   
\hline
\multirow{2}{*}{20}&F1 score&76.65(2.46)&76.74(1.43)&76.83(1.18)&76.99(0.93)&77.54(0.65)&77.90(0.55)\\ 

& Kappa&71.69(2.35)&71.82(1.37)&70.98(1.14)&72.43(0.96)&72.52(0.77)&72.83(0.54)\\   
\hline
\multirow{2}{*}{40}&F1 score&77.55(1.41)&77.61(1.41)&77.86(1.01)&78.01(0.79)&78.52(0.59)&78.96(0.48)\\
& Kappa&72.72(2.10)&72.67(1.53)&72.91(1.10)&73.12(0.88)&73.48(0.59)&73.92(0.49)\\ 
\hline

\end{tabular}}}
\end{center}
\label{table:SleepEDFsc3}
\end{subtable}
\label{table:SleepEDF-tot}
\end{table*}

\section{Experimental evaluation}
In this section, we evaluate the effectiveness of our method using sets of available activity recognition \rs{and}\hlbreakable{,} stress detection\hlbreakable{, and sleep stage scoring} datasets. A summary of datasets can be found in table \ref{table:datasets}. \rs{SemiPFL was developed using Python 3.8, Pytorch 1.10.2, and trained on a PC with Nvidia 2080Ti 11GB GPU and 96GB RAM.}

\hlbreakable{In order to evaluate SemiPFL in real-time, we designed a set of experiments using Raspberry pi 4 boards. We considered the Raspberry pi as the user communicating over a LAN with a PC as the server. In our experiments, we considered three scenarios: a) all users have system 1 processing power (Fig. \ref{fig:exp} a), b) about half of the users are considered with system 1 processing power and the rest with system 2 hardware specs (Fig. \ref{fig:exp} b), and c) we assume around half of the users have system 1 hardware, and the rest are considered with system 3 processing power (Fig. \ref{fig:exp} c). A comparison between system 1, system 2, and system 3 hardware resources can be found in Table \ref{table:system1and2}. To provide a fair comparison, each type of users receive the same processing time.}

In this chapter, first, we cite datasets used in this study with their corresponding processing. Second, we explain our experimental setup. Third, we compare SemiPFL with other related federated learning frameworks. Fourth, we explain the impact of the number of available labeled instances\rs{. Finally}\hlbreakable{, fifth}, we evaluate the impact of the number of users collaborating during training\hlbreakable{, and finally, we study the effect of user hardware heterogeneity on the overall performance.}

\subsection{Dataset and prepossessing}
We employed five available human action recognition \rs{and}\hlbreakable{,} one stress detection \hlbreakable{, and one sleep stage scoring} datasets to evaluate our method. A summary of datasets can be found in table \ref{table:datasets}. In this study, we used the following datasets:
\subsubsection{Mobiact\cite{10.1007/978-3-319-62704-5_7}} Mobiact includes accelerometer, gyroscope, and orientation data gathered from smartphones in participants' pockets. Twenty activities are recorded, such as standing, walking, jogging, and jumping. In Mobiact, 61 subjects with six trials for each subject are recorded. To compare our results with \st{FehHAR I}\hlbreakable{FedAR} \cite{Bettini2021PersonalizedRecognition} and FedSCN \cite{Saeed2021FederatedIntelligence}, We evaluated our algorithm based on two different sets of outputs, five activities: standing, walking, sitting, jumping, and jogging, and ten activities: standing, car-step in, sitting on a chair, car-step out, jogging, jumping, Stand to sit, stairs down, walking, and stairs up.

\subsubsection{WISDM \cite{weiss2019smartphone}} WISDM dataset contains accelerometer and gyroscope data from smartphones and smartwatches separately, summarizing 18 different activities collected from 51 subjects. In our experiment, to compare our results with \rs{FedHAR I}\hlbreakable{FedAR} \cite{Bettini2021PersonalizedRecognition}, we only considered the following daily activities: walking, jogging, stairs, standing, sitting. 
\hlbreakable{\subsubsection{HAR-UCI \cite{anguita2013public}} HAR-UCI contains activity data collected from thirty participants doing six activities: walking, walking upstairs,  walking downstairs,  sitting,  standing,  and lying while users are wearing a smartphone on the waist, which records 3-axial acceleration and 3-axial angular velocity at a constant rate of 50Hz.}

\subsubsection{HHAR \cite{stisen2015smart}} HHAR is the most widely-used benchmark for human activity recognition algorithms. Data is collected by smartphones and smartwatches separately, including 3-dimensional accelerometer and gyroscope data from 9 users. Six daily activities (biking, sitting, standing, walking, stair up, and stair down) are collected using different devices.

\subsubsection{PAMAP2 \cite{reiss2012introducing,reiss2012creating}} PAMAP2 contains data with 52 dimensions from 3 synchronized IMUs on the chest, dominant side's ankle, and over the wrist on the dominant arm separately. A total of 24 activities are recorded from 8 subjects. We eventually selected ten activities from all eight subjects with 39-dimensional data to address missing data and an imbalance class problem. 

\subsubsection{WESAD \cite{schmidt2018introducing}} WESAD  features physiological data and motion modalities. Fifteen different subjects' data are recorded from wearable sensors on both wrist and chest. WESAD contains three outputs: normal, amusement, and stress.

{\rs{6) HAR-UCI \cite{anguita2013public}} \rs{HAR-UCI contains activity data collected from thirty participants doing six activities: walking, walking upstairs,  walking downstairs,  sitting,  standing,  and lying while users are wearing a smartphone on the waist, which records 3-axial acceleration and 3-axial angular velocity at a constant rate of 50Hz.}
\hlbreakable{\subsubsection{Sleep-EDF \cite{goldberger2000physiobank,stisen2015smart}} Sleep-EDF is a sleep-scoring dataset containing data collected from 20 subjects. It consists of different modalities such as containing Electroencephalogram (EEG), Electrooculogram (EOG), Electromyography (EMG) collected from subjects' chin, event markers, respiration, and body temperature. Data are manually scored by sleep analysis experts to 5 classes (wake, N1, N2, N3,  and rapid  eye  movement). }
\subsection{Experimental setup }
In the past years, many researchers tried to introduce metrics to evaluate performance in federated learning settings\cite{divi2021new}. To evaluate our method, we calculated federated average F1 score and Kappa score for all users (Eqn. (\ref{equ:average})), where $E_j$ is the evaluation dataset for user $j$, and $f_j$ is the personalized model for user j, and $K'$ is the total number of available users during training. We run each scenario \rs{multiple} \hlbreakable{ten} times with different seed values to calculate each performance metric's average and standard deviation values. It is important to note that evaluation data is not used during the training phase. Also, since the server randomly selects users, we eliminated those who did not participate in the training session from the final evaluation.

\begin{equation}
 \begin{matrix}
{F1}_{total}=\sum_{j=1}^{K'} F1(E_j,f_j)\\\\
{Kappa}_{total}=\sum_{j=1}^{K'} Kappa(E_j,f_j)
\label{equ:average}
 \end{matrix}
\end{equation}

We randomly selected a set of users for each dataset to be considered server datasets. We kept the same users for our complete analysis to have consistent results. For Mobiact, HAR-UCI, WISDM, HHAR, PAMAP, \rs{and} WESAD\hlbreakable{, and Sleep-EDF} we selected six, six, six, three, two, \rs{and} three\hlbreakable{, and two} users datasets, respectively, as server instances. We selected our hyperparameters using grid search. For the model architecture, we used four linear layers in the autoencoder (two layers for encoder and two layers for decoder). For the base model, we use two linear layers. After each linear layer, we added a ReLU layer as an activation function and a dropout layer (dropout=$0.2$). 

We use Adam optimizer with the learning rate of $0.001$. The batch size for training was 128. For threshold value, we selected $\tau = 0.05$. We created our data tensor from datasets using a sliding window with length $W=30$. We also normalize and balance class distributions before use. 

For all datasets, we considered 30\% of the data as an evaluation dataset and used the rest for training. We considered 20\% of the remaining data points as labeled datasets and the rest as unlabeled data points. We randomly added \rs{5}, 10, \rs{15}, 20, \rs{25}, \rs{30}, and 40 labeled datapoints per class, respectively, to our setup and started the training from scratch. We reported all the results at \rs{$r=2000$} \hlbreakable{$r=200$} rounds. We also choose $T=T'=10$. For the Hyper-network, we borrowed the structure from pFedHN \cite{shamsian2021personalized} and updated the architecture so that it outputs our autoencoder parameters.

\begin{figure*}
     \centering
     \begin{subfigure}[b]{0.24\textwidth}
         \centering
         \includegraphics[width=\textwidth]{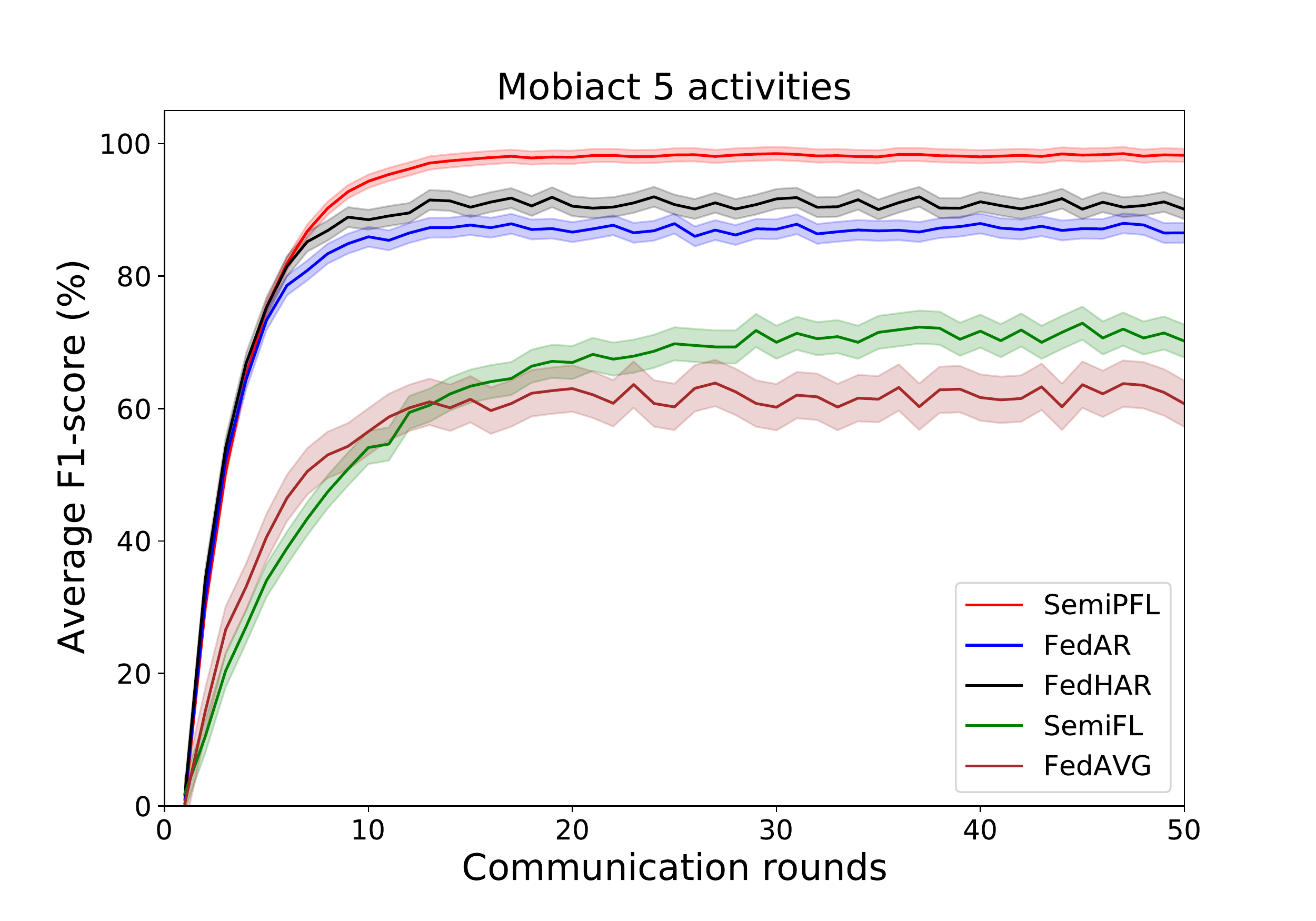}
         \caption{}
     \end{subfigure}
     \hfill
     \begin{subfigure}[b]{0.24\textwidth}
         \centering
         \includegraphics[width=\textwidth]{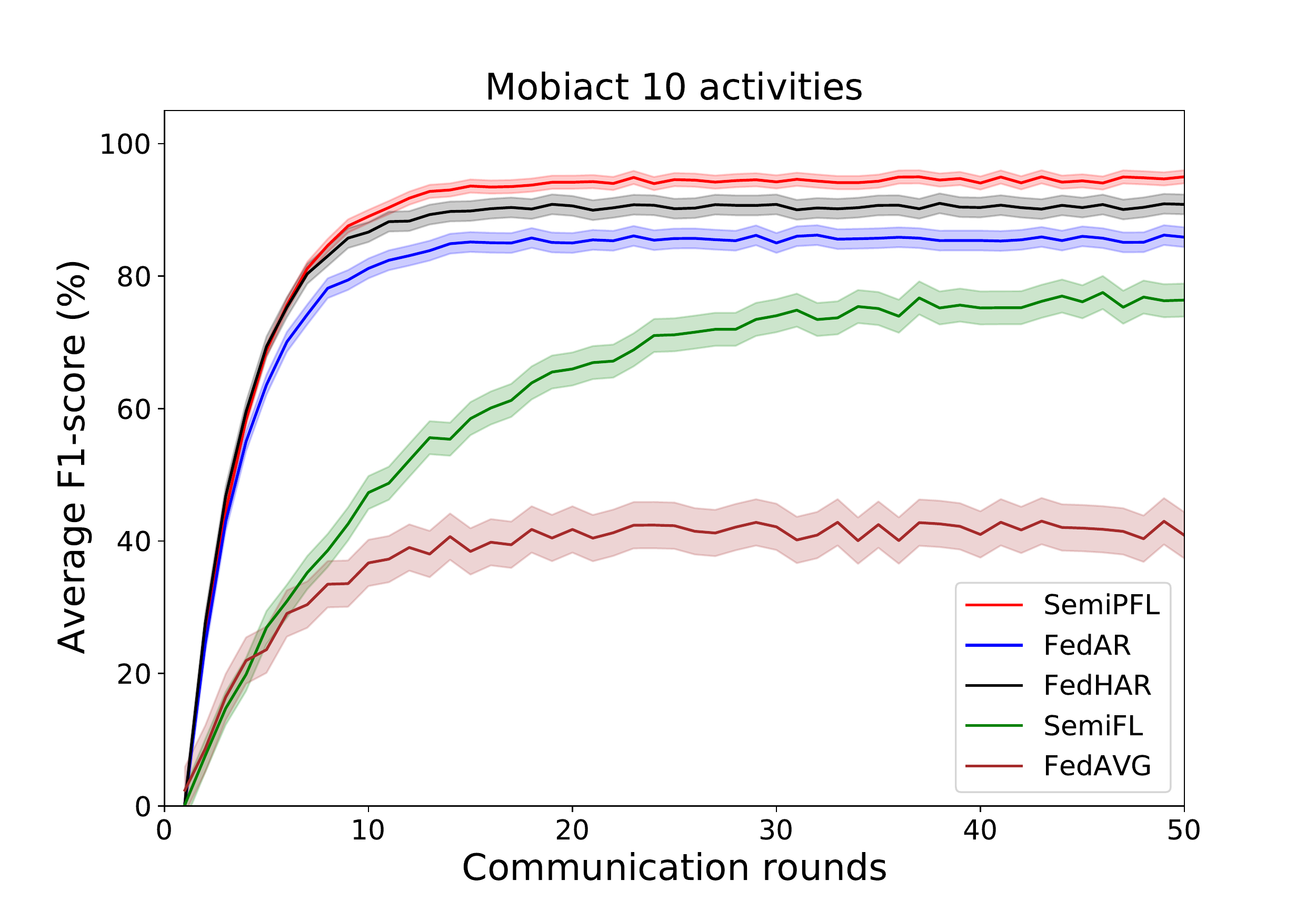}
         \caption{}
     \end{subfigure}
     \hfill
     \begin{subfigure}[b]{0.24\textwidth}
         \centering
         \includegraphics[width=\textwidth]{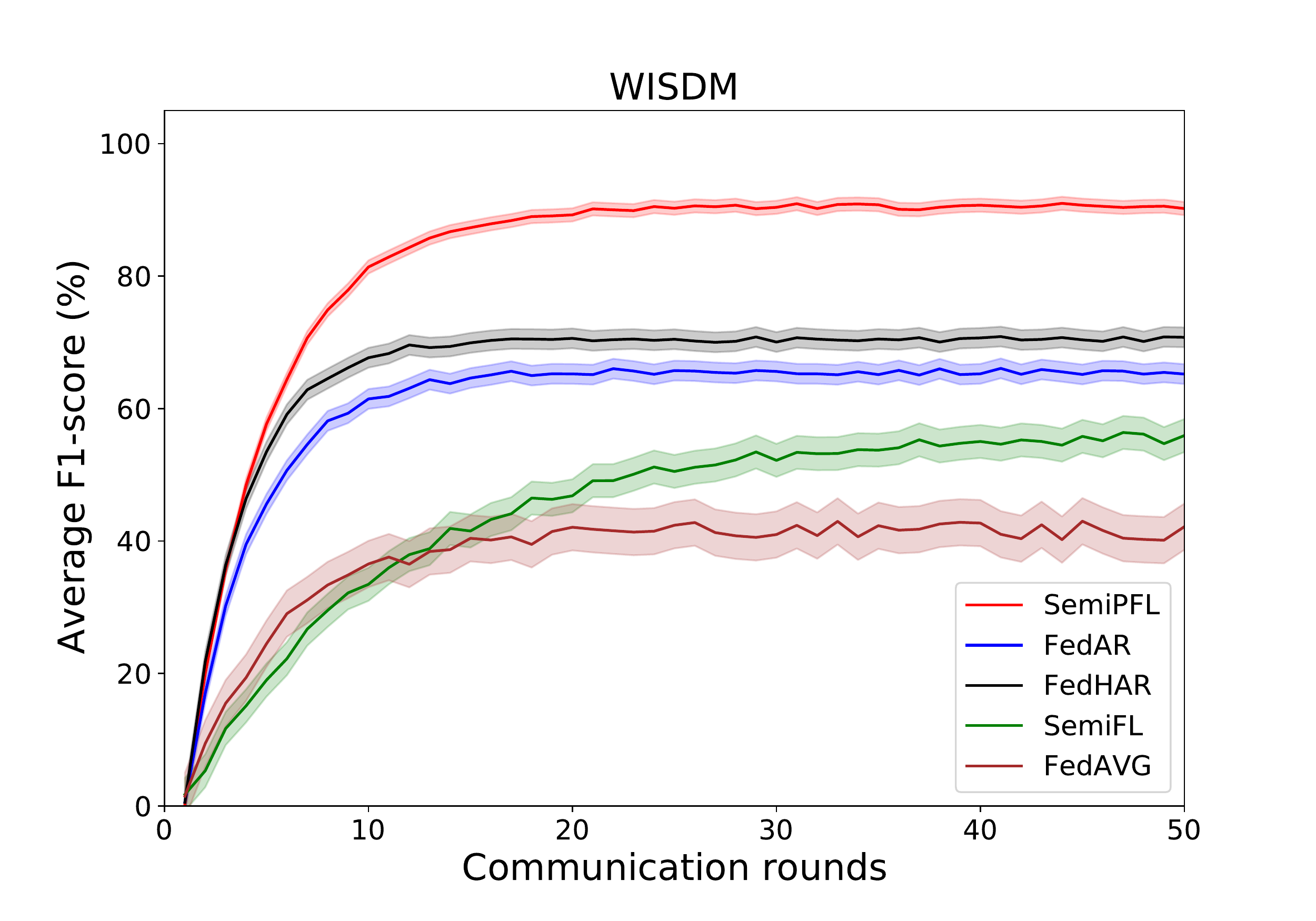}
         \caption{}
     \end{subfigure}
     \hfill
     \begin{subfigure}[b]{0.24\textwidth}
         \centering
         \includegraphics[width=\textwidth]{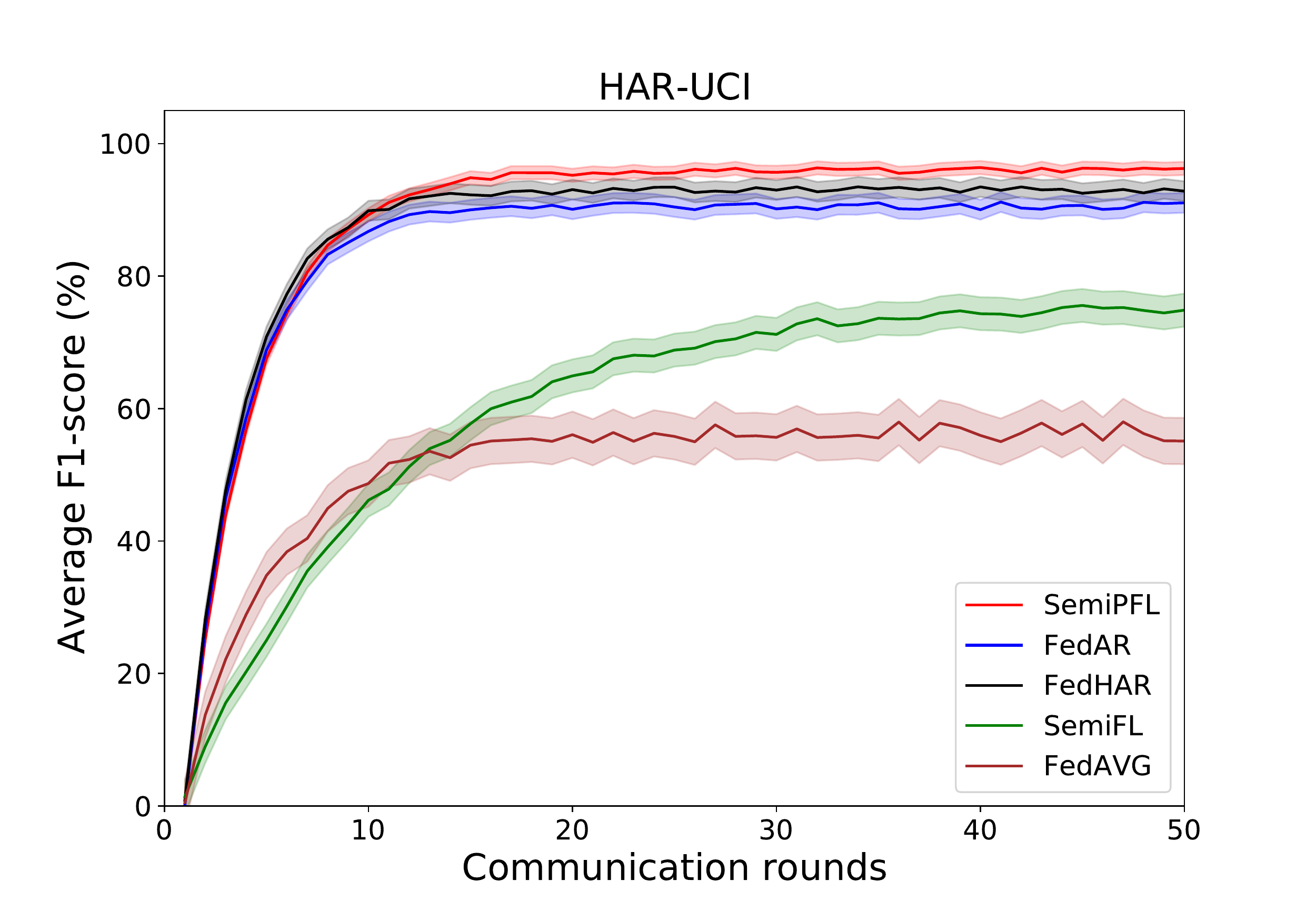}
         \caption{}
     \end{subfigure}
     \hfill
     \begin{subfigure}[b]{0.24\textwidth}
         \centering
         \includegraphics[width=\textwidth]{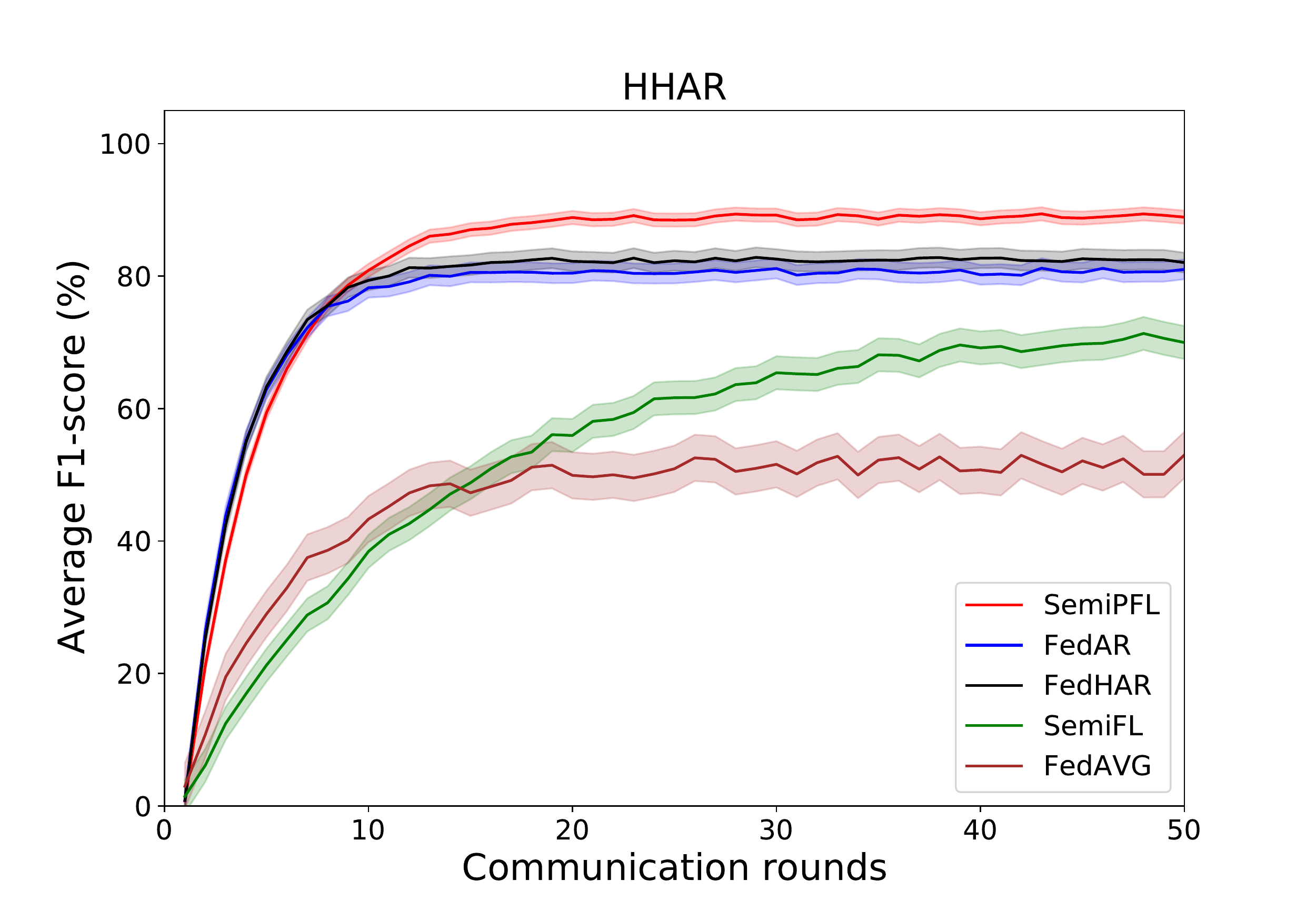}
         \caption{}
     \end{subfigure}
     \hfill
     \begin{subfigure}[b]{0.24\textwidth}
         \centering
         \includegraphics[width=\textwidth]{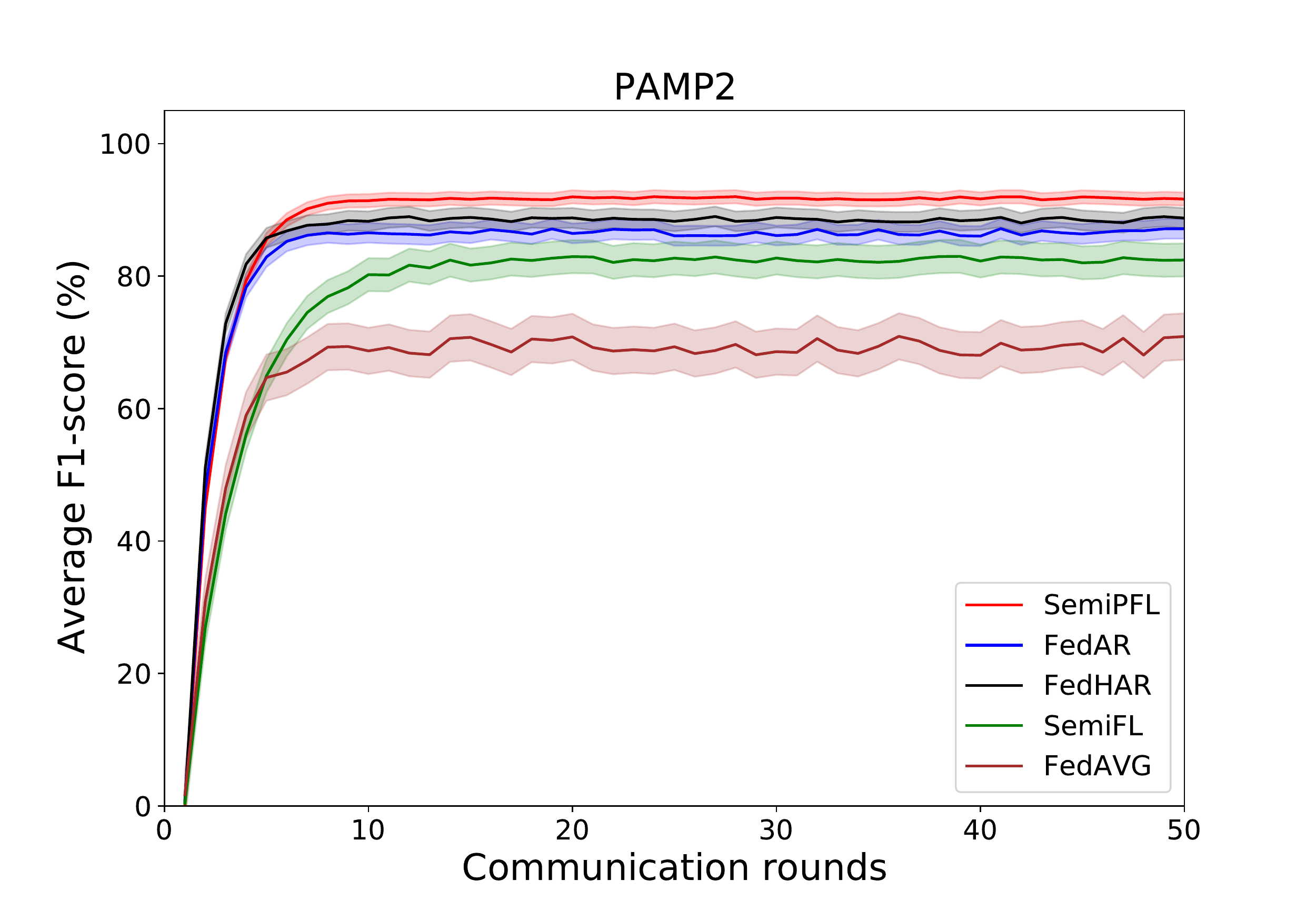}
         \caption{}
     \end{subfigure}
     \hfill
     \begin{subfigure}[b]{0.24\textwidth}
         \centering
         \includegraphics[width=\textwidth]{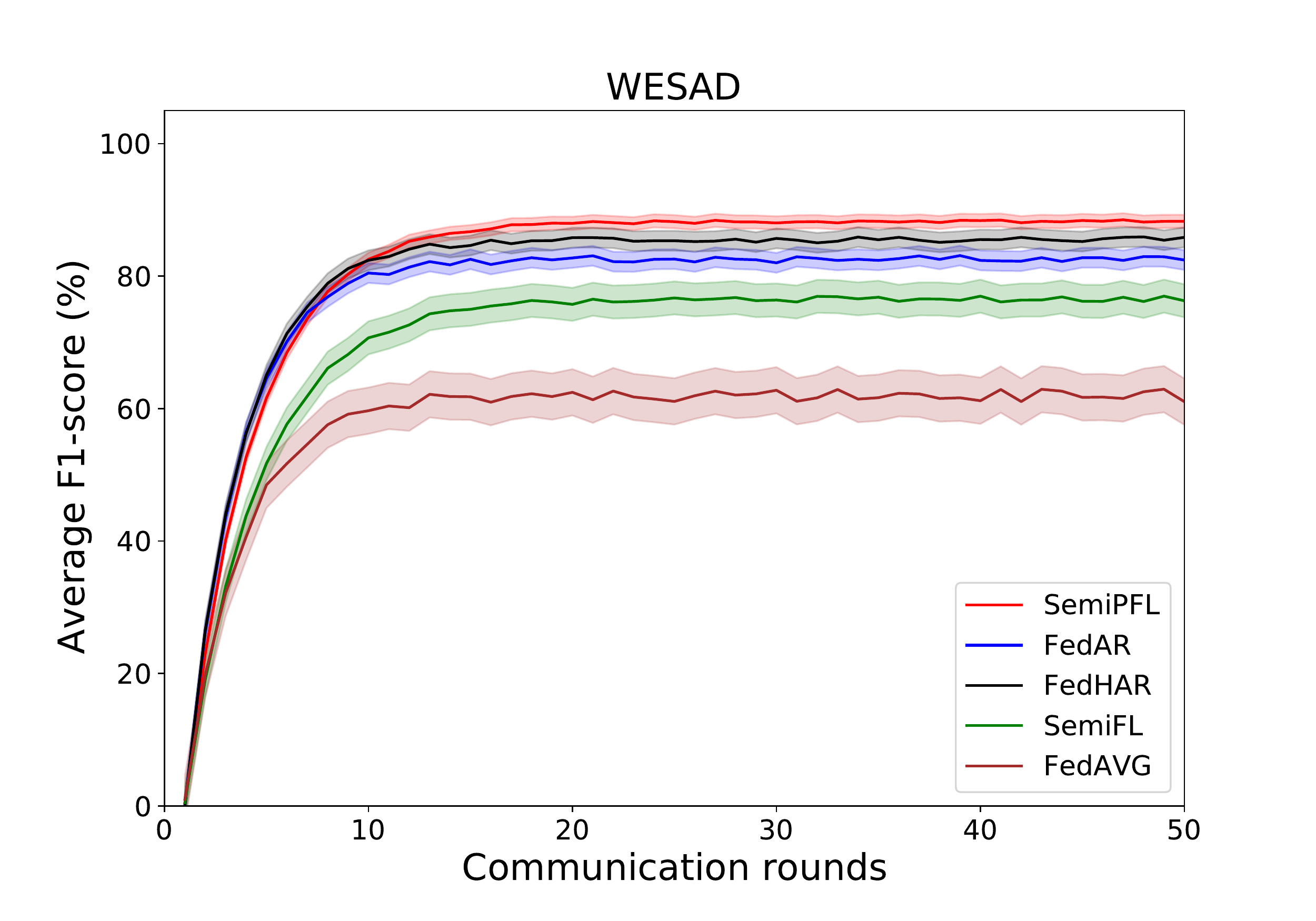}
         \caption{}
     \end{subfigure}
     \hfill
     \begin{subfigure}[b]{0.24\textwidth}
         \centering
         \includegraphics[width=\textwidth]{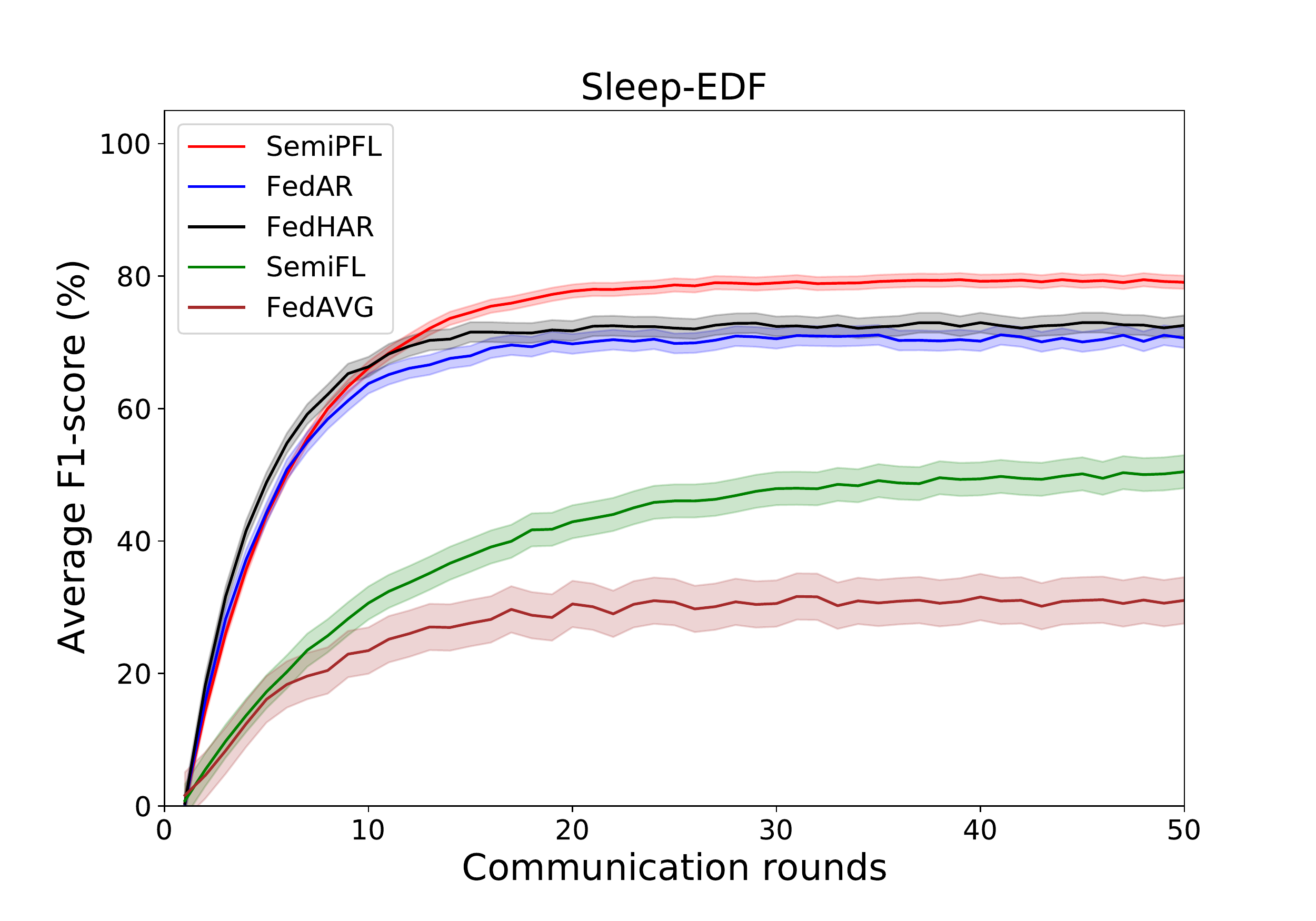}
         \caption{}
     \end{subfigure}
     \caption{\hlbreakable{Comparison of SemiPFL and other methods in terms of the average F1 score over different communication rounds for Mobiact ((a) 5 activities, (b) 10 activities), (c) WISDM, (d) HAR-UCI, (e) HHAR, (f) PAMP2, (g) WESAD, and (h) Sleep-EDF datasets.}}
      \label{fig:com}
\end{figure*}

\begin{figure*}
         \centering
         \includegraphics[width=0.8\textwidth]{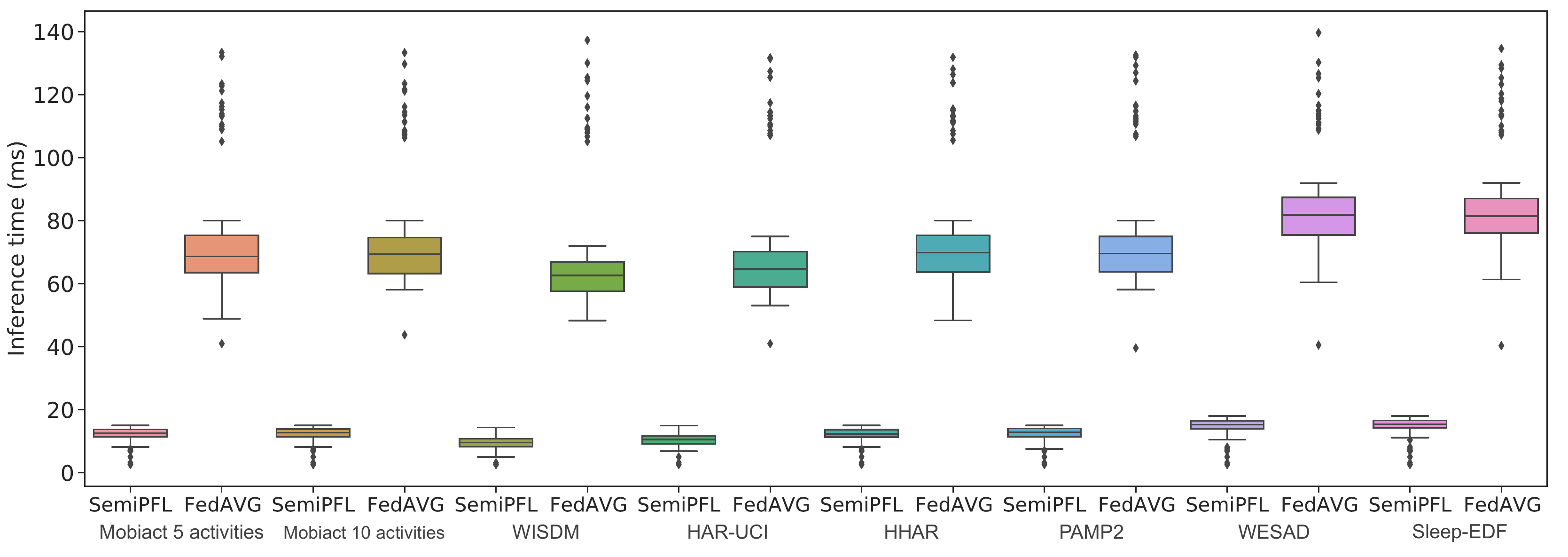}
         \caption{\hlbreakable{Comparison of SemiPFL and FedAVG in terms of inference time for Mobiact (5 activities, 10 activities), WISDM, HAR-UCI, HHAR, PAMP2, WESAD, and Sleep-EDF datasets.}}
         \label{fig:inference}
\end{figure*}
\begin{figure*}
     \centering
     \begin{subfigure}[b]{0.24\textwidth}
         \centering
         \includegraphics[width=\textwidth]{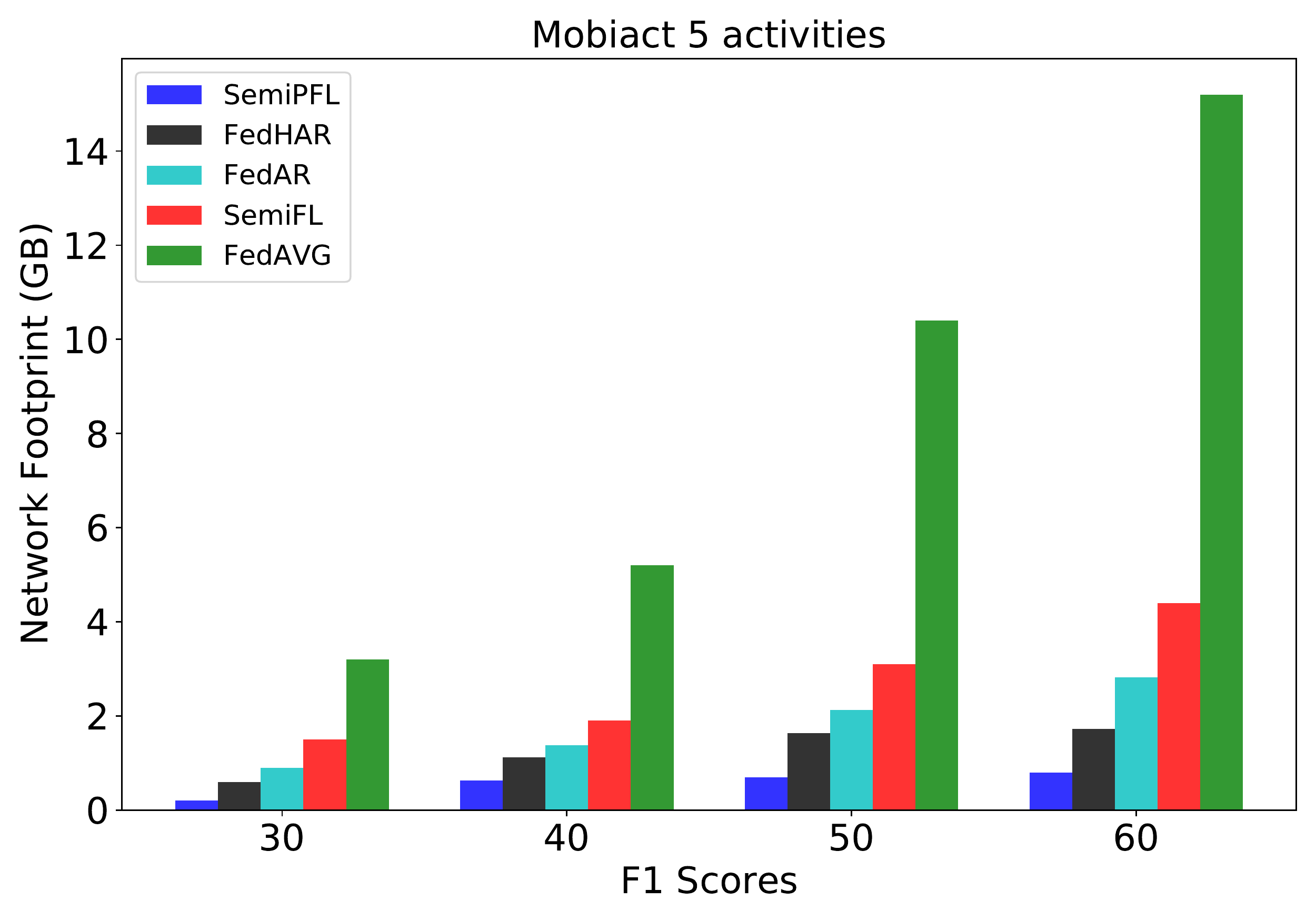}
         \caption{}
     \end{subfigure}
     \hfill
     \begin{subfigure}[b]{0.24\textwidth}
         \centering
         \includegraphics[width=\textwidth]{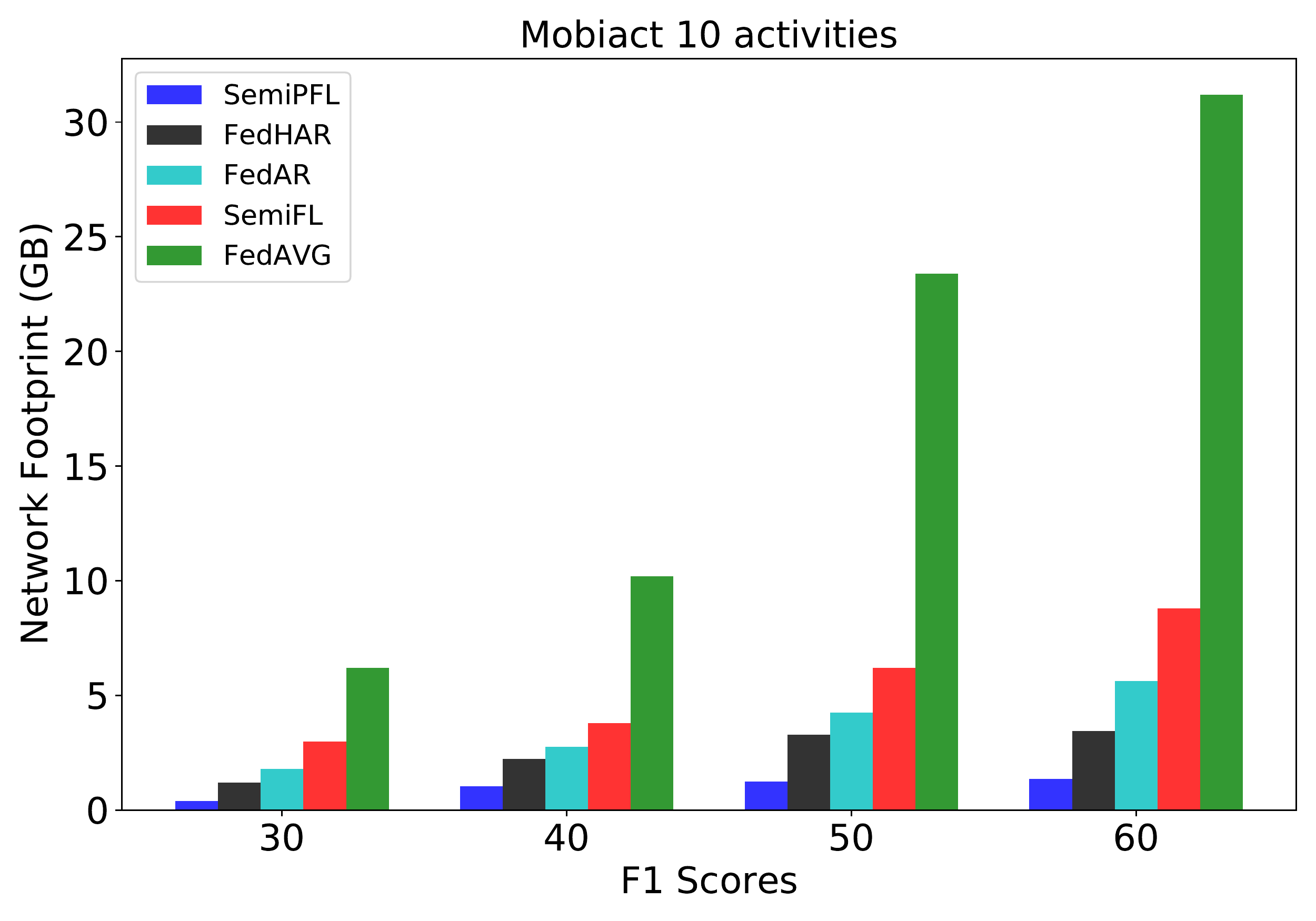}
         \caption{}
     \end{subfigure}
     \hfill
     \begin{subfigure}[b]{0.24\textwidth}
         \centering
         \includegraphics[width=\textwidth]{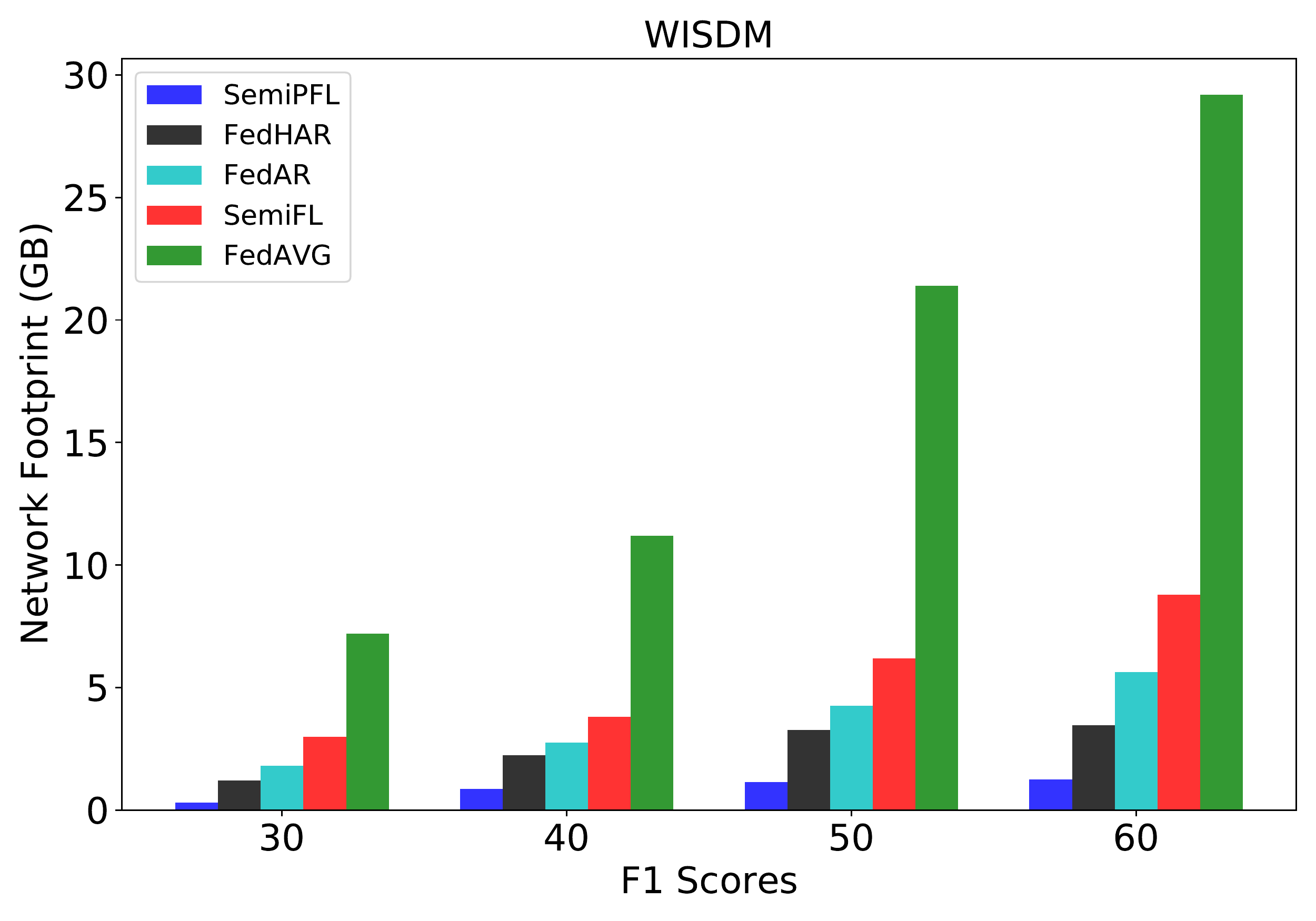}
         \caption{}
     \end{subfigure}
     \begin{subfigure}[b]{0.24\textwidth}
         \centering
         \includegraphics[width=\textwidth]{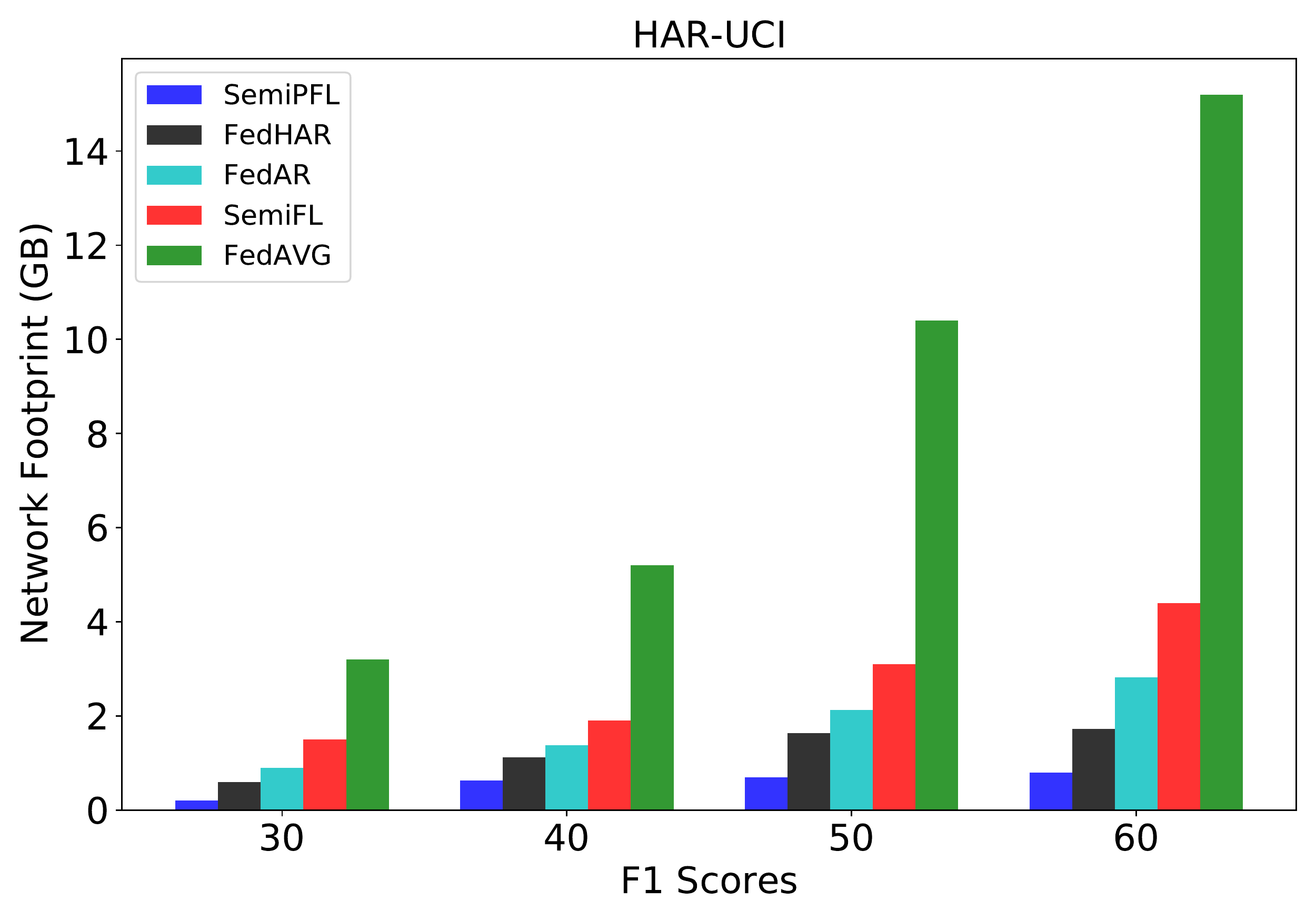}
         \caption{}
     \end{subfigure}
     \hfill
     \begin{subfigure}[b]{0.24\textwidth}
         \centering
         \includegraphics[width=\textwidth]{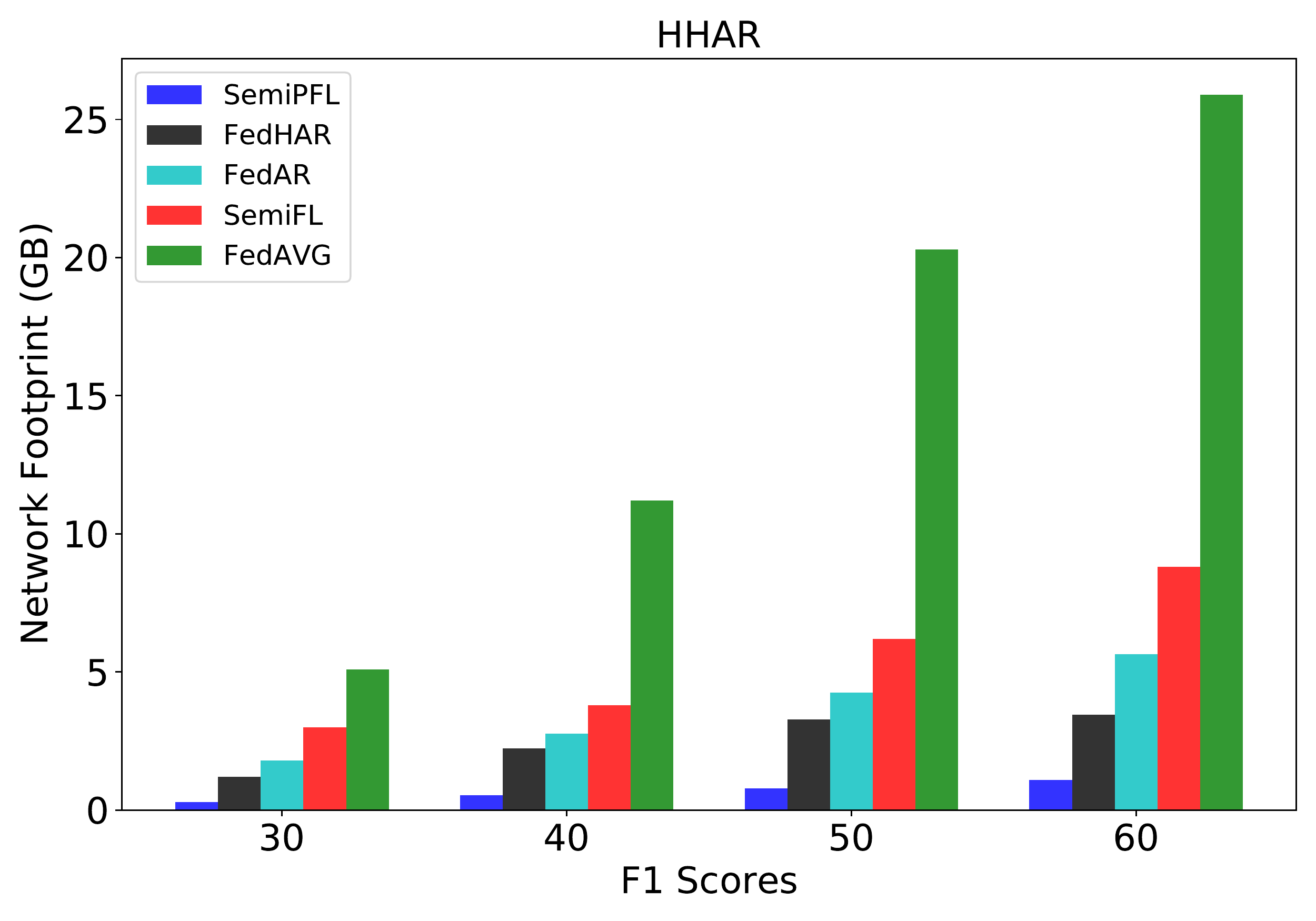}
         \caption{}
     \end{subfigure}
     \hfill
     \begin{subfigure}[b]{0.24\textwidth}
         \centering
         \includegraphics[width=\textwidth]{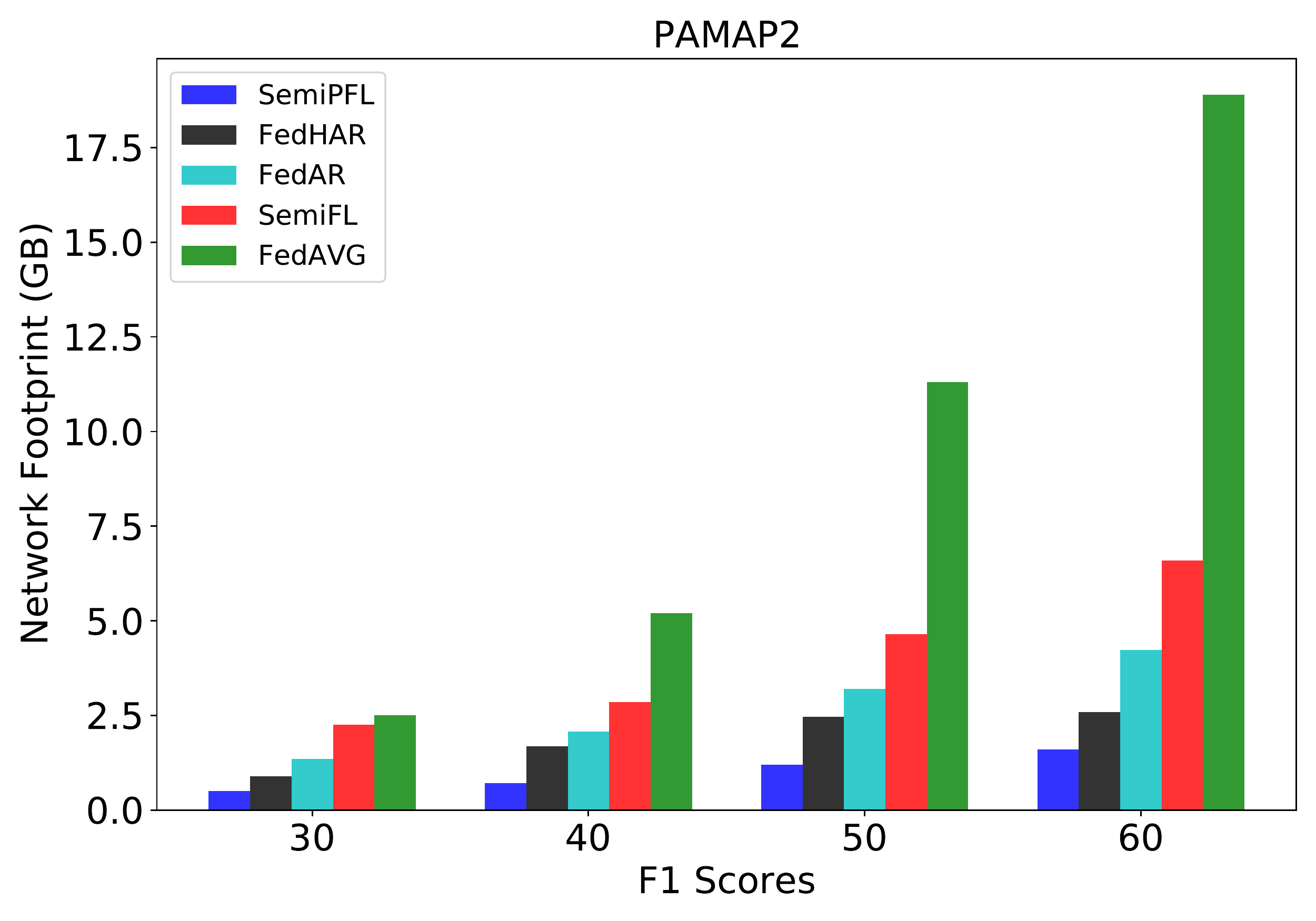}
         \caption{}
     \end{subfigure}
     \hfill
     \begin{subfigure}[b]{0.24\textwidth}
         \centering
         \includegraphics[width=\textwidth]{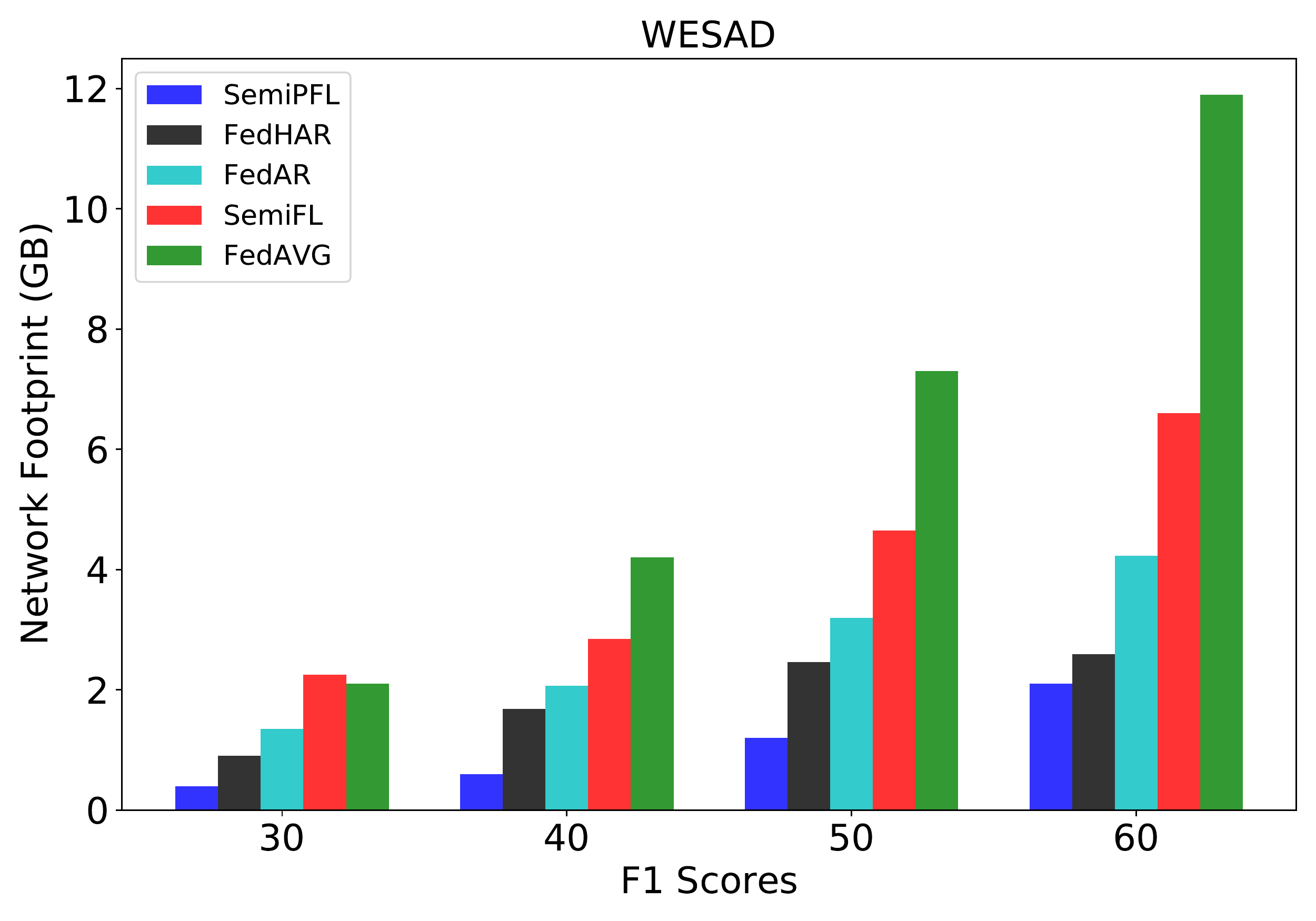}
         \caption{}
     \end{subfigure}
     \hfill
     \begin{subfigure}[b]{0.24\textwidth}
         \centering
         \includegraphics[width=\textwidth]{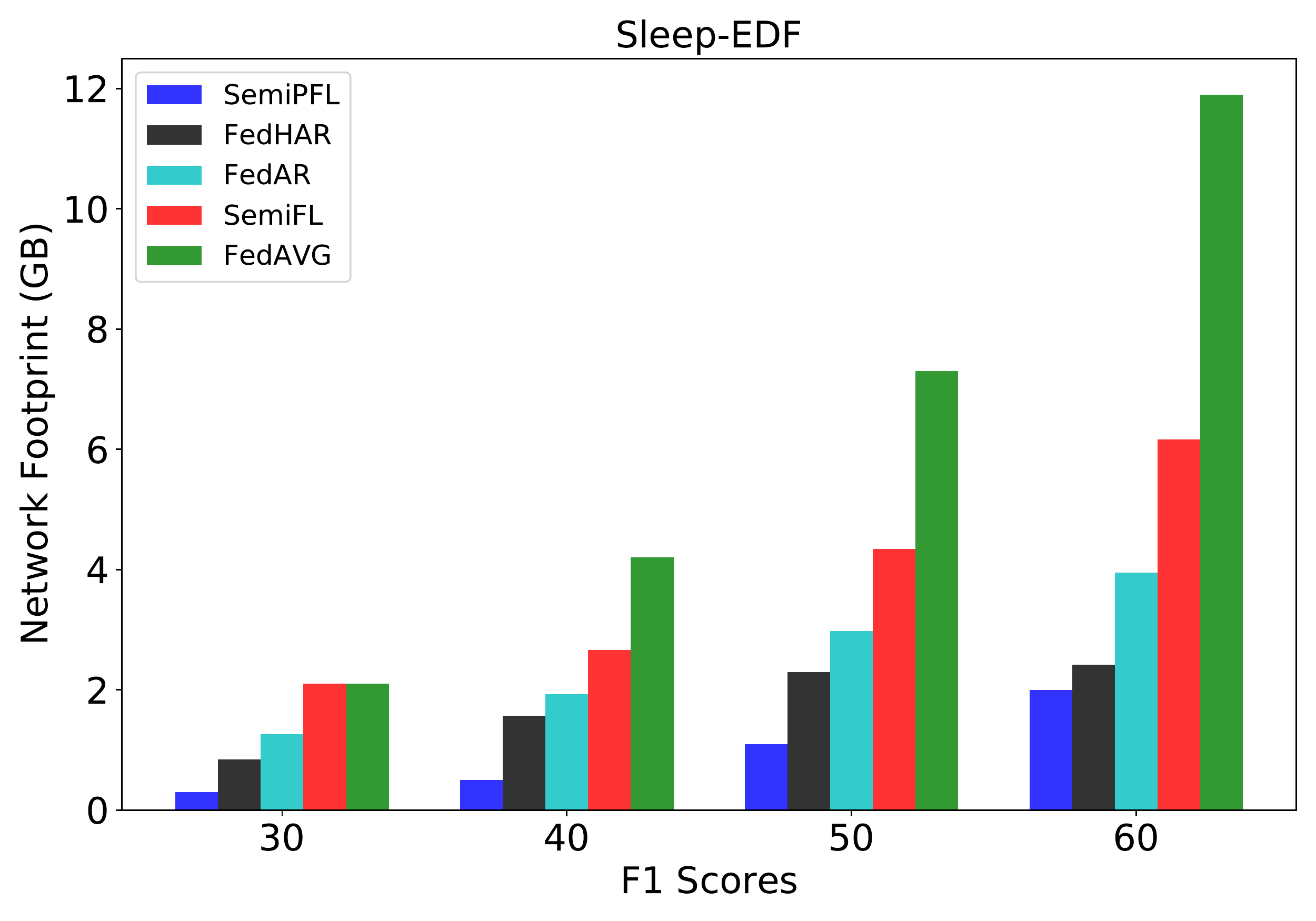}
         \caption{}
     \end{subfigure}
     \caption{\hlbreakable{Comparison of the network footprint between SemiPFL, FedAR, FedHAR, SemiFL, and FedAVG for Mobiact ((a) 5 activities, (b) 10 activities), (c) WISDM, (d) HAR-UCI, (e) HHAR, (f) PAMP2, (g) WESAD, and (h) Sleep-EDF datasets.}}
\label{fig:compare}
\end{figure*}

\subsection{Comparison with other methods}
\subsubsection{\hlbreakable{Comparison in terms of overall performance}}
Table \ref{table:litreview} demonstrates a comparison amongst our method and the most related studies investigating the effectiveness of federated learning frameworks for embedded edge intelligence. Here, we have classified these methods based on their objectives: personlized, label in server, label in user, and the model type. In order to be able to accurately compare our results with selected methods, we chose similar pre-processing steps. 

Most of these works focus on federated learning, where users collaborate to generate a global model. These methods are not performing well, having lower average Kappa or F1 scores (as seen in Table \ref{table:litreview}). While SemiPFL use FCNN which is simpler and faster to train than the more complex architectures such as CNN and LSTM, it outperforms them all in terms of respective average F1 and Kappa values. \rs{SemiPFL method with its attributes such as some label in server, and no requirement for label for user addresses broader application scenarios as well as providing better performance metrics as compared to both recent semi-personalized work \mbox{\cite{arivazhagan2019federated,li2020federated,li2021fedbn,yu2021fedhar,Bettini2021PersonalizedRecognition,Chen2021FedHealthHealthcare}}, and earlier federated learning methods \mbox{\cite{zhao2020semi,mcmahan2017communicationefficient,Saeed2021FederatedIntelligence}}.} \hlbreakable{SemiPFL performs better than recent semi-personalized work \mbox{\cite{arivazhagan2019federated,li2020federated,li2021fedbn,yu2021fedhar,Bettini2021PersonalizedRecognition,Chen2021FedHealthHealthcare}}, and earlier federated learning methods \mbox{\cite{zhao2020semi,mcmahan2017communicationefficient,Saeed2021FederatedIntelligence}}. It also has a more relaxed attribute regarding the need for labeled data on the user side.} 
\subsubsection{\hlbreakable{Comparison in terms of convergence speed}}
\hlbreakable{Fig. \ref{fig:com} demonstrates a comparison of SemiPFL with other methods in average F1-score over communication rounds, using Mobiact 5 activities (Fig. \ref{fig:com} a), Mobiact 10 activities (Fig. \ref{fig:com} b), WISDM (Fig. \ref{fig:com} c), HAR-UCI (Fig. \ref{fig:com} d), HHAR (Fig. \ref{fig:com} e), PAMP2 (Fig. \ref{fig:com} f), WESAD (Fig. \ref{fig:com} g), and Sleep-EDF (Fig. \ref{fig:com} h) datasets. The center lines in plots correspond to average values, and the shaded area represents the standard deviation, calculated over 10 different trials. In all datasets, FedAR (blue line) and FedHAR (black line) converge slightly faster than SemiPFL (red line) initially. However, SemiPFL in the same context converges to a higher average F1-score. This can result from having one user communicating with the central server at a time. In both FedAR and FedHAR, every round, a set of users collaborate with the central server. SemiFL (green line) and FedAVG (brown line) have slower or lower convergence values compared to SemiPFL, FedAR, and FedHAR in the average F1-score.}

\subsubsection{\hlbreakable{Comparison in terms of inference time}}

\hlbreakable{Fig. \ref{fig:inference} reports a comparison between SemiPFL and FedAVG regarding inference time tested on Raspberry Pi 4 with system 1 specs using Mobiact 5 activities, Mobiact 10 activities, WISDM, HAR-UCI, HHAR, PAMP2, WESAD, and Sleep-EDF datasets (A summary can be found in Fig. \ref{fig:inference}). Similar to \cite{Zhao2020Semi-supervisedRecognition}, which uses autoencoder to train a global model for users, the processing time of SemiPFL is significantly lower ($p<0.001$) than FedAVG. The reason is that in SemiPFL, users first encode their dataset into smaller representations. Despite the fact that the autoencoder adds up to the user's processing time, the encoder transforms the user dataset into a smaller representation, which requires smaller models compared to methods such as FedAVG.}
\subsubsection{\hlbreakable{Comparison in terms of network footprint}}
\hlbreakable{Similar to \cite{luping2019cmfl}, we have compared the network footprint between SemiPFL, FedAR \cite{Bettini2021PersonalizedRecognition}, FedHAR \cite{yu2021fedhar}, SemiFL \cite{Zhao2020Semi-supervisedRecognition}, and FedAVG\cite{mcmahan2017communicationefficient}, based on the average F1-score for all users using Mobiact 5 activities (Fig. \ref{fig:compare} a), Mobiact 10 activities (Fig. \ref{fig:compare} b), WISDM (Fig. \ref{fig:compare} c), HAR-UCI (Fig. \ref{fig:compare} d), HHAR (Fig. \ref{fig:compare} e), PAMP2 (Fig. \ref{fig:compare} f), WESAD (Fig. \ref{fig:compare} g), and Sleep-EDF (Fig. \ref{fig:compare} h) datasets. Based on the results, SemiPFL (blue bar) has less network footprint than FedAR (cyan bar), FedHAR (black bar), SemiFL (red bar), and FedAVG (green bar) in terms of the same average F1 score. One of the main reasons is that in our method, only one user communicates with the server every communication round, making transmission cost more efficient.} 
\subsection{Impact of number of available labeled instances}
SemiPFL covers a wide range of scenarios from no labeled data to fully supervised setting, unlike other previous publications. This design has lent itself to investigate the effect of increased available labeling by incrementally adding to the labeled instances per class and measuring the improvements in average Kappa and F1 scores as shown in Tables \ref{table:mobiact5}, \ref{table:mobiact11}, \hlbreakable{\ref{table:wisdm}}, \hlbreakable{\ref{table:uci}, } \ref{table:hhar}, \ref{table:PAMAP2}, \ref{table:WESAD}\hlbreakable{, and \ref{table:SleepEDF}}. \hlbreakable{We also see the same trend in \ref{table:mobiact5sc2}, \ref{table:mobiact5sc3}, \ref{table:mobiact11sc2}, \ref{table:mobiact11sc3}, \ref{table:wisdmsc2},
\ref{table:wisdmsc3}, \ref{table:ucisc2},
\ref{table:ucisc3}, \ref{table:hharsc2}, \ref{table:hharsc3}, \ref{table:PAMAP2sc2}, \ref{table:PAMAP2sc3}, \ref{table:WESADsc2}, \ref{table:WESADsc3}, \ref{table:SleepEDFsc2}, and \ref{table:SleepEDFsc3}, where hardware heterogeneity exists among half of users.} Having some labeled datasets on the server side in the SemiPFL design enables us to achieve high scores even without any user labels and increasing the performance as the number of labeled data increases. This trend has been observed in this study for different \rs{activity numbers and }datasets, while in some instances it highlights the possibility to have an optimum number of labeled instances as the performance metrics saturate beyond certain number of labeled datasets.

\subsection{Impact of number of users collaborating during training}

To investigate the impact of the number of users collaborating during the training process, we randomly selected five users for Mobiact and WISDM, and then added five more users each time and started training SemiPFL from scratch. For PAMAP2 and HHAR, we randomly selected one user, and added one more user in each training and testing of SemiPFL. For HAR-UCI \rs{dataset}\hlbreakable{ and Sleep-EDF datasets}, we randomly selected three users, and then added three more users every time. For WESAD, we randomly selected two users, and then added two more users every time. The results for the performance metrics average Kappa and F1 scores as the different number of users are added to SemiPFL is also reported in Tables \ref{table:mobiact5}, \ref{table:mobiact11}, \hlbreakable{\ref{table:wisdm}}, \hlbreakable{\ref{table:uci}, } \ref{table:hhar}, \ref{table:PAMAP2}, \rs{\mbox{\ref{table:wisdm}}},\ref{table:WESAD}, and \ref{table:SleepEDF}. \hlbreakable{We also see the same trend in \ref{table:mobiact5sc2}, \ref{table:mobiact5sc3}, \ref{table:mobiact11sc2}, \ref{table:mobiact11sc3}, \ref{table:wisdmsc2},
\ref{table:wisdmsc3}, \ref{table:ucisc2},
\ref{table:ucisc3}, \ref{table:hharsc2}, \ref{table:hharsc3}, \ref{table:PAMAP2sc2}, \ref{table:PAMAP2sc3}, \ref{table:WESADsc2}, \ref{table:WESADsc3}, \ref{table:SleepEDFsc2}, and \ref{table:SleepEDFsc3}, where hardware heterogeneity exists among half of users.} Unlike previous publications, we observe that due to the sophisticated personalized representation of SemiPFL adding more users increase the performance scores and not necessarily decrease the performance due to the edge data heterogeneity. This is an important outcome as it demonstrates the possibility for collaborative learning from edge nodes for the entire SemiPFL model. 
\subsection{\hlbreakable{Impact of user hardware resource heterogeneity}}
\hlbreakable{Tables \ref{table:mobiact-tot}, \ref{table:wisdm-tot}, \ref{table:uci-tot}, \ref{table:hhar-tot}, \ref{table:PAMAP2-tot}, \ref{table:WESAD-tot}, and \ref{table:SleepEDF-tot} present our experimental results for scenarios 1, 2, and 3. Tables \ref{table:mobiact5}, \ref{table:mobiact11}, \ref{table:wisdm}, \ref{table:uci},  \ref{table:hhar}, \ref{table:PAMAP2}, \ref{table:WESAD},  and \ref{table:SleepEDF} present the overall performance in terms of the F1-score and Kappa score for scenario 1, where all users have the same processing specs.  Tables \ref{table:mobiact5sc2}, \ref{table:mobiact11sc2}, \ref{table:wisdmsc2},
\ref{table:ucisc2},
\ref{table:hharsc2}, \ref{table:PAMAP2sc2}, \ref{table:WESADsc2}, and \ref{table:SleepEDFsc2}, demonstrate the effect of increasing labeled data for users, and the effect of increasing active users during training when we have scenario 2 user hardware heterogeneity. Tables \ref{table:mobiact5sc3}, 
\ref{table:mobiact11sc3}, 
\ref{table:wisdmsc3}, 
\ref{table:ucisc3}, 
\ref{table:hharsc3}, 
\ref{table:PAMAP2sc3}, 
\ref{table:WESADsc3}, 
and \ref{table:SleepEDFsc3}, show the overall performance of scenario 3, when the number of labeled data increases, and the number of active users increases. One visible trend is that scenario 1 performance is generally always higher than scenario 2, and scenario 2 is always higher than scenario 3 in terms of the average F1-score and Kappa score. However, based on our results, all three scenarios converge to nearly the same values in the reported performance measure as the number of users increases and the number of labeled data increases. We can conclude that our method shows good performance stability when hardware resource heterogeneity exists in scenarios 2, and 3. To our knowledge, this work is the first attempt to quantify such effects, and it is of interest to explore further study of such variations.}

\section{Conclusion}
This paper introduces SemiPFL, a novel personalized semi-supervised learning method focusing on edge intelligence. Our approach trains a Hyper-network that generates a personalized autoencoder to enable learning from user data representation. Furthermore, based on the user autoencoder and the sets of the available dataset in server-side from different distributions, the server generates a group of base models to the corresponding user. Finally, the user fine-tunes the weighted average of such base models to generate a personalized base model. We extensively evaluated the proposed method \hlbreakable{in three different real-time experimental scenarios }on five publicly available human action recognition\rs{ and}\hlbreakable{,} one stress detection\hlbreakable{, and one sleep-stage scoring} datasets collected from wearable devices. Our method outperforms personalized models and available federated learning frameworks under the same assumptions in terms of average F1 and Kappa scores. \hlbreakable{We also compare SemiPFL and other related methods in terms of convergence speed, inference time, and network footprint. In all three categories, SemiPFL demonstrates superior results compared to related methods. We also studied the effect of user hardware heterogeneity in three scenarios and demonstrated stable performance in the presence of such variations. However, this work is the first attempt to quantify such effects and unlock new avenues for further study of such variations.} While SemiPFL can perform well for cases without labeled data at the edge, the performance increases with increasing labeled data and saturates signifying an optimum number of labeled data for a given edge node. In addition, we demonstrated that SemiPFL performance increases with increasing number of users unlike other publications highlighting the possibility to incorporate edge data heterogeneity in SemiPFL platform. By leveraging semi-supervised learning, our framework prohibitively reduces the need for annotating data. 

\highlightReference{zhang2022federated}
\highlightReference{li2021fedrs}
\highlightReference{huang2022learn}
\highlightReference{li2022data}
\highlightReference{nguyen2021federated}
\highlightReference{ha2016hypernetworks}
\highlightReference{mendieta2022local}
\highlightReference{brik2020federated}
\highlightReference{ferrag2021federated}
\highlightReference{stisen2015smart}
\highlightReference{goldberger2000physiobank}
\highlightReference{ding2022federated}
\highlightReference{cho2022flame}
\highlightReference{luping2019cmfl}
\highlightReference{khatun2022deep}
\highlightReference{tan2022towards}
\highlightReference{wu2020personalized}
\highlightReference{pei2022personalized}
\highlightReference{hu2020personalized}
\highlightReference{imteaj2021survey}
\bibliographystyle{unsrt}
\bibliography{references}

\begin{thebibliography}{10}

\bibitem{yousefi2017survey}
Siamak Yousefi, Hirokazu Narui, Sankalp Dayal, Stefano Ermon, and Shahrokh
  Valaee.
\newblock A survey on behavior recognition using wifi channel state
  information.
\newblock {\em IEEE Communications Magazine}, 55(10):98--104, 2017.

\bibitem{chen2012sensor}
Liming Chen, Jesse Hoey, Chris~D Nugent, Diane~J Cook, and Zhiwen Yu.
\newblock Sensor-based activity recognition.
\newblock {\em IEEE Transactions on Systems, Man, and Cybernetics, Part C
  (Applications and Reviews)}, 42(6):790--808, 2012.

\bibitem{ravi2005activity}
Nishkam Ravi, Nikhil Dandekar, Preetham Mysore, and Michael~L Littman.
\newblock Activity recognition from accelerometer data.
\newblock In {\em Aaai}, volume~5, pages 1541--1546. Pittsburgh, PA, 2005.

\bibitem{supratak2017deepsleepnet}
Akara Supratak, Hao Dong, Chao Wu, and Yike Guo.
\newblock Deepsleepnet: A model for automatic sleep stage scoring based on raw
  single-channel eeg.
\newblock {\em IEEE Transactions on Neural Systems and Rehabilitation
  Engineering}, 25(11):1998--2008, 2017.

\bibitem{mubashir2013survey}
Muhammad Mubashir, Ling Shao, and Luke Seed.
\newblock A survey on fall detection: Principles and approaches.
\newblock {\em Neurocomputing}, 100:144--152, 2013.

\bibitem{kim2009human}
Eunju Kim, Sumi Helal, and Diane Cook.
\newblock Human activity recognition and pattern discovery.
\newblock {\em IEEE pervasive computing}, 9(1):48--53, 2009.

\bibitem{Wang2019DeepSurvey}
Jindong Wang, Yiqiang Chen, Shuji Hao, Xiaohui Peng, and Lisha Hu.
\newblock {Deep learning for sensor-based activity recognition: A survey}.
\newblock {\em Pattern Recognition Letters}, 119:3--11, 2019.

\bibitem{lara2012survey}
Oscar~D Lara and Miguel~A Labrador.
\newblock A survey on human activity recognition using wearable sensors.
\newblock {\em IEEE communications surveys \& tutorials}, 15(3):1192--1209,
  2012.

\bibitem{Saeed2021FederatedIntelligence}
Aaqib Saeed, Flora~D. Salim, Tanir Ozcelebi, and Johan Lukkien.
\newblock {Federated Self-Supervised Learning of Multisensor Representations
  for Embedded Intelligence}.
\newblock {\em IEEE Internet of Things Journal}, 8(2):1030--1040, 2021.

\bibitem{nguyen2021federated}
Dinh~C Nguyen, Ming Ding, Pubudu~N Pathirana, Aruna Seneviratne, Jun Li, and
  H~Vincent Poor.
\newblock Federated learning for internet of things: A comprehensive survey.
\newblock {\em IEEE Communications Surveys \& Tutorials}, 23(3):1622--1658,
  2021.

\bibitem{ferrag2021federated}
Mohamed~Amine Ferrag, Othmane Friha, Leandros Maglaras, Helge Janicke, and Lei
  Shu.
\newblock Federated deep learning for cyber security in the internet of things:
  Concepts, applications, and experimental analysis.
\newblock {\em IEEE Access}, 9:138509--138542, 2021.

\bibitem{zhao2020semi}
Yuchen Zhao, Hanyang Liu, Honglin Li, Payam Barnaghi, and Hamed Haddadi.
\newblock Semi-supervised federated learning for activity recognition.
\newblock {\em arXiv preprint arXiv:2011.00851}, 2020.

\bibitem{imteaj2021survey}
Ahmed Imteaj, Urmish Thakker, Shiqiang Wang, Jian Li, and M~Hadi Amini.
\newblock A survey on federated learning for resource-constrained iot devices.
\newblock {\em IEEE Internet of Things Journal}, 9(1):1--24, 2021.

\bibitem{Liu2020DeepApproach}
Yi~Liu, Sahil Garg, Jiangtian Nie, Yang Zhang, Zehui Xiong, Jiawen Kang, and
  M~Shamim Hossain.
\newblock Deep anomaly detection for time-series data in industrial iot: A
  communication-efficient on-device federated learning approach.
\newblock {\em IEEE Internet of Things Journal}, 8(8):6348--6358, 2020.

\bibitem{zhang2020blockchain}
Weishan Zhang, Qinghua Lu, Qiuyu Yu, Zhaotong Li, Yue Liu, Sin~Kit Lo, Shiping
  Chen, Xiwei Xu, and Liming Zhu.
\newblock Blockchain-based federated learning for device failure detection in
  industrial iot.
\newblock {\em IEEE Internet of Things Journal}, 8(7):5926--5937, 2020.

\bibitem{liu2020federated}
Yi~Liu, Jiangtian Nie, Xuandi Li, Syed~Hassan Ahmed, Wei Yang~Bryan Lim, and
  Chunyan Miao.
\newblock Federated learning in the sky: Aerial-ground air quality sensing
  framework with uav swarms.
\newblock {\em IEEE Internet of Things Journal}, 8(12):9827--9837, 2020.

\bibitem{Liuyi2020IPT}
Yi~Liu, James J.~Q. Yu, Jiawen Kang, Dusit Niyato, and Shuyu Zhang.
\newblock Privacy-preserving traffic flow prediction: A federated learning
  approach.
\newblock {\em IEEE Internet of Things Journal}, 7(8):7751--7763, 2020.

\bibitem{pei2022personalized}
Jiaming Pei, Kaiyang Zhong, Mian~Ahmad Jan, and Jinhai Li.
\newblock Personalized federated learning framework for network traffic anomaly
  detection.
\newblock {\em Computer Networks}, 209:108906, 2022.

\bibitem{yu2020deep}
Shuai Yu, Xu~Chen, Zhi Zhou, Xiaowen Gong, and Di~Wu.
\newblock When deep reinforcement learning meets federated learning:
  Intelligent multitimescale resource management for multiaccess edge computing
  in 5g ultradense network.
\newblock {\em IEEE Internet of Things Journal}, 8(4):2238--2251, 2020.

\bibitem{saha2020fogfl}
Rituparna Saha, Sudip Misra, and Pallav~Kumar Deb.
\newblock Fogfl: Fog-assisted federated learning for resource-constrained iot
  devices.
\newblock {\em IEEE Internet of Things Journal}, 8(10):8456--8463, 2020.

\bibitem{nguyen2020efficient}
Van-Dinh Nguyen, Shree~Krishna Sharma, Thang~X Vu, Symeon Chatzinotas, and
  Bj{\"o}rn Ottersten.
\newblock Efficient federated learning algorithm for resource allocation in
  wireless iot networks.
\newblock {\em IEEE Internet of Things Journal}, 8(5):3394--3409, 2020.

\bibitem{xue2021resource}
Zeyue Xue, Pan Zhou, Zichuan Xu, Xiumin Wang, Yulai Xie, Xiaofeng Ding, and
  Shiping Wen.
\newblock A resource-constrained and privacy-preserving edge-computing-enabled
  clinical decision system: A federated reinforcement learning approach.
\newblock {\em IEEE Internet of Things Journal}, 8(11):9122--9138, 2021.

\bibitem{chhikara2020federated}
Prateek Chhikara, Prabhjot Singh, Rajkumar Tekchandani, Neeraj Kumar, and
  Mohsen Guizani.
\newblock Federated learning meets human emotions: A decentralized framework
  for human--computer interaction for iot applications.
\newblock {\em IEEE Internet of Things Journal}, 8(8):6949--6962, 2020.

\bibitem{kwon2020multiagent}
Dohyun Kwon, Joohyung Jeon, Soohyun Park, Joongheon Kim, and Sungrae Cho.
\newblock Multiagent ddpg-based deep learning for smart ocean federated
  learning iot networks.
\newblock {\em IEEE Internet of Things Journal}, 7(10):9895--9903, 2020.

\bibitem{zhang2021dynamic}
Weishan Zhang, Tao Zhou, Qinghua Lu, Xiao Wang, Chunsheng Zhu, Haoyun Sun,
  Zhipeng Wang, Sin~Kit Lo, and Fei-Yue Wang.
\newblock Dynamic-fusion-based federated learning for covid-19 detection.
\newblock {\em IEEE Internet of Things Journal}, 8(21):15884--15891, 2021.

\bibitem{dayan2021federated}
Ittai Dayan, Holger~R Roth, Aoxiao Zhong, Ahmed Harouni, Amilcare Gentili,
  Anas~Z Abidin, Andrew Liu, Anthony~Beardsworth Costa, Bradford~J Wood,
  Chien-Sung Tsai, et~al.
\newblock Federated learning for predicting clinical outcomes in patients with
  covid-19.
\newblock {\em Nature medicine}, 27(10):1735--1743, 2021.

\bibitem{kaissis2020secure}
Georgios~A Kaissis, Marcus~R Makowski, Daniel R{\"u}ckert, and Rickmer~F
  Braren.
\newblock Secure, privacy-preserving and federated machine learning in medical
  imaging.
\newblock {\em Nature Machine Intelligence}, 2(6):305--311, 2020.

\bibitem{lim2020dynamic}
Wei Yang~Bryan Lim, Sahil Garg, Zehui Xiong, Dusit Niyato, Cyril Leung, Chunyan
  Miao, and Mohsen Guizani.
\newblock Dynamic contract design for federated learning in smart healthcare
  applications.
\newblock {\em IEEE Internet of Things Journal}, 8(23):16853--16862, 2020.

\bibitem{Tang2021SelfHAR:Data}
Chi~Ian Tang, Ignacio Perez-Pozuelo, Dimitris Spathis, Soren Brage, Nick
  Wareham, and Cecilia Mascolo.
\newblock {SelfHAR: Improving Human Activity Recognition through Self-training
  with Unlabeled Data}.
\newblock {\em Proceedings of the ACM on Interactive, Mobile, Wearable and
  Ubiquitous Technologies}, 5(1), 3 2021.

\bibitem{Zhao2020Semi-supervisedRecognition}
Yuchen Zhao, Hanyang Liu, Honglin Li, Payam Barnaghi, and Hamed Haddadi.
\newblock {Semi-supervised Federated Learning for Activity Recognition}.
\newblock {\em ACM Transactions on Intelligent Systems and Technology},
  1(1):1--21, 2020.

\bibitem{Chen2021FedHealthHealthcare}
Yiqiang Chen, Wang Lu, Jindong Wang, and Xin Qin.
\newblock Fedhealth 2: Weighted federated transfer learning via batch
  normalization for personalized healthcare.
\newblock {\em arXiv preprint arXiv:2106.01009}, 2021.

\bibitem{Chen2019FedHealth:Healthcare}
Yiqiang Chen, Xin Qin, Jindong Wang, Chaohui Yu, and Wen Gao.
\newblock Fedhealth: A federated transfer learning framework for wearable
  healthcare.
\newblock {\em IEEE Intelligent Systems}, 35(4):83--93, 2020.

\bibitem{yu2021fedhar}
Hongzheng Yu, Zekai Chen, Xiao Zhang, Xu~Chen, Fuzhen Zhuang, Hui Xiong, and
  Xiuzhen Cheng.
\newblock Fedhar: Semi-supervised online learning for personalized federated
  human activity recognition.
\newblock {\em IEEE Transactions on Mobile Computing}, 2021.

\bibitem{Lim2019FederatedSurvey}
Wei Yang~Bryan Lim, Nguyen~Cong Luong, Dinh~Thai Hoang, Yutao Jiao, Ying-Chang
  Liang, Qiang Yang, Dusit Niyato, and Chunyan Miao.
\newblock Federated learning in mobile edge networks: A comprehensive survey.
\newblock {\em IEEE Communications Surveys \& Tutorials}, 22(3):2031--2063,
  2020.

\bibitem{MLSYS2019_bd686fd6}
Keith Bonawitz, Hubert Eichner, Wolfgang Grieskamp, Dzmitry Huba, Alex
  Ingerman, Vladimir Ivanov, Chlo\'{e} Kiddon, Jakub Kone\v{c}n\'{y}, Stefano
  Mazzocchi, Brendan McMahan, Timon Van~Overveldt, David Petrou, Daniel Ramage,
  and Jason Roselander.
\newblock Towards federated learning at scale: System design.
\newblock In A.~Talwalkar, V.~Smith, and M.~Zaharia, editors, {\em Proceedings
  of Machine Learning and Systems}, volume~1, pages 374--388, 2019.

\bibitem{xu2021federated}
Jie Xu, Benjamin~S Glicksberg, Chang Su, Peter Walker, Jiang Bian, and Fei
  Wang.
\newblock Federated learning for healthcare informatics.
\newblock {\em Journal of Healthcare Informatics Research}, 5(1):1--19, 2021.

\bibitem{mcmahan2017communicationefficient}
H.~Brendan McMahan, Eider Moore, Daniel Ramage, Seth Hampson, and
  Blaise~Agüera y~Arcas.
\newblock Communication-efficient learning of deep networks from decentralized
  data, 2017.

\bibitem{Bettini2021PersonalizedRecognition}
Riccardo Presotto, Gabriele Civitarese, and Claudio Bettini.
\newblock Semi-supervised and personalized federated activity recognition based
  on active learning and label propagation.
\newblock {\em Personal and Ubiquitous Computing}, 26(5):1281--1298, 2022.

\bibitem{li2020federated}
Tian Li, Anit~Kumar Sahu, Manzil Zaheer, Maziar Sanjabi, Ameet Talwalkar, and
  Virginia Smith.
\newblock Federated optimization in heterogeneous networks.
\newblock {\em Proceedings of Machine Learning and Systems}, 2:429--450, 2020.

\bibitem{Aledhari2020FederatedApplications}
Mohammed Aledhari, Rehma Razzak, Reza~M. Parizi, and Fahad Saeed.
\newblock {Federated Learning: A Survey on Enabling Technologies, Protocols,
  and Applications}.
\newblock {\em IEEE Access}, 8:140699--140725, 2020.

\bibitem{brik2020federated}
Bouziane Brik, Adlen Ksentini, and Maha Bouaziz.
\newblock Federated learning for uavs-enabled wireless networks: Use cases,
  challenges, and open problems.
\newblock {\em IEEE Access}, 8:53841--53849, 2020.

\bibitem{cho2022flame}
Hyunsung Cho, Akhil Mathur, and Fahim Kawsar.
\newblock Flame: Federated learning across multi-device environments.
\newblock {\em arXiv preprint arXiv:2202.08922}, 2022.

\bibitem{tan2022towards}
Alysa~Ziying Tan, Han Yu, Lizhen Cui, and Qiang Yang.
\newblock Towards personalized federated learning.
\newblock {\em IEEE Transactions on Neural Networks and Learning Systems},
  2022.

\bibitem{li2021fedrs}
Xin-Chun Li and De-Chuan Zhan.
\newblock Fedrs: Federated learning with restricted softmax for label
  distribution non-iid data.
\newblock In {\em Proceedings of the 27th ACM SIGKDD Conference on Knowledge
  Discovery \& Data Mining}, pages 995--1005, 2021.

\bibitem{zhang2022federated}
Jie Zhang, Zhiqi Li, Bo~Li, Jianghe Xu, Shuang Wu, Shouhong Ding, and Chao Wu.
\newblock Federated learning with label distribution skew via logits
  calibration.
\newblock In {\em International Conference on Machine Learning}, pages
  26311--26329. PMLR, 2022.

\bibitem{li2022data}
Zonghang Li, Yihong He, Hongfang Yu, Jiawen Kang, Xiaoping Li, Zenglin Xu, and
  Dusit Niyato.
\newblock Data heterogeneity-robust federated learning via group client
  selection in industrial iot.
\newblock {\em IEEE Internet of Things Journal}, 2022.

\bibitem{huang2022learn}
Wenke Huang, Mang Ye, and Bo~Du.
\newblock Learn from others and be yourself in heterogeneous federated
  learning.
\newblock In {\em Proceedings of the IEEE/CVF Conference on Computer Vision and
  Pattern Recognition}, pages 10143--10153, 2022.

\bibitem{mendieta2022local}
Matias Mendieta, Taojiannan Yang, Pu~Wang, Minwoo Lee, Zhengming Ding, and Chen
  Chen.
\newblock Local learning matters: Rethinking data heterogeneity in federated
  learning.
\newblock In {\em Proceedings of the IEEE/CVF Conference on Computer Vision and
  Pattern Recognition}, pages 8397--8406, 2022.

\bibitem{wu2020personalized}
Qiong Wu, Kaiwen He, and Xu~Chen.
\newblock Personalized federated learning for intelligent iot applications: A
  cloud-edge based framework.
\newblock {\em IEEE Open Journal of the Computer Society}, 1:35--44, 2020.

\bibitem{hu2020personalized}
Rui Hu, Yuanxiong Guo, Hongning Li, Qingqi Pei, and Yanmin Gong.
\newblock Personalized federated learning with differential privacy.
\newblock {\em IEEE Internet of Things Journal}, 7(10):9530--9539, 2020.

\bibitem{fallah2020personalized}
Alireza Fallah, Aryan Mokhtari, and Asuman Ozdaglar.
\newblock Personalized federated learning with theoretical guarantees: A
  model-agnostic meta-learning approach.
\newblock {\em Advances in Neural Information Processing Systems},
  33:3557--3568, 2020.

\bibitem{Shamsian2021PersonalizedHypernetworks}
Aviv Shamsian, Aviv Navon, Ethan Fetaya, and Gal Chechik.
\newblock Personalized federated learning using hypernetworks.
\newblock In {\em International Conference on Machine Learning}, pages
  9489--9502. PMLR, 2021.

\bibitem{Dinh2020PersonalizedEnvelopes}
Canh T~Dinh, Nguyen Tran, and Josh Nguyen.
\newblock Personalized federated learning with moreau envelopes.
\newblock {\em Advances in Neural Information Processing Systems},
  33:21394--21405, 2020.

\bibitem{achituve2021personalized}
Idan Achituve, Aviv Shamsian, Aviv Navon, Gal Chechik, and Ethan Fetaya.
\newblock Personalized federated learning with gaussian processes.
\newblock {\em Advances in Neural Information Processing Systems}, 34, 2021.

\bibitem{Fallah2020PersonalizedApproach}
Alireza Fallah, Aryan Mokhtari, and Asuman Ozdaglar.
\newblock {Personalized federated learning: A meta-learning approach}.
\newblock {\em arXiv}, pages 1--29, 2020.

\bibitem{li2021ditto}
Tian Li, Shengyuan Hu, Ahmad Beirami, and Virginia Smith.
\newblock Ditto: Fair and robust federated learning through personalization.
\newblock In {\em International Conference on Machine Learning}, pages
  6357--6368. PMLR, 2021.

\bibitem{cifar10}
Alex Krizhevsky, Vinod Nair, and Geoffrey Hinton.
\newblock Cifar-10 (canadian institute for advanced research).

\bibitem{cifar100}
2009 Krizhevsky.
\newblock Cifar100.

\bibitem{yang2015deep}
Jianbo Yang, Minh~Nhut Nguyen, Phyo~Phyo San, Xiao~Li Li, and Shonali
  Krishnaswamy.
\newblock Deep convolutional neural networks on multichannel time series for
  human activity recognition.
\newblock In {\em Twenty-fourth international joint conference on artificial
  intelligence}, 2015.

\bibitem{ronao2016human}
Charissa~Ann Ronao and Sung-Bae Cho.
\newblock Human activity recognition with smartphone sensors using deep
  learning neural networks.
\newblock {\em Expert systems with applications}, 59:235--244, 2016.

\bibitem{hammerla2016deep}
Nils~Y Hammerla, Shane Halloran, and Thomas Pl{\"o}tz.
\newblock Deep, convolutional, and recurrent models for human activity
  recognition using wearables.
\newblock {\em arXiv preprint arXiv:1604.08880}, 2016.

\bibitem{van2020survey}
Jesper~E Van~Engelen and Holger~H Hoos.
\newblock A survey on semi-supervised learning.
\newblock {\em Machine Learning}, 109(2):373--440, 2020.

\bibitem{Saeed2021FederatedIntelligenceb}
Aaqib Saeed, Flora~D. Salim, Tanir Ozcelebi, and Johan Lukkien.
\newblock {Federated Self-Supervised Learning of Multisensor Representations
  for Embedded Intelligence}.
\newblock {\em IEEE Internet of Things Journal}, 8(2):1030--1040, 2021.

\bibitem{goodfellow2016deep}
Ian Goodfellow, Yoshua Bengio, and Aaron Courville.
\newblock {\em Deep learning}.
\newblock MIT press, 2016.

\bibitem{Kairouz2019AdvancesLearning}
Peter Kairouz, H~Brendan McMahan, Brendan Avent, Aur{\'e}lien Bellet, Mehdi
  Bennis, Arjun~Nitin Bhagoji, Kallista Bonawitz, Zachary Charles, Graham
  Cormode, Rachel Cummings, et~al.
\newblock Advances and open problems in federated learning.
\newblock {\em Foundations and Trends{\textregistered} in Machine Learning},
  14(1--2):1--210, 2021.

\bibitem{li2021fedbn}
Xiaoxiao Li, Meirui Jiang, Xiaofei Zhang, Michael Kamp, and Qi~Dou.
\newblock Fedbn: Federated learning on non-iid features via local batch
  normalization.
\newblock {\em arXiv preprint arXiv:2102.07623}, 2021.

\bibitem{arivazhagan2019federated}
Manoj~Ghuhan Arivazhagan, Vinay Aggarwal, Aaditya~Kumar Singh, and Sunav
  Choudhary.
\newblock Federated learning with personalization layers.
\newblock {\em arXiv preprint arXiv:1912.00818}, 2019.

\bibitem{shamsian2021personalized}
Aviv Shamsian, Aviv Navon, Ethan Fetaya, and Gal Chechik.
\newblock Personalized federated learning using hypernetworks.
\newblock In {\em International Conference on Machine Learning}, pages
  9489--9502. PMLR, 2021.

\bibitem{ha2016hypernetworks}
David Ha, Andrew Dai, and Quoc~V Le.
\newblock Hypernetworks.
\newblock {\em arXiv preprint arXiv:1609.09106}, 2016.

\bibitem{ding2022federated}
Jie Ding, Eric Tramel, Anit~Kumar Sahu, Shuang Wu, Salman Avestimehr, and Tao
  Zhang.
\newblock Federated learning challenges and opportunities: An outlook.
\newblock In {\em ICASSP 2022-2022 IEEE International Conference on Acoustics,
  Speech and Signal Processing (ICASSP)}, pages 8752--8756. IEEE, 2022.

\bibitem{10.1007/978-3-319-62704-5_7}
Charikleia Chatzaki, Matthew Pediaditis, George Vavoulas, and Manolis
  Tsiknakis.
\newblock Human daily activity and fall recognition using a smartphone's
  acceleration sensor.
\newblock In Carsten R{\"o}cker, John O'Donoghue, Martina Ziefle, Markus
  Helfert, and William Molloy, editors, {\em Information and Communication
  Technologies for Ageing Well and e-Health}, pages 100--118, Cham, 2017.
  Springer International Publishing.

\bibitem{weiss2019smartphone}
Gary~M Weiss, Kenichi Yoneda, and Thaier Hayajneh.
\newblock Smartphone and smartwatch-based biometrics using activities of daily
  living.
\newblock {\em IEEE Access}, 7:133190--133202, 2019.

\bibitem{anguita2013public}
Davide Anguita, Alessandro Ghio, Luca Oneto, Xavier Parra~Perez, and Jorge~Luis
  Reyes~Ortiz.
\newblock A public domain dataset for human activity recognition using
  smartphones.
\newblock In {\em Proceedings of the 21th international European symposium on
  artificial neural networks, computational intelligence and machine learning},
  pages 437--442, 2013.

\bibitem{stisen2015smart}
Allan Stisen, Henrik Blunck, Sourav Bhattacharya, Thor~Siiger Prentow,
  Mikkel~Baun Kj{\ae}rgaard, Anind Dey, Tobias Sonne, and Mads~M{\o}ller
  Jensen.
\newblock Smart devices are different: Assessing and mitigatingmobile sensing
  heterogeneities for activity recognition.
\newblock In {\em Proceedings of the 13th ACM conference on embedded networked
  sensor systems}, pages 127--140, 2015.

\bibitem{reiss2012introducing}
Attila Reiss and Didier Stricker.
\newblock Introducing a new benchmarked dataset for activity monitoring.
\newblock In {\em 2012 16th international symposium on wearable computers},
  pages 108--109. IEEE, 2012.

\bibitem{reiss2012creating}
Attila Reiss and Didier Stricker.
\newblock Creating and benchmarking a new dataset for physical activity
  monitoring.
\newblock In {\em Proceedings of the 5th International Conference on PErvasive
  Technologies Related to Assistive Environments}, pages 1--8, 2012.

\bibitem{schmidt2018introducing}
Philip Schmidt, Attila Reiss, Robert Duerichen, Claus Marberger, and Kristof
  Van~Laerhoven.
\newblock Introducing wesad, a multimodal dataset for wearable stress and
  affect detection.
\newblock In {\em Proceedings of the 20th ACM international conference on
  multimodal interaction}, pages 400--408, 2018.

\bibitem{goldberger2000physiobank}
Ary~L Goldberger, Luis~AN Amaral, Leon Glass, Jeffrey~M Hausdorff, Plamen~Ch
  Ivanov, Roger~G Mark, Joseph~E Mietus, George~B Moody, Chung-Kang Peng, and
  H~Eugene Stanley.
\newblock Physiobank, physiotoolkit, and physionet: components of a new
  research resource for complex physiologic signals.
\newblock {\em circulation}, 101(23):e215--e220, 2000.

\bibitem{khatun2022deep}
Mst~Alema Khatun, Mohammad~Abu Yousuf, Sabbir Ahmed, Md~Zia Uddin, Salem~A
  Alyami, Samer Al-Ashhab, Hanan~F Akhdar, Asaduzzaman Khan, Akm Azad, and
  Mohammad~Ali Moni.
\newblock Deep cnn-lstm with self-attention model for human activity
  recognition using wearable sensor.
\newblock {\em IEEE Journal of Translational Engineering in Health and
  Medicine}, 10:1--16, 2022.

\bibitem{divi2021new}
Siddharth Divi, Yi-Shan Lin, Habiba Farrukh, and Z~Berkay Celik.
\newblock New metrics to evaluate the performance and fairness of personalized
  federated learning.
\newblock {\em arXiv preprint arXiv:2107.13173}, 2021.

\bibitem{luping2019cmfl}
WANG Luping, WANG Wei, and LI~Bo.
\newblock Cmfl: Mitigating communication overhead for federated learning.
\newblock In {\em 2019 IEEE 39th international conference on distributed
  computing systems (ICDCS)}, pages 954--964. IEEE, 2019.

\end{thebibliography}
\end{document}